\documentclass[a4paper,fleqn]{cas-dc}

\usepackage[numbers]{natbib}
\usepackage[accsupp]{axessibility}  
\usepackage{bibentry}
\usepackage{amsmath}
\usepackage{multirow}
\usepackage{tabularx}
\usepackage{makecell}
\usepackage{bbding}
\usepackage{pifont}
\usepackage{wasysym}
\usepackage{amssymb}
\usepackage{svg}
\usepackage{booktabs}
\usepackage{graphicx}
\usepackage{caption}
\usepackage{subcaption}

\newcommand{\figref}[1]{Fig.~\ref{#1}}

\newcommand{\tableref}[1]{Tab.~\ref{#1}}
\newcommand{\eqnref}[1]{Eq.~(\ref{#1})}

\newcommand{\secref}[1]{Sec.~\ref{#1}}

\newcommand{\Tabref}[1]{Table~\ref{#1}}
\newcommand{\Eqnref}[1]{Equation~(\ref{#1})}

\definecolor{Aquamarine}{rgb}{0.32, 0.70, 0.73}
\definecolor{watermelon_green}{rgb}{0, 0.42, 0.30}

\newcommand{\new}[1]{\textcolor{black}{#1}}

\def\tsc#1{\csdef{#1}{\textsc{\lowercase{#1}}\xspace}}
\tsc{WGM}
\tsc{QE}
\tsc{EP}
\tsc{PMS}
\tsc{BEC}
\tsc{DE}

\begin{document}
\let\WriteBookmarks\relax
\def\floatpagepagefraction{1}
\def\textpagefraction{.001}
\shorttitle{Dual Interaction Network}
\shortauthors{J. Noh et~al.}

\title [mode = title]{Dual Interaction Network with Cross-Image Attention for Medical Image Segmentation}

\author[1]{Jeonghyun Noh}[orcid=0000-0001-7327-1811]
\ead{wjdgus0967@pusan.ac.kr}

\credit{Conceptualization, Data Curation, Software, Methodology, Validation, Investigation, Formal analysis, Visualization, Writing - Original draft preparation}

\affiliation[1]{organization={Department of Information Convergence Engineering, Pusan National University},
                addressline={2, Busandaehak-ro 63beon-gil, Geumjeong-gu}, 
                city={Busan},
                postcode={46241}, 
                country={Republic of Korea}}

\author[2]{Wangsu Jeon}[orcid=0000-0001-8887-2513]

\ead{jws2218@naver.com}
\credit{Formal analysis, Validation, Methodology, Investigation, Visualization, Writing - Review and editing}

\affiliation[2]{organization={School of Computer Engineering, Kyungnam University},
                addressline={7, Gyeongnamdaehak-ro, Masanhappo-gu}, 
                postcode={51767}, 
                city={Changwon},
                country={Republic of Korea}}

\author[3,4]{Jinsun Park}[orcid=0000-0002-2296-819X]
\ead{jspark@pusan.ac.kr}

\credit{Conceptualization, Formal analysis, Funding acquisition, Methodology, Project administration, Resources, Supervision, Writing – review and editing}

\affiliation[3]{organization={School of Computer Science and Engineering, Pusan National University},
                addressline={2, Busandaehak-ro 63beon-gil, Geumjeong-gu}, 
                city={Busan},
                postcode={46241}, 
                country={Republic of Korea}}



\begin{abstract}
Medical image segmentation is a crucial method for assisting professionals in diagnosing various diseases through medical imaging.
However, various factors such as noise, blurriness, and low contrast often hinder the accurate diagnosis of diseases.
While numerous image enhancement techniques can mitigate these issues, they may also alter crucial information needed for accurate diagnosis in the original image.
Conventional image fusion strategies, such as feature concatenation can address this challenge.
However, they struggle to fully leverage the advantages of both original and enhanced images while suppressing the side effects of the enhancements.
To overcome the problem, we propose a dual interactive fusion module (DIFM) that effectively exploits mutual complementary information from the original and enhanced images.
DIFM employs cross-attention bidirectionally to simultaneously attend to corresponding spatial information across different images, subsequently refining the complementary features via global spatial attention.
This interaction leverages low- to high-level features implicitly associated with diverse structural attributes like edges, blobs, and object shapes, resulting in enhanced features that embody important spatial characteristics.
In addition, we introduce a multi-scale boundary loss based on gradient extraction to improve segmentation accuracy at object boundaries.
Experimental results on the ACDC and Synapse datasets demonstrate the superiority of the proposed method quantitatively and qualitatively.
Code available at: \url{https://github.com/JJeong-Gari/DIN}
\end{abstract}




\maketitle

\section{Introduction}
\label{sec:intro}
\begin{figure}
    \centering
    \renewcommand{\arraystretch}{0.2}
    \begin{tabular}{@{}c@{\hskip 0.003\linewidth}c@{\hskip 0.003\linewidth}c@{\hskip 0.003\linewidth}c@{\hskip 0.003\linewidth}}
        \includegraphics[width=0.245\linewidth]{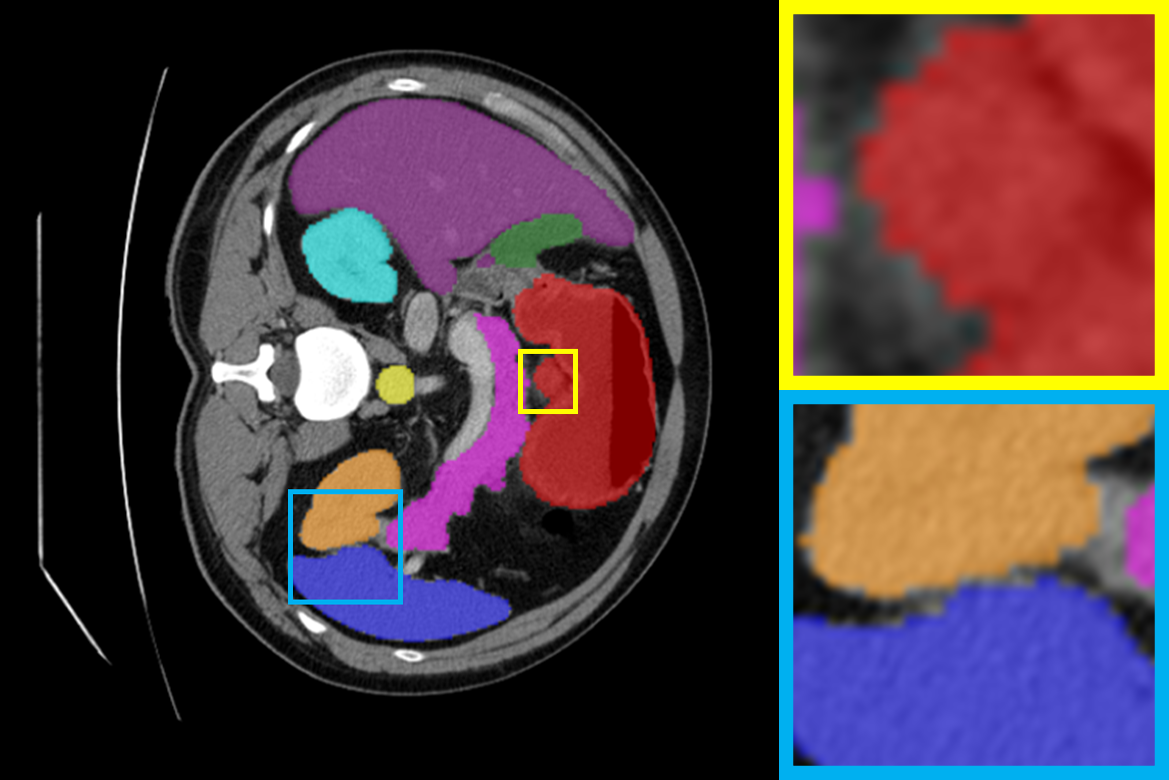} & 
        \includegraphics[width=0.245\linewidth]{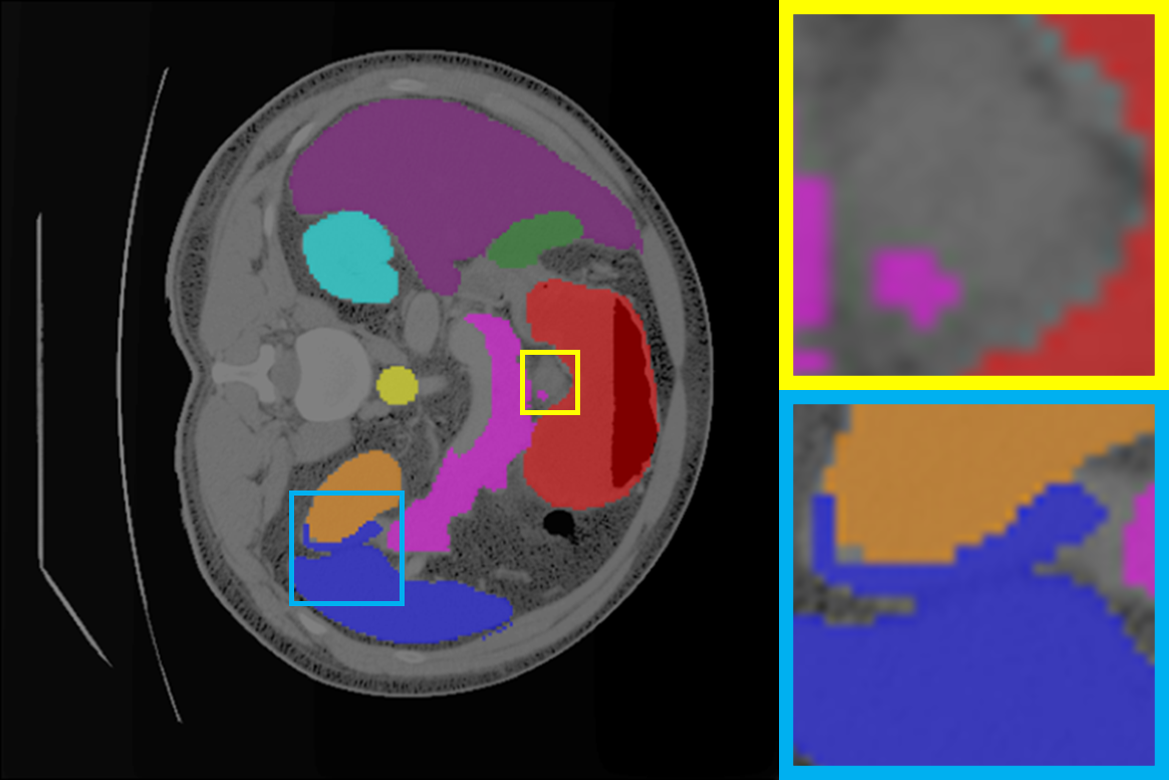} & 
        \includegraphics[width=0.245\linewidth]{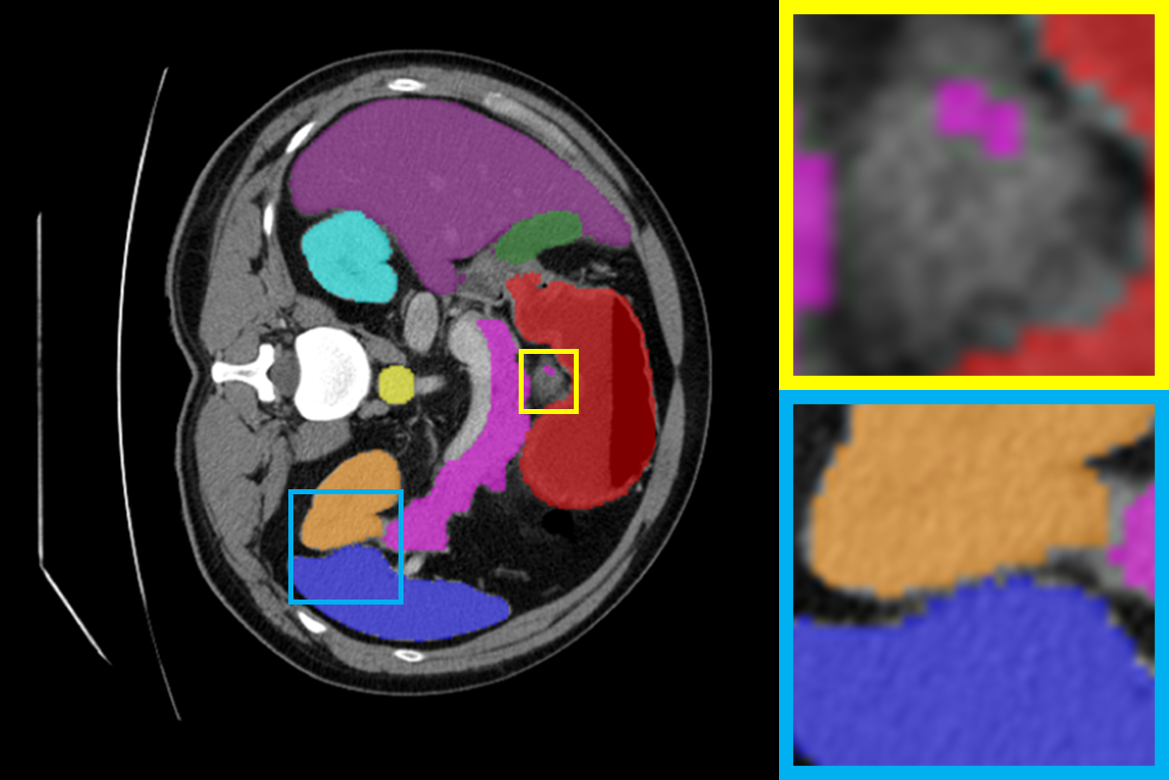} &
        \includegraphics[width=0.245\linewidth]{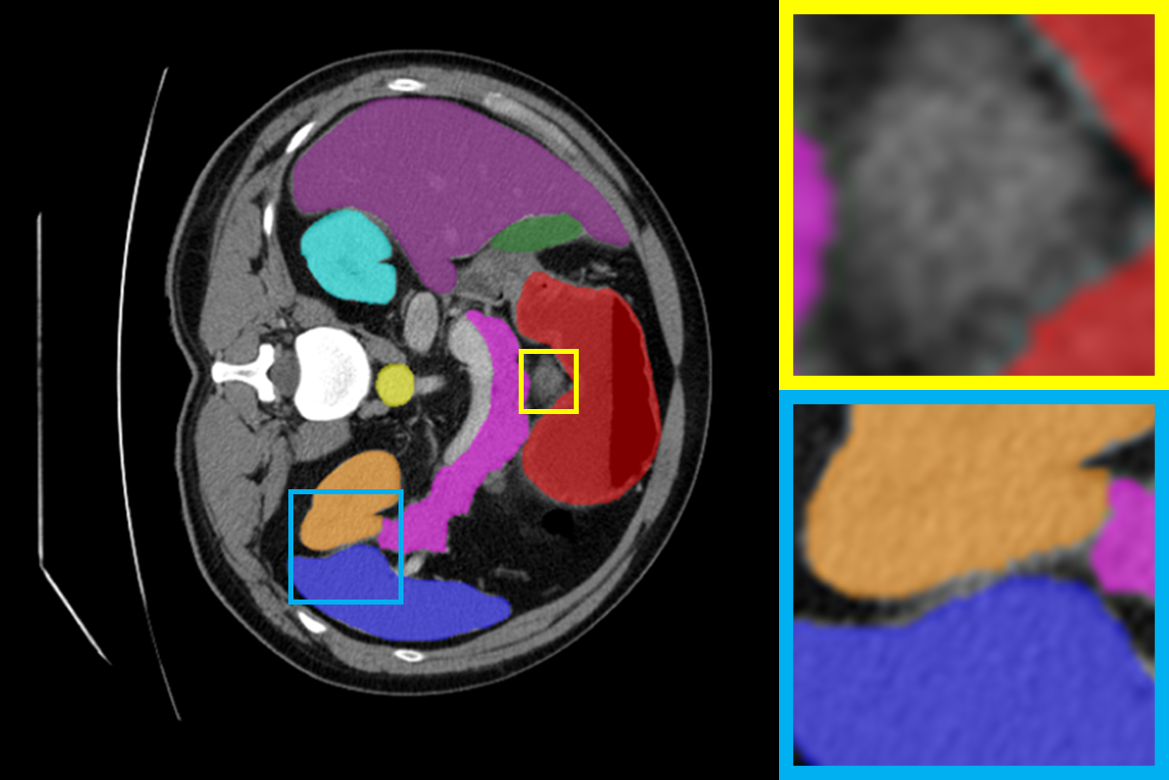} \\
        \footnotesize (a) & \footnotesize (b) & \footnotesize (c) & \footnotesize (d)
    \end{tabular}
    \caption{Medical image segmentation results with various input configurations: (a) Original image only, (b) Enhanced image only, (c) Proposed, and (d) Ground truth. The proposed method shows more accurate prediction compared to single-image outputs (a) and (b) because it benefits from both the original and enhanced images effectively.}
    \label{fig:teaser}
\end{figure}

Medical image segmentation plays a critical role in various diagnostic workflows, as it enables accurate delineation of anatomical structures and pathological regions, thereby enhancing disease interpretation, treatment planning, and outcome prediction.
Jang et al.~\cite{jang2020deep} has demonstrated that improved segmentation performance can lead directly to reduced diagnostic error rates and increased clinician confidence.
Building on the remarkable success of deep learning across diverse domains~\cite{lee2024crossformer,kim2024adnet,ha2024interdimensional,son2025mc}, recent progress in medical image segmentation has been primarily driven by deep learning.
Since the introduction of U-Net~\cite{ronneberger2015unet}, segmentation methods have rapidly evolved, with convolutional neural networks (CNN)-based models~\cite{zhou2018unet++,huang2020unet3+,isensee2021nnu} and Transformer-based models~\cite{chen2021transunet,cao2022swinunet,tragakis2023fully} showing superior performance in computed tomography (CT) and magnetic resonance imaging (MRI) segmentation.
However, various challenges such as blurriness, noise, and low contrast often hinder the accurate diagnosis of diseases.
Applying various image enhancement techniques to generate an enhanced image from the input can alleviate this problem, as existing methods have
demonstrated improved segmentation accuracy~\cite{gupta2021instaconvnet,iqball2022covid,saifullah2024advanced}.
Nevertheless, input images may unintentionally lose crucial information contained in the original image during enhancement.
As a result, the segmentation model suffers from performance degradation (\figref{fig:teaser}(a) and (b)).
Therefore, we argue that it is crucial to leverage the advantages of both the original and enhanced images through image fusion strategies.

One simple fusion strategy is the input fusion, which concatenates images along the channel dimension.
While applicable to previous works~\cite{huang2020unet3+, cao2022swinunet, tragakis2023fully}, it struggles to explore non-linear relationships between two images.
Layer fusion addresses this by processing images through separate encoders and merging features by addition or concatenation~\cite{dolz2018hyperdense,zhang2022mmformer,zhu2023crcnet} and cross-attention~\cite{xing2022nestedformer,zheng2023casf,lin2023few} at multiple layers.
However, simple fusion, addition, or concatenation limit their ability to selectively highlight critical features, and cross-attention typically focuses on unidirectional information flow, potentially limiting the depth of contextual comprehension, like morphological details.
Therefore, integrating diverse information sources simultaneously is essential for a deeper understanding of the context.

In this paper, we propose a novel approach leveraging both the original and enhanced images with a dual-interactive fusion module (DIFM).
Our DIFM first combines shallow and deep features extracted from image-specific encoders~\cite{liu2022convnet}.
Subsequently, bidirectional cross-attention enables the exchange of mutually complementary information between images.
Finally, global spatial attention is employed to enhance structural features.
To obtain a segmentation prediction, the fused feature is fed into a multi-layer perceptron (MLP) decoder~\cite{xie2021segformer}.
We also introduce a multi-scale boundary loss to improve segmentation accuracy on object boundaries further.
Our multi-scale boundary loss minimizes the gradient difference between prediction and ground truth (GT).
As a result, our method has achieved a state-of-the-art (SOTA) performance of 93.25\% on the automatic cardiac diagnosis challenge (ACDC)~\cite{ACDC} dataset and a competitive result of 85.49\% on the synapse multi-organ segmentation (Synapse)~\cite{Synapse} dataset.
The main contributions of this study are summarized as follows:

\begin{itemize}
\item We suggest a novel approach using the original and enhanced images for medical image segmentation.
\item We propose a DIFM that effectively leverages the advantage of both original and enhanced images, utilizing a cross-attention bidirectionally.
\item We introduce a multi-scale boundary loss based on gradient extraction to enhance segmentation accuracy, particularly at object boundaries.
\item The proposed method has achieved SOTA performance on the ACDC dataset and competitive results on the Synapse dataset.
\end{itemize}

\section{Related Work}
\label{sec:related}

\subsection{Single Image-based Methods}
U-Net~\cite{ronneberger2015unet} proposed a U-shaped, CNN-based encoder-decoder structure initially for medical image segmentation.
%
%
UNet++\cite{zhou2018unet++} improved U-Net's skip connections by using dense convolution blocks, while U-Net3+\cite{huang2020unet3+} introduced full-scale skip connections for better multi-scale feature use.
nnU-Net~\cite{isensee2021nnu} is an open-source framework that automatically configures neural networks for segmentation.

Recently, Transformer has been actively researched in numerous vision tasks, leading to the proposal of Vision Transformer (ViT)~\cite{dosovitskiy2020vit} and Swin Transformer (Swin-T)~\cite{liu2021swin}, which have shown superior results over CNNs.
TransUNet~\cite{chen2021transunet} was the first to apply the transformer architecture, ViT to medical segmentation.
SwinUNet~\cite{cao2022swinunet} proposed using Swin-T to extract and leverage multi-scale feature maps.
MissFormer~\cite{huang2022missformer} introduced the ReMix-FFN to re-integrate the local context and global dependencies.
FCT~\cite{tragakis2023fully} proposed a fully convolutional transformer structure that effectively captures long-term dependencies.
\new{
EMCAD~\cite{rahman2024emcad} proposed a multi-scale convolutional attention decoder to improve hierarchical feature integration across different resolution levels.
}

\new{
More recently, diverse strategies, such as diffusion or visual mamba, have been applied to improve medical image segmentation performance.
HiDiff~\cite{chen2024hidiff} introduced a diffusion-based framework to iteratively refine segmentation masks, improving boundary smoothness and consistency.
VM-UNet~\cite{ruan2024vm} leveraged the Vision Mamba architecture~\cite{liu2024vmamba}, a state space model (SSM)-based approach, to model long-range dependencies within a U-Net framework.
}

\subsection{Simple and Cross-attention based Fusion}
Existing methods relying on a single image struggle to capture critical diagnostic information and fail to exploit non-linear relationships, even with input fusion strategies fully.
To address this, prior works have introduced attention and concatenation-based fusion strategies.
HyperdenseNet~\cite{dolz2018hyperdense} suggested a dense connection structure for sharing network-specific features.
MAML~\cite{zhang2021modality} introduced a modality-aware module to integrate features via attention maps.
MMFormer~\cite{zhang2022mmformer} proposed a modality-correlated encoder with addition and multi-head self-attention (MHSA) for feature integration.
CRCNet~\cite{zhu2023crcnet} generated attention maps from predicted masks and feature maps, integrating them via multiplication for improved segmentation.

However, addition or concatenation-based fusion lacks selective filtering, leading to the integration of irrelevant features that negatively affect medical image segmentation.
To address this, the cross-attention mechanism~\cite{chen2021crossvit} has been employed.
NestedFormer~\cite{xing2022nestedformer} introduced a module that utilizes cross-attention between intra-information and encoder-specific feature maps.
CASF~\cite{zheng2023casf} fused different features from CNN and Transformer branches through cross-attention.
CAT-Net~\cite{lin2023few} used cross-attention between support and query images, leveraging previous masks as pseudo-labels to enhance object focus.
\new{
TranSiam~\cite{li2024transiam} presented a multi-modal feature aggregation framework that utilizes locality-aware modules to effectively combine complementary information from different modalities.
}

Previous methods, limited by one direction of cross-attention, restrict complementary interactions.
In contrast, our DIFM employs bidirectional cross-attention to enhance information exchange between original and enhanced images, producing a mutually enriched feature map.
Moreover, DIFM refines object details and spatial context through edge features and global spatial attention.

\section{Method}
\label{sec:method}

\begin{figure*}[t]
\centering
\includegraphics[width=0.92\textwidth]{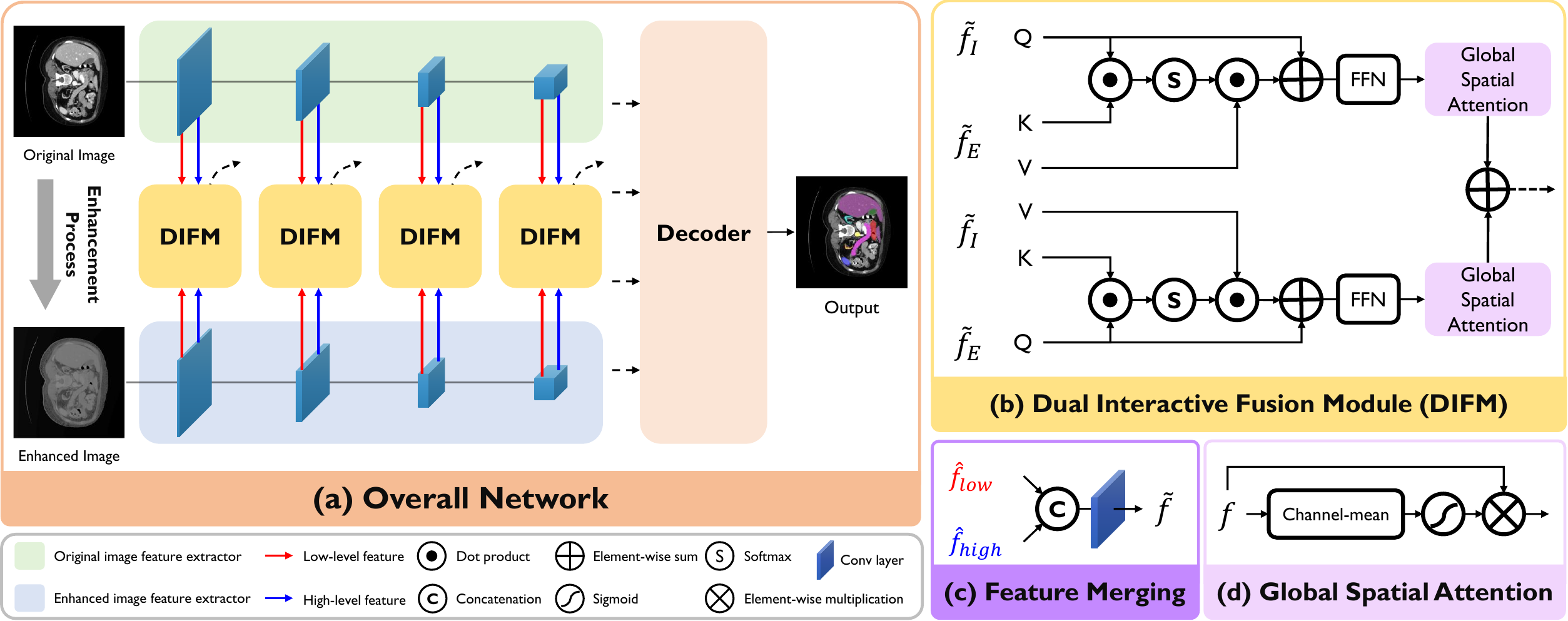}
\hfill
\caption{
The overall architecture of the proposed network.
Fuzzy image enhancement~\cite{FuzzyImageEnhancement} is used to improve the quality of the original image.
The ConvNext-base~\cite{liu2022convnet} encoder extracts image features, while the DIFM exploits the interaction between original and enhanced feature maps utilizing an attention mechanism to fuse them.
The decoder is an MLP decoder~\cite{xie2021segformer}, which generates a segmentation mask.
}
\label{fig:overall_architecture}
\end{figure*}

The proposed network comprises an encoder-decoder structure with DIFM integrated into the skip connection between the encoder and decoder, as shown in \figref{fig:overall_architecture}a.
To generate enhanced images, we use fuzzy image enhancement~\cite{FuzzyImageEnhancement}, which has proven effective in various works~\cite{gupta2021instaconvnet,iqball2022covid,noh2025object}.
ConvNext-base~\cite{liu2022convnet} and the MLP decoder~\cite{xie2021segformer} are used as the encoder and decoder.
We also introduce a multi-scale boundary loss to enhance segmentation performance around object boundaries.
Details of the proposed method will be described in the following sections.

\subsection{Dual Interactive Fusion Module (DIFM)}
\label{sec:DIFM}
In medical image segmentation, noise, blurriness, and low contrast can hinder accurate diagnosis.
While image enhancement techniques can address these issues, they risk losing critical features from the original image.
Therefore, leveraging both original and enhanced images is essential.
However, previous methods~\cite{tragakis2023fully, wang2021modality, zheng2023casf} have not fully utilized the strengths of the two images.
To tackle the problem, we propose DIFM, which facilitates effective interaction between original and enhanced images, generating mutually complementary feature maps, as illustrated in \figref{fig:overall_architecture}b.
%
%
Our network is equipped with four DIFMs to fully exploit features from the hierarchical structure.

\subsubsection{Feature Merging}
Neural networks are designed to extract progressively abstract representations of the input images.
Early-stage layers capture low-level features such as edges and textures, while deeper layers extract high-level features including shapes and semantic elements.
Specifically, precise segmentation of object boundaries requires a comprehensive understanding of both low- and high-level features.
To achieve this, we extract and fuse image-specific features at different network depths: low-level features $\hat{f}_{low}$ after the first pooling layer, providing boundary information, and high-level features $\hat{f}_{high}$ from the last layer of each encoder block, encompassing semantic characteristics.
The feature merging shown in \figref{fig:overall_architecture}c is formulated as follows: 
\begin{equation}
    \label{equ:F_merge}
    \tilde{f} = \mathtt{conv}(\mathtt{concat}(\hat{f}_{low}, \hat{f}_{high})),
\end{equation}
where $\mathtt{concat}(\cdot, \cdot)$, $\mathtt{conv}$, and $\tilde{f}$ denote the concatenation, a convolution layer, and the merged feature map, respectively.
\Eqnref{equ:F_merge} applies to each original and enhanced image.

\subsubsection{Dual Cross-attention}
The feature maps generated from the original and enhanced images by each encoder exhibit distinct feature distributions.
Fusing these features through simple fusion strategies integrates relevant and irrelevant features, potentially degrading performance (shown in \tableref{tab:fusion_strategy}).
While cross-attention~\cite{chen2021crossvit} filters irrelevant features through interaction between feature maps, it operates unidirectionally, refining only one set of features.
This restricts contextual comprehension.
To address these issues, we propose a dual cross-attention, leveraging mutually complementary information extracted from different images.
It provides a deeper understanding by integrating diverse perspectives from both the original and enhanced images.
The proposed dual cross-attention is formulated as follows:
\begin{equation}
    \label{equ:attention}
    \mathtt{Atten}(Q_{m},K_{m},V_{m}) = \mathtt{softmax}(\frac{Q_{m}K_{m}^{T}}{\sqrt{d}})V_{m},
\end{equation}
where $\mathtt{Atten}$ is the cross-attention and $\mathtt{softmax(\cdot)}$ is the Softmax function.
$Q_{m}$, $K_{m}$, and $V_{m}$ denote query, key, and value extracted from the original image feature $\hat{f}_{I}$ $(m=i)$ or enhanced image feature $\hat{f}_{E}$ $(m=e)$, respectively.
$d$ is the dimension of the feature map.
Based on \eqnref{equ:attention}, we generate dual interactive features $\tilde{f}_{IE}$ and $\tilde{f}_{EI}$ as follows:
\begin{equation}
    \begin{split}
        \label{equ:cross}
        \tilde{f}_{IE} = \mathtt{Atten}(Q_{i}, K_{e}, V_{e}) + Q_{i}, \\
        \tilde{f}_{EI} = \mathtt{Atten}(Q_{e}, K_{i}, V_{i}) + Q_{e}.
    \end{split}
\end{equation}

Our dual cross-attention operation is followed by a feed-forward network (FFN) to refine $\hat{f}_{IE}$ and $\hat{f}_{EI}$.
The FFN enhances the model's capacity for capturing complex relationships and aggregating features from different images.
The initial stage of the FFN is formulated as follows:
\begin{equation}
    \label{equ:FFN_1}
    \tilde{f}_{1}, \tilde{f}_{2} = \mathtt{chunk}(\mathtt{conv}(\mathtt{LN}(\tilde{f}))), \quad \tilde{f} \in \{\tilde{f}_{IE}, \tilde{f}_{EI}\},
\end{equation}
where $\mathtt{chunck}$ and $\mathtt{LN}$ denote the channel-wise splitting operation and layer normalization, respectively.
$\mathtt{conv}$ is to expand the feature space by doubling the channel dimension.
The subsequent stage of the FFN is formulated as follows:
\begin{equation}
    \label{equ:FFN_2}
    \hat{F} = \mathtt{conv}(\mathtt{GELU}(\tilde{f}_{1}) \otimes \tilde{f}_{2}) + \tilde{f},
\end{equation}
where $\mathtt{GELU}$ and $\otimes$ represent GELU activation function~\cite{hendrycks2016gaussian} and element-wise multiplication, respectively.
The FFN stabilizes learning through $\mathtt{LN}$ and introduces non-linearity via the GELU.
This FFN enhances the model's ability to capture complex relationships through cross-attention.

\subsubsection{Global Spatial Attention}
Our dual cross-attention is based on a window-based attention approach, computing attention independently within each window.
While effective, this method may disrupt spatial continuity at window boundaries.
To address this limitation, we implement a spatial attention mechanism for feature map refinement, integrating global feature information through mean operations to incorporate broader context into a local feature representation.
Subsequently, a sigmoid function and element-wise multiplication act as a gating mechanism, enabling dynamic modulation of specific features based on the global context.
By learning channel-wise weights, the model can adaptively prioritize or suppress different feature channels according to their importance, selectively enhancing spatial information.
Our global spatial attention shown in \figref{fig:overall_architecture}d is formulated as follows:
\begin{equation}
    \label{equ:spatial_attention}
    F = \mathtt{sigm}(\mathtt{mean}(f)) \otimes f, \quad f \in \{\hat{F}_{IE}, \hat{F}_{EI}\},
\end{equation}
where $\mathtt{sigm(\cdot)}$, $\mathtt{mean(\cdot)}$, and $\otimes$ denote the sigmoid function, channel mean operation, and element-wise multiplication with broadcasting, respectively.
To generate the final feature for prediction, $F_{IE}$ and $F_{EI}$ are generated from $\hat{F}_{IE}$ and $\hat{F}_{EI}$, respectively, and then added element-wise.

\subsection{Loss Function}
\label{sec:loss_function}
We use various loss functions to conduct efficient learning and improve segmentation performance.
We use cross-entropy (CE), Dice, and the proposed multi-scale boundary losses for training, which will be described in detail.

\subsubsection{Multi-Scale Boundary Loss}
The CE loss, which measures the probability difference between predictions and GT, is effective in segmentation tasks but sensitive to class imbalances, potentially degrading performance in skewed distributions.
To address this, we use CE loss and Dice loss, which quantify overlap and optimize for similarity, effectively managing class imbalances.
These loss functions are defined as follows:
\begin{gather}
    \label{equ:CE}
    L_{CE} = - \frac{1}{N} \sum_{i=1}^{N} \sum_{c=1}^{C} y_{ic} \log(x_{ic}),
\\
    \label{equ:Dice}
    L_{Dice} = 1 - \frac{1}{C} \sum_{c=1}^{C} \frac{2 \sum_{i=1}^{N} x_{ic} y_{ic}}{\sum_{i=1}^{N} x_{ic} + \sum_{i=1}^{N} y_{ic}},
\end{gather}
where $x_{ic}$ and $y_{ic}$ represent the prediction, GT, respectively, for the $i$-th sample and $c$-th class.
$N$, $C$, $L_{CE}$, and $L_{Dice}$ denote the total number of samples, total number of classes, CE, and Dice loss, respectively.

However, both losses may overlook fine-level object boundary details, risking inaccurate segmentation by merging adjacent segments.
To address this, Kervadec et al.~\cite{kervadec2019boundary} proposed boundary loss, a distance metric on contours using integral-based computations.
Inspired by this, we introduce a multi-scale boundary loss $L_{bnd}$, computed from the intensity difference based on a gradient extractor.
The $L_{bnd}$ extracts intensity by gradient extractor and quantifies the discrepancy between predicted and GT boundaries using the L1 norm, preserving fine-grained edge details.
Also, by computing at original and downsampled scales ($1/2$ and $1/4$), our approach captures both coarse and fine boundary details.
$L_{bnd}$ enhances segmentation accuracy by aligning predictions more closely with optimal boundaries.
The proposed $L_{bnd}$ is defined as follows:
\begin{equation}
    \label{equ:multi-scale_boundary_loss}
    L_{bnd} = \sum_{i\in \{ 1,2,4 \} }^{} |G(\mathtt{avg}(x, i)) - G(\mathtt{avg}(y, i))|_{1},
\end{equation}
where $G(\cdot)$ is the gradient extractor and $\mathtt{avg}(\cdot, i)$ denotes average pooling operation with $i \times i$ kernel.
\new{
The gradient extractor uses the Sobel filter to calculate efficiently.
To enable multi-class boundary extraction, we apply a softmax to the model output and convert the GT into a one-hot encoded format.
This formulation allows for the identification of class-specific boundaries across all target classes in a consistent and differentiable manner.
}
The total loss is defined as follows:
\begin{equation}
    \label{equ:total_loss}
    L_{total} = \alpha L_{CE} + \beta L_{Dice} + L_{bnd},
\end{equation}
where $\alpha$ and $\beta$ are set to 0.3 and 0.7, respectively, like~\cite{rahman2023multi}.

\section{Experiments}
\label{sec:experiments}

\subsection{Dataset}
Our model is evaluated on the Synapse~\cite{Synapse} and ACDC~\cite{ACDC} datasets.
Synapse is an abdominal CT image dataset with 30 images containing eight organs: the aorta, gallbladder (GB), left kidney (KL), right kidney (KR), liver, pancreas (PC), spleen (SP), and stomach (SM).
18 and 12 scans are used for training and evaluation, respectively.
ACDC provides MRI data containing three organs of 100 patients: the right ventricle (RV), left ventricle (LV), and myocardium (Myo).
70 images for training, 10 for validation, and 20 images for evaluation are used.

\subsection{Implementation Details}
\label{sec:implementation}
Experimental results are obtained using a machine with an NVIDIA A6000 GPU and PyTorch 1.12.
The AdamW optimizer was employed for 300 epochs with a learning rate of 1e-4 and a batch size of 3.
Input images are resized to 224$\times$224, and random rotation as well as horizontal and vertical flip augmentations are applied.
Evaluation metrics comprise Dice scores for ACDC and Synapse datasets and 95\% Hausdorff Distance (HD95) for Synapse datasets.
The Dice score and HD95 measure, respectively, the similarity and maximum distance between the prediction and GT.

\subsection{Experimental Results on Synapse and ACDC}
The proposed method is compared with previous methods based on the average Dice score and HD95.
Our model has achieved competitive results on the Synapse and SOTA performance on the ACDC datasets.
Detailed analyses of the experimental results will be provided in this section.

\begin{table*}
\caption{Quantitative results on the Synapse test set with previous methods.}
\label{tab:synapse}
\centering
\resizebox{0.75\textwidth}{!}{
\begin{tabular}{c|c|c|c|c|c|c|c|c|c|c}
\Xhline{3\arrayrulewidth}
Methods & Dice$\uparrow$ & HD95$\downarrow$ & Aorta & GB & KL & KR & Liver & PC & SP & SM \\ \hline\hline
UNet~\cite{ronneberger2015unet} & 70.11 & 44.69 & 84.00 & 56.70 & 72.41 & 62.64 & 86.98 & 48.73 & 81.48 & 67.96 \\
R50+AttnUNet~\cite{chen2021transunet} & 75.57 & 36.97 & 55.92 & 63.91 & 79.20 & 72.71 & 93.56 & 49.37 & 87.19 & 74.95 \\
TransUNet~\cite{chen2021transunet} & 77.48 & 31.69 & 87.23 & 63.13 & 81.87 & 77.02 & 94.08 & 55.86 & 85.08 & 75.62 \\
SwinUNet~\cite{cao2022swinunet}  & 79.13 & 21.55 & 85.47 & 66.53 & 83.28 & 79.61 & 94.29 & 56.58 & 90.66 & 76.60\\
MT-UNet~\cite{wang2022mixed} & 78.59 & 26.59 & 87.92 & 64.99 & 81.47 & 77.29 & 93.06 & 59.46 & 87.75 & 76.81 \\
MISSFormer~\cite{huang2022missformer} & 81.96 & 18.20 & 86.99 & 68.65 & 85.21 & 82.00 & 94.41 & 65.67 & 91.92 & 80.81 \\
CASTFormer~\cite{you2022class} & 82.55 & 22.73 & 89.05 & 67.48 & 86.05 & 82.17 & 95.61 & 67.49 & 91.00 & 81.55 \\
TrasnsCASCADE~\cite{rahman2023medical} & 82.68 & 17.34 & 86.63 & 68.48 & 87.66 & 84.56 & 94.43 & 65.33 & 90.79 & 83.52 \\
MERIT~\cite{rahman2023multi} & 84.90 & 13.22 & 87.71 & \textbf{74.40} & \underline{87.79} & 84.85 & 95.26 & 71.81 & \underline{92.01} & 85.38 \\
FCT~\cite{tragakis2023fully} & 83.53 & - & \underline{89.85} & 72.73 & \textbf{88.45} & \textbf{86.60} & \underline{95.62} & 66.25 & 89.77 & 79.42 \\ 
nnFormer~\cite{zhou2021nnformer} & \textbf{86.57} & \textbf{10.63} & \textbf{92.04} & 70.17 & 86.57 & \underline{86.25} & \textbf{96.84} & \textbf{83.35} & 90.51 & \textbf{86.83} \\ \hline
Ours & \underline{85.49} & \underline{10.74} & 89.12 & \underline{74.32} & 87.47 & 85.85 & 94.71 & \underline{73.82} & \textbf{92.30} & \underline{86.31} \\
\Xhline{3\arrayrulewidth}
\end{tabular}
}
\end{table*}

\begin{table}[t]
\caption{Quantitative results on the ACDC test set with previous methods.}
\label{tab:ACDC}
\centering
\resizebox{0.81\linewidth}{!}{
\begin{tabular}{c|c|c|c|c}
\Xhline{3\arrayrulewidth}
Methods & Dice$\uparrow$ & RV & Myo & LV \\ \hline\hline
R50+AttnUNet\cite{chen2021transunet} & 86.75 & 87.58 & 79.20 & 93.47 \\
TransUNet\cite{chen2021transunet} & 89.71 & 88.86 & 84.53 & 95.73 \\
SwinUNet\cite{cao2022swinunet}  & 90.00 & 88.55 & 85.62 & 95.83\\
MT-UNet\cite{wang2022mixed} & 90.43 & 86.64 & 89.04 & 95.62 \\
MISSFormer\cite{huang2022missformer} & 90.86 & 89.55 & 88.04 & 94.99 \\
nnUNet\cite{isensee2021nnu} & 91.61 & 90.24 & 89.24 & 95.36 \\
TrasnsCASCADE\cite{rahman2023medical} & 91.63 & 89.14 & 90.25 & 95.50 \\
nnFormer\cite{zhou2021nnformer} & 92.06 & 90.94 & 89.58 & 95.65 \\
MERIT\cite{rahman2023multi} & 92.32 & 90.87 & 90.00 & \underline{96.08} \\
FCT$_{224}$\cite{tragakis2023fully} & 92.84 & 92.02 & \underline{90.61} & 95.89 \\
FCT$_{384}$\cite{tragakis2023fully} & \underline{93.02} & \textbf{92.64} & 90.51 & 95.90 \\ \hline
Ours & \textbf{93.25} & \underline{92.16} & \textbf{91.08} & \textbf{96.50} \\ 
\Xhline{3\arrayrulewidth}
\end{tabular}
}
\end{table}

\begin{figure}
    \centering
    \renewcommand{\arraystretch}{0.2}
    \begin{subtable}{\linewidth}
        \centering
        \resizebox{0.90\linewidth}{!}{
        \begin{tabular}{@{}c@{\hskip 0.003\linewidth}c@{\hskip 0.003\linewidth}c@{\hskip 0.003\linewidth}c@{\hskip 0.003\linewidth}}
        \multicolumn{4}{l}{\includegraphics[width=0.91\linewidth]{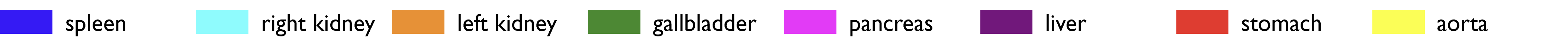}} \\
            \includegraphics[width=0.245\linewidth]{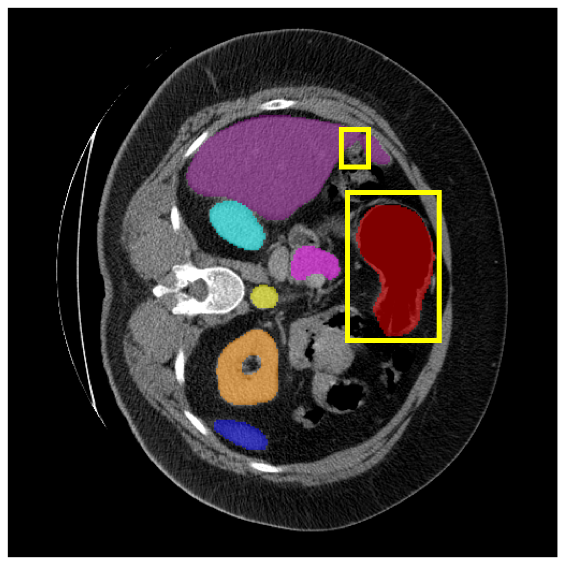} &
            \includegraphics[width=0.245\linewidth]{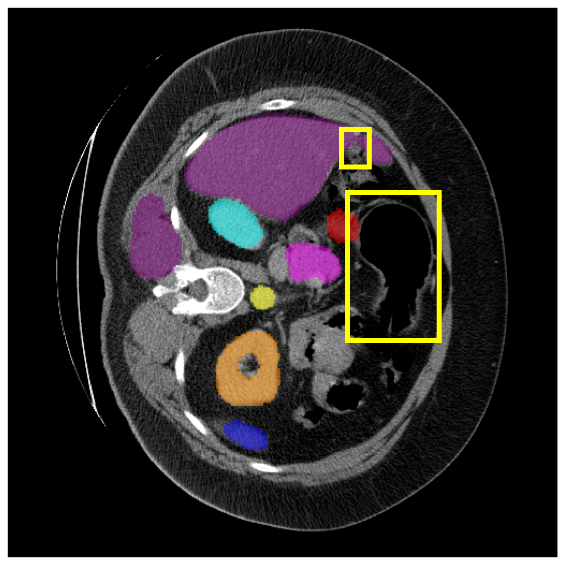} &
            \includegraphics[width=0.245\linewidth]{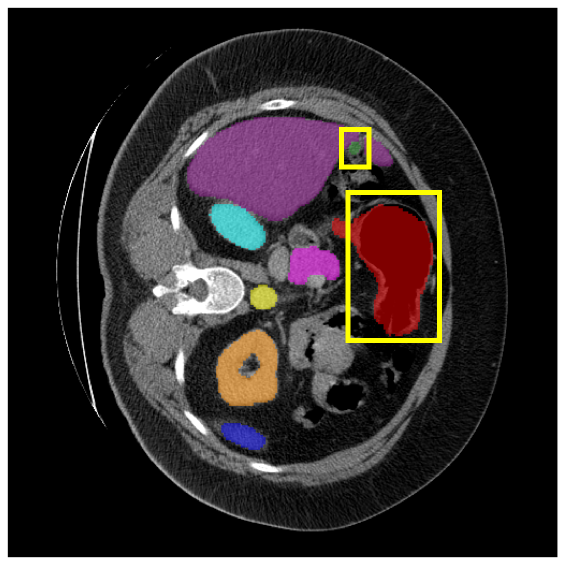} &
            \includegraphics[width=0.245\linewidth]{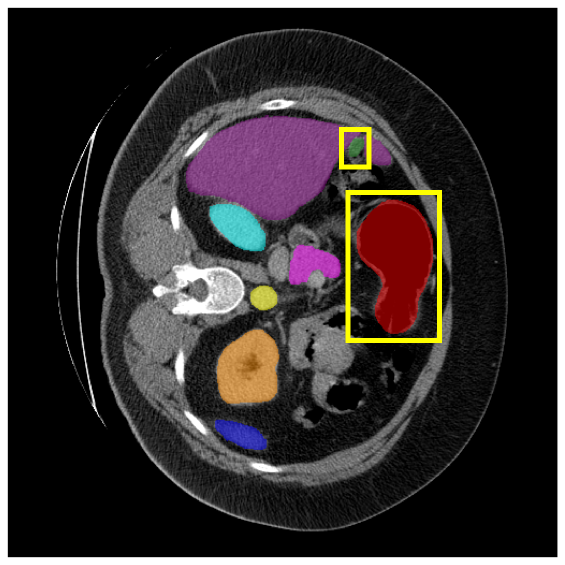} \\
            \includegraphics[width=0.245\linewidth]{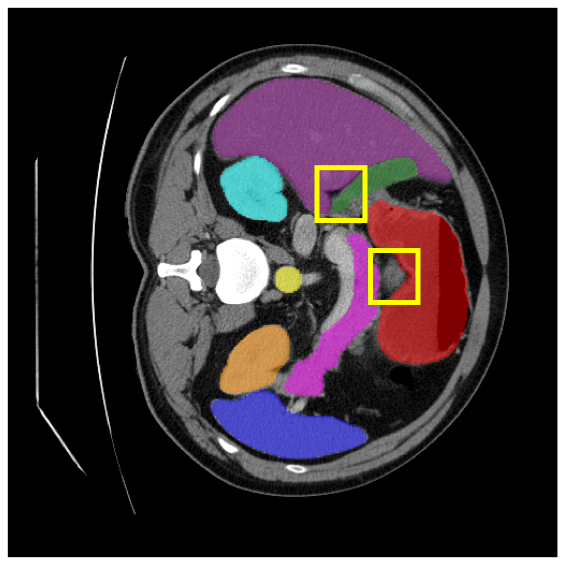} &
            \includegraphics[width=0.245\linewidth]{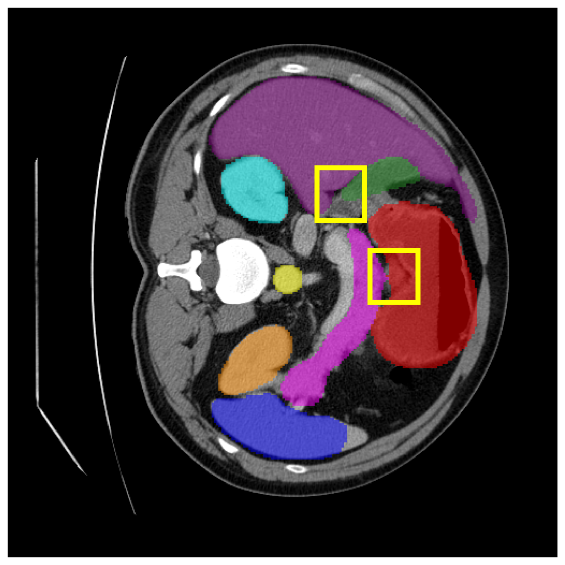} &
            \includegraphics[width=0.245\linewidth]{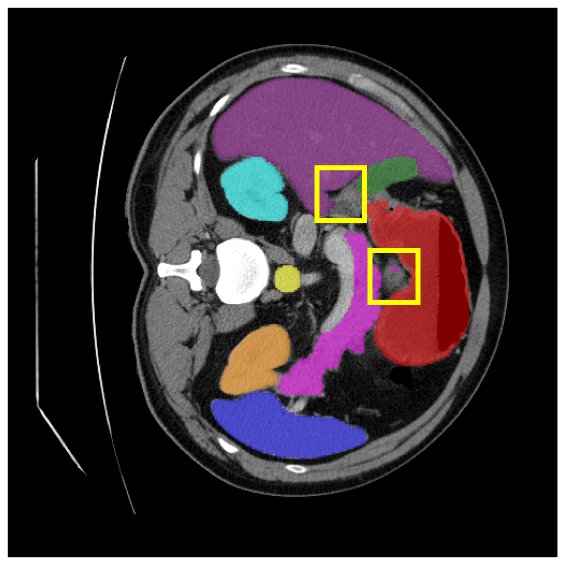} &
            \includegraphics[width=0.245\linewidth]{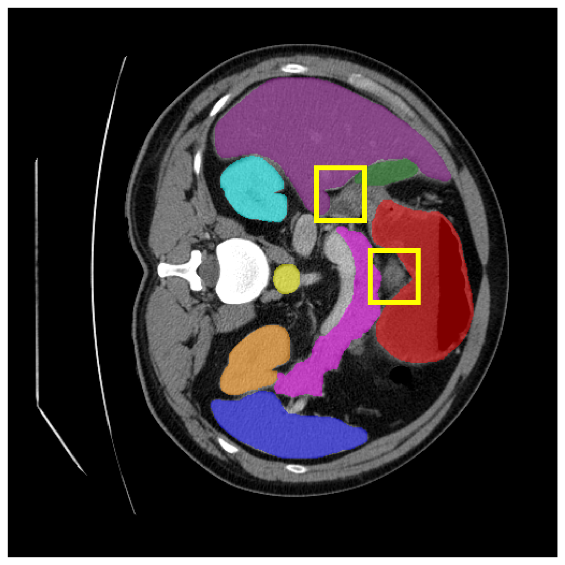} \\
            \footnotesize \footnotesize MERIT\cite{rahman2023multi} & \footnotesize FCT\cite{tragakis2023fully} & \footnotesize Ours & \footnotesize GT
        \end{tabular}
        }
        \caption{}
        \label{subfig:synapse}
    \end{subtable}
    \begin{subtable}{\linewidth}
        \centering
            \resizebox{0.90\linewidth}{!}{
        \begin{tabular}{@{}c@{\hskip 0.003\linewidth}c@{\hskip 0.003\linewidth}c@{\hskip 0.003\linewidth}c@{\hskip 0.003\linewidth}}
        \multicolumn{2}{l}{\includegraphics[width=0.430\linewidth]{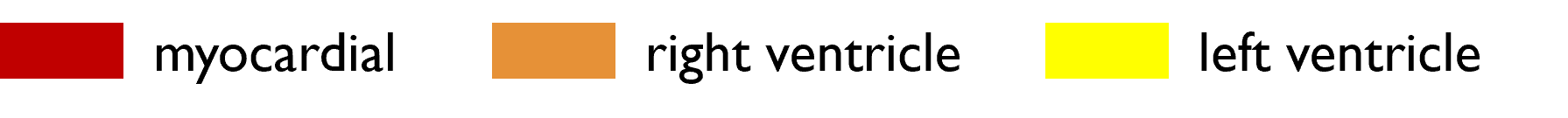}} \\
            \includegraphics[width=0.245\linewidth]{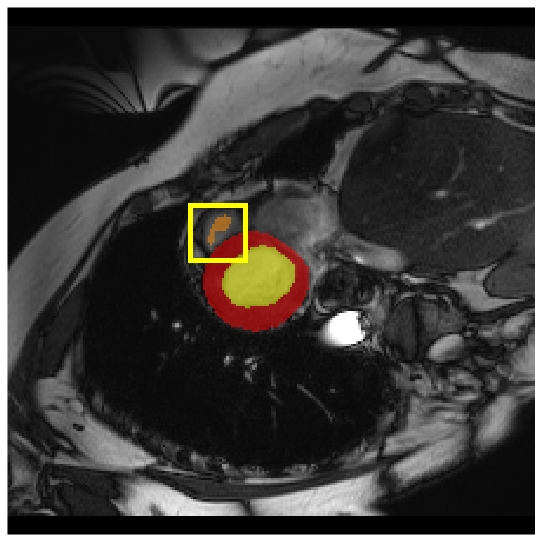} &
            \includegraphics[width=0.245\linewidth]{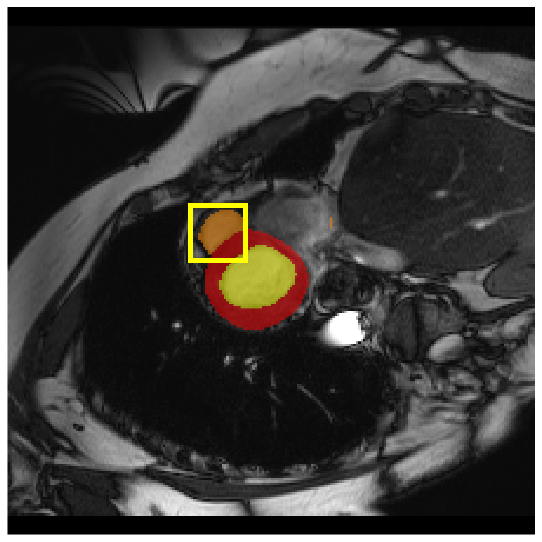} &
            \includegraphics[width=0.245\linewidth]{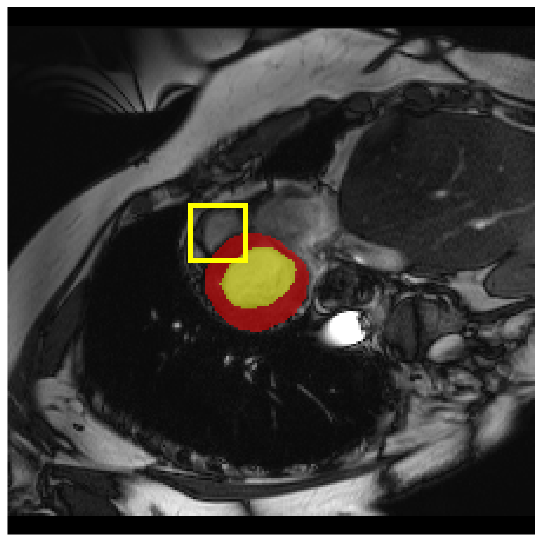} &
            \includegraphics[width=0.245\linewidth]{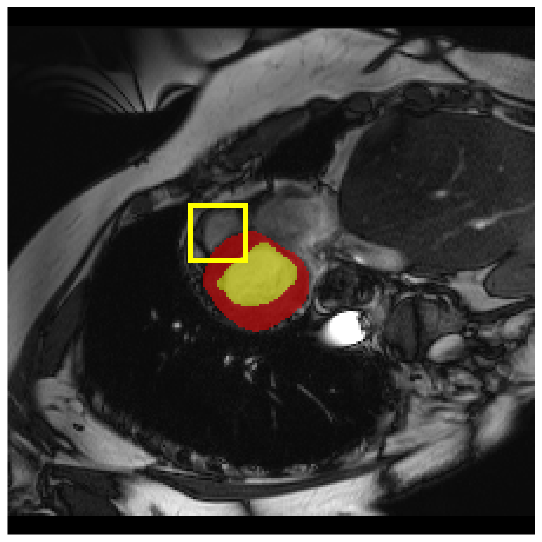} \\
            \includegraphics[width=0.245\linewidth]{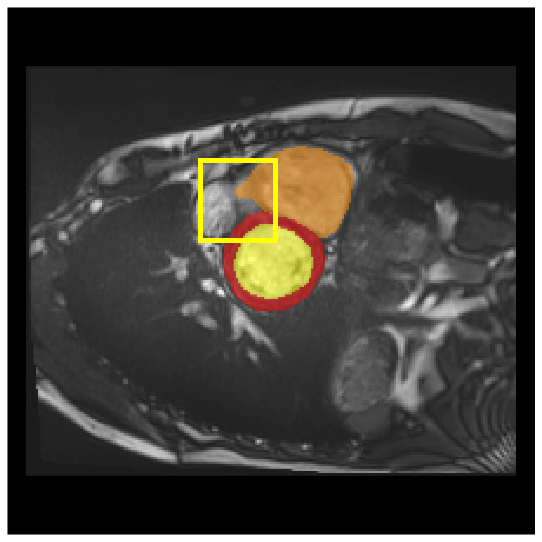} &
            \includegraphics[width=0.245\linewidth]{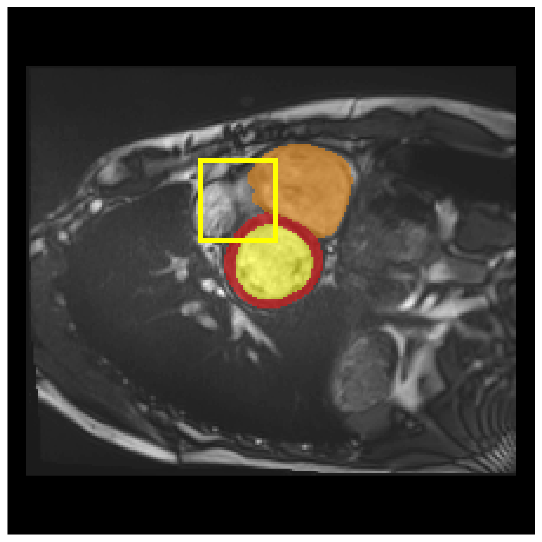} &
            \includegraphics[width=0.245\linewidth]{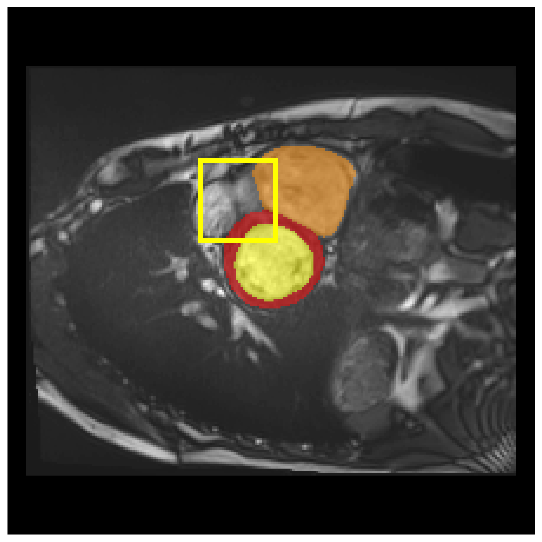} &
            \includegraphics[width=0.245\linewidth]{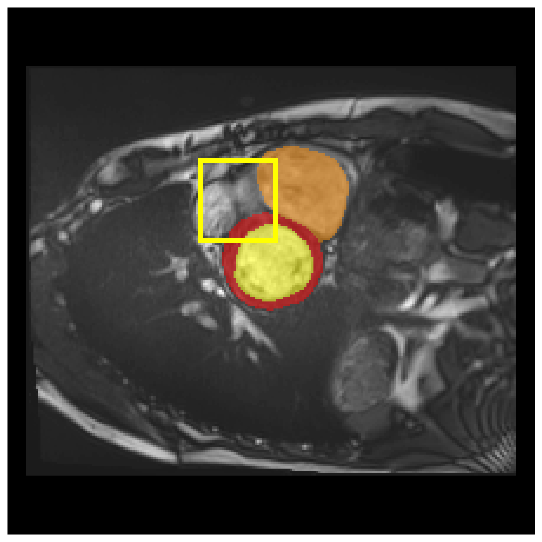} \\
            \footnotesize \footnotesize MERIT\cite{rahman2023multi} & \footnotesize FCT\cite{tragakis2023fully} & \footnotesize Ours & \footnotesize GT
        \end{tabular}
        }
        \caption{}
        \label{subfig:acdc}
    \end{subtable}
    \label{fig:visual_comparsion}
    \caption{
    Visual comparisons of segmentation results on the (a) Synapse and (b) ACDC datasets.
    Yellow boxes highlight regions in which our method excels at segmentation.
    }
\end{figure}

\subsubsection{Quantitative Results}

\Tabref{tab:synapse} provides quantitative evaluation results on the Synapse dataset.
Note that bold and underlined values indicate the best and second-best results.
The proposed method achieved the second-best performance, with a Dice score of 85.49\% and an HD95 of 10.74, following nnFormer~\cite{zhou2021nnformer}, which exploits 3D volumetric information instead of 2d images.
Despite this, ours outperformed nnFormer in GB and SP, with Dice score improvements of 4.15\%p and 1.79\%p, respectively.
Also, significant improvements were observed over MERIT~\cite{rahman2023multi} and FCT~\cite{tragakis2023fully} in terms of PC, SP, and SM. 
This is attributed to the integration of enhanced images that highlight object details.
A notable decrease of 2.48 in HD95 compared to MERIT indicates the effectiveness of our method in capturing boundary spatial context.

Experimental results on the ACDC datasets are presented in \tableref{tab:ACDC}.
FCT$_{224}$ and FCT$_{384}$ represent the FCT applied to images resized to 224$\times$224 and 384$\times$384, respectively.
Our model achieved the SOTA performance of 93.25\%, surpassing the FCT$_{224}$ and FCT$_{384}$, which achieved accuracy of 92.84\% and 93.02\%, respectively.
Notably, our approach demonstrated superior performance for the Myo and LV, with Dice scores of 91.08\% and 96.50\%.
By effectively extracting and integrating features from original and enhanced images using DIFM, our method achieves significant performance improvements compared to single-image approaches.
We demonstrated the robustness of our method across both CT and MRI.

\subsubsection{Visual Comparisons}
Visualization of our method on the Synapse and ACDC datasets is shown in \figref{subfig:synapse} and \figref{subfig:acdc}.
For the Synapse dataset illustrated in \figref{subfig:synapse}, FCT failed to accurately segment SM and GB, while MERIT achieved precise segmentation of SM but struggled with GB.
In contrast, our method achieved accurate segmentation of both SM and GB.
Regarding the ACDC dataset shown in \figref{subfig:acdc}, while previous methods achieve comparable segmentation of the Myo and LV to the GT, they exhibit noticeable errors on the RV, including invasion into adjacent organs and misrecognition.
On the other hand, our method accurately segments across all three structures Myo, LV, and RV, performing as precisely as the GT.
We demonstrate the superiority of our method quantitatively and qualitatively.

\subsection{Ablation Study}
We have conducted an ablation study on the ACDC dataset to evaluate performance depending on the structure of the network and multi-scale boundary loss.
Also, we demonstrated the efficacy of our method through various input configurations and fusion strategies.

\subsubsection{DIFM and Multi-Scale Boundary Loss}
To investigate the effect of DIFM and $L_{bnd}$, we trained our model with various configurations as shown in \tableref{tab:difm}.
\new{
The baseline model, which simply fuses the original and enhanced images through concatenation without attributes of DIFM, achieved a 92.50\%.
This result demonstrates the importance of using cross-attention for effective information exchange between the two images, compared to simple concatenation.
However, we speculate that with sufficient training, the concatenation-based model may also learn to exchange information to some extent, which could explain the relatively slight performance gap.
}
Feature merging improved accuracy by 0.26\%p, demonstrating the importance of interacting with low and high features.
Adding global spatial attention increased accuracy by 0.2\%p, highlighting the need to emphasize key features after cross-attention.
Incorporating all components achieved 92.91\% accuracy, showcasing the value of integrating both feature types. $L_{bnd}$ further improved accuracy by 0.34\%p, emphasizing the importance of boundary information in segmentation tasks.
These experimental results demonstrate the superiority of our DIFM structure and $L_{bnd}$.

\begin{table}
\caption{Ablation study results on the network structure and loss function.
(a) Dual Cross-Attention, (b) Feature Merging, and (c) Global Spatial Attention.
}\label{tab:difm}
\centering
\resizebox{0.71\linewidth}{!}{
\begin{tabular}{ccc|ccc|c}
\Xhline{3\arrayrulewidth}
\multicolumn{3}{c|}{\textbf{DIFM Structure}} & \multicolumn{3}{c|}{\textbf{Loss}} & \textbf{Metrics} \\ 
(a) & (b) & (c) & $L_{CE}$ & $L_{Dice}$ & $L_{bnd}$ & Dice$\uparrow$ \\ \hline\hline
           &            &            & \new{\checkmark} & \new{\checkmark} &            & \new{92.50} \\ \hline
\checkmark &            &            & \checkmark & \checkmark &            & 92.55 \\ \hline
\checkmark & \checkmark &            & \checkmark & \checkmark &            & 92.81 \\ \hline
\checkmark &            & \checkmark & \checkmark & \checkmark &            & 92.75 \\ \hline
\checkmark & \checkmark & \checkmark & \checkmark & \checkmark &            & \underline{92.91} \\ \hline
\checkmark & \checkmark & \checkmark & \checkmark & \checkmark & \checkmark & \textbf{93.25} \\ \hline
\Xhline{3\arrayrulewidth} \\ 
\end{tabular}
}
\end{table}

\subsubsection{Various Inputs Configurations}
Experiment results on various input configurations are shown in \tableref{tab:various_inputs}.
\new{
The `only $I$' and `only $E$' using only the original or fuzzy-enhanced~\cite{FuzzyImageEnhancement} images with multi-head self-attention (MHSA).
Additionally, to evaluate the effectiveness of different enhancement techniques, we employ histogram equalization (HE), contrast-limited adaptive histogram equalization (CLAHE), and the deep learning-based Zero-DCE~\cite{li2021learning} method.
The results show that using two original images ($I+I$) resulted in increases of 0.04\%p and 0.24\%p compared to only $I$ and only $E$, respectively.
It can be attributed that, although the same features are used, the DIFM introduces additional computational.
However, it shows that $I+I$ yields slightly lower performance compared to using the original and enhanced images.
This suggests that enhanced images can provide complementary information not present in the original input.
When using images enhanced by Zero-DCE, the Dice score improved by 0.08\%p and 0.03\%p compared to those enhanced with HE and CLAHE, respectively.
However, the Dice score is 0.2\%p lower than that using the fuzzy~\cite{FuzzyImageEnhancement} method.
This can be attributed to the fact that Zero-DCE performs implicit enhancement through a general-purpose deep learning model, rather than explicitly highlighting medically relevant features such as object boundaries, edges, fine or shape details.
As a result, the enhanced images contain less task-specific information for organ segmentation than those produced by the fuzzy enhancement approach.
Nevertheless, incorporating enhanced images, regardless of the enhancement method, consistently outperforms using only a single original image or even two original images.
This highlights the importance of combining original and enhanced images to achieve optimal segmentation.
}

\begin{table}
\caption{Ablation study results on various input configurations.
$I$: original image, $E$: enhanced image.
}\label{tab:various_inputs}
\resizebox{0.76\linewidth}{!}{
\centering
\begin{tabular}{c|c|c|c|c}
\Xhline{3\arrayrulewidth}
Inputs & DICE$\uparrow$ & RV & Myo & LV \\ \hline\hline
only $I$ & 92.88 & \underline{91.74} & 90.64 & 96.25 \\
only $E$ & 92.66 & 91.09 & 90.64 & 96.26 \\
\new{$I +I$} & \new{92.92} & \new{91.38} & \new{90.96} & \new{96.42} \\
$I +$ HE & 92.97 & 91.58 & 91.03 & 96.31 \\ 
$I +$ CLAHE & 93.02 & 91.64 & \underline{91.07} & 96.35 \\
\new{$I +$ Zero-dce~\cite{li2021learning}} & \new{\underline{93.05}} & \new{91.64} & \new{91.06} & \new{\underline{96.46}} \\
Ours($I + E$~\cite{FuzzyImageEnhancement}) & \textbf{93.25} & \textbf{92.16} & \textbf{91.08} & \textbf{96.50}  \\
\Xhline{3\arrayrulewidth}
\end{tabular}
}
\end{table}

\begin{table}
\caption{Ablation study results on various fusion strategies.}\label{tab:fusion_strategy}
\centering
\resizebox{0.81\linewidth}{!}{
\begin{tabular}{c|c|c|c|c|c}
\Xhline{3\arrayrulewidth}
Strategy & Methods & Dice$\uparrow$ & RV & Myo & LV \\ \hline\hline
\multirow{3}{*}{Non-fusion} & SwinUNet~\cite{cao2022swinunet} & 91.32 & 90.10 & 88.47 & 95.40 \\
         & FCT~\cite{tragakis2023fully} & 91.36 & 89.19 & 89.28 & 95.60 \\
       & only $I$ & 92.88 & \underline{91.74} & 90.64 & 96.25 \\ \hline
\multirow{3}{*}{Input fusion} & SwinUNet & 91.15 & 89.09 & 88.91 & 95.44 \\
         & FCT & 91.15 & 88.84 & 89.07 & 95.55 \\
         & Ours$_{IF}$ & 92.63 & 90.90 & 90.75 & 96.23 \\ \hline
\multirow{4}{*}{Layer fusion} & SwinUNet$_{DIFM}$ & 91.93 & 90.50 & 89.63 & 95.66 \\
        & FCT$_{DIFM}$ & 91.83 & 89.49 & 90.10 & 95.91 \\
        & Ours$_{concat}$ & \underline{92.90} & 91.25 & \textbf{91.10} & \underline{96.35} \\
        & Ours & \textbf{93.25} & \textbf{92.16} & \underline{91.08} & \textbf{96.50} \\
\Xhline{3\arrayrulewidth}
\end{tabular}
}
\end{table}

\subsubsection{Various Fusion Strategies}

The performance comparison of various fusion strategies is presented in \tableref{tab:fusion_strategy}.
These experiments were conducted under the same conditions as \secref{sec:implementation}, using the loss function in \eqnref{equ:total_loss} to ensure a fair comparison.
The `Non-fusion' refers to using a single original image as input.
In Ours$_{IF}$, MHSA is employed instead of DIFM.
Ours$_{concat}$ represents the use of concatenation operations in place of DIFM.
SwinUNet$_{DIFM}$ and FCT$_{DIFM}$ indicate the application of DIFM for layer fusion.
Input fusion yields lower performance than non-fusion, likely due to irrelevant information from enhanced images.
Ours$_{concat}$ shows a slight improvement of 0.02\%p over only $I$, suggesting benefits from layer-level concatenation.
However, our method with DIFM outperforms Ours${concat}$ by 0.35\%p, highlighting DIFM's ability to selectively exploit relevant features.
Applying DIFM to SwinUNet and FCT improves performance by 0.61\%p and 0.47\%p, respectively.
Overall, these results emphasize the importance of selective feature fusion for effective medical image segmentation.

\new{
\subsection{Discussion}
\label{sec:discussion}
We discuss the computational cost of our method and its potential extension to 3D tasks.
This study is primarily focused on achieving high segmentation accuracy to support precise diagnostic applications.
The proposed method comprises a total of 323.31M parameters, including 178M for the encoder, 119.94M for the DIFM, and 25.27M for the MLP decoder, and the GFLOPs is 1860.09.
This result indicates that due to the separate processing of two images by two encoders, along with the use of the DIFM module to fully leverage complementary information.
While these costs are relatively high, the method consistently outperforms the various datasets, highlighting its effectiveness and suitability for high-precision medical image segmentation.
In future work, we aim to explore the development of more accurate yet lightweight architectures to reduce computational overhead while maintaining or improving performance.
}

\new{
In comparison to nnFormer~\cite{zhou2021nnformer}, our approach achieves superior performance on the ACDC and delivers competitive results on the Synapse dataset.
These results suggest that our proposed method, which integrates both the original and enhanced images, can rival 3D-based methods.
This implies that if our method were extended to directly process 3D volumes, it could potentially achieve even greater performance.
However, due to resource constraints, particularly the significantly increased computational and memory demands associated with 3D processing, we were unable to conduct experiments in the 3D domain within the scope of this study.
Processing full 3D volumes typically requires considerably more GPU memory, which exceeded our available resources.
One of the key advantages of DIFM is its modular architecture, which allows it to be flexibly integrated with different types of encoder representations (as shown in \tableref{tab:fusion_strategy}).
For instance, if the backbone model employs a ray-based representation, as in implicit neural representations or cone-beam CT reconstruction, DIFM can be adapted to include ray-aware attention mechanisms.
These mechanisms would enable the model to better capture the spatial and directional priors inherent in ray-sampled data.
For 3D convolution-based volumetric approaches, DIFM can be extended by reshaping feature maps along specific axes, for example, treating 3D volumes as sequences of 2D slices or applying axial flattening.
Such strategies, used in axial attention and tokenized volume modeling, are compatible with DIFM and represent a promising direction for future research.
}

\section{Conclusion}
\label{sec:conclusion}
In this paper, we propose DIFM for medical image segmentation, leveraging the advantages of both original and enhanced images.
Integrated into the network's skip connections, DIFM comprises three stages: feature merging, dual cross-attention for complementary integration, and global spatial attention for refining key attributes.
We also introduced the multi-scale boundary loss using gradient extraction to enhance accuracy at object boundaries.
As a result, our model has achieved a SOTA performance of 93.25\% on the ACDC dataset and a competitive result of 85.49\% on the Synapse dataset.

In future work, we will enhance our module to process the original image along with multiple enhanced versions simultaneously, using various image enhancement techniques.
Additionally, as mentioned in \secref{sec:discussion}, we plan to extend this approach further.
We believe that the proposed method of combining the original and enhanced image holds promise for advancing research in medical image segmentation.




\printcredits

\section*{Acknowledgments}
This work was supported in part by the IITP (Institute of Information \& Communications Technology Planning \& Evaluation)-ITRC (Information Technology Research Center) grant funded by the Korea Government (MSIT) ({IITP-2025-RS-2023-00260098}, 50\%) and in part by the National Research Foundation of Korea (NRF) \href{http://dx.doi.org/10.13039/501100003725}{South Korea} grant funded by the Korea Government (MSIT)({RS-2023-00217689}, 50\%).

\section*{Data availability}
We have utilized public data only.

\bibliographystyle{elsarticle-num}

\bibliography{template}

\clearpage
\twocolumn[
\begin{center}
    {\LARGE \bfseries Supplementary: Dual Interaction Network with Cross-Image Attention for Medical Image Segmentation \par}
\end{center}
\vspace{2em}
]

\section{Sup: Introduction}
In this supplementary material, we present additional ablation studies, image enhancement methods, the structure of the encoder and decoder, the gradient extractor of multi-scale boundary loss, limitation,s and further visual comparison.
The details of these supplementary components are described in the following sections.

\section{Additional Ablation Studies}
\label{sec:additional}
We have conducted additional ablation studies on the ACDC and Synapse datasets.
We demonstrated the efficacy of our method through a number of DIFMs.

We evaluate the cross-attention method and demonstrate the superiority of our multi-scale boundary loss function $L_{bnd}$.
Moreover, we compare performance across three scenarios: using only original images (only $I$), using only enhanced images (only $E$), and our proposed method.
Additionally, we conduct decoder constraints and other types of datasets.
Note that all methods were trained under the same experimental environment as our proposed method, to ensure fair comparisons.

\subsection{Number of DIFM}

\begin{table}[H]
\caption{Ablation study results on the number of DIFM.}\label{tab:number_of_DIFM}
\centering
\resizebox{0.70\linewidth}{!}{
\begin{tabular}{c|c|c|c|c}
\Xhline{3\arrayrulewidth}
\# of DIFM & Dice$\uparrow$ & RV & Myo & LV \\ \hline\hline
1 & 86.86 & 81.21 & 85.88 & 93.48 \\
2 & 91.31 & 89.17 & 89.44 & 95.33 \\ 
3 & \underline{92.78} & \underline{91.36} & \underline{90.79} & \underline{96.18} \\
4 & \textbf{93.25} & \textbf{92.16} & \textbf{91.08} & \textbf{96.50} \\
\Xhline{3\arrayrulewidth}
\multicolumn{5}{r}{\textbf{Bold}: The best, \underline{Underline}: The second-best}
\end{tabular}
}
\end{table}

\Tabref{tab:number_of_DIFM} presents the performance evaluation based on the number of DIFMs utilized.
In our experiments, we systematically removed DIFMs by first excluding the feature map with the smallest resolution.
Results indicate a positive correlation between the number of DIFMs and accuracy.
Notably, using four DIFMs yielded a 0.47\%p increase in accuracy compared to three, underscoring the importance of inter-image feature fusion.
This experiment demonstrates that segmentation accuracy generally improves with an increasing number of DIFMs utilizing various features.

\subsection{Comparison with cross-attention}

The experiments on cross-attention are shown in \tableref{tab:uni_direction}.
The method that only uses the original image as Q and the enhanced image as K and V is denoted as $\tilde{f}_{IE}$.
In contrast, the method that only uses the enhanced image as Q and the original image as K and V is denoted as $\tilde{f}_{EI}$.
The results indicate that the dual cross-attention method outperforms the $\tilde{f}_{IE}$ and $\tilde{f}_{EI}$ methods by 0.15\%p and 0.09\%p, respectively.
This performance gap arises because the uni-directional methods exchange information in only one direction, either from $E$ to $I$ or from $I$ to $E$, thereby restricting the depth of contextual comprehension.
In contrast, dual cross-attention allows for bidirectional information exchange between the $I$ and $E$ features, enabling a deeper understanding of complex relations.
Therefore, dual cross-attention is more effective in enhancing the richness of contextual understanding compared to uni-directional cross-attention.

\begin{table}
\caption{Ablation study results on the cross-attention.
$I$: original image, $E$: enhanced image.
}
\label{tab:uni_direction}
\resizebox{0.70\linewidth}{!}{
\centering
\begin{tabular}{c|c|c|c|c}
\Xhline{3\arrayrulewidth}
Methods & DICE$\uparrow$ & RV & Myo & LV \\ \hline\hline
$\tilde{f}_{IE}$ & 93.10 & 91.82 & \textbf{91.13} & \underline{96.37} \\
$\tilde{f}_{EI}$ & \underline{93.14} & \textbf{92.21} & 90.89 & 96.30 \\
Ours & \textbf{93.25} & \underline{92.16} & \underline{91.08} & \textbf{96.50} \\
\Xhline{3\arrayrulewidth}
\multicolumn{5}{r}{\textbf{Bold}: The best, \underline{Underline}: The second-best}
\end{tabular}
}
\end{table}

\subsection{Comparison with Multi-Scale Boundary Loss}

The performance comparison for $L_{bnd}$ is shown in \tableref{tab:loss_bnd}.
When utilizing the $L_{bnd}$, the performance for SwinUNet, MERIT, FCT, and the proposed method increased by 0.28\%p, 0.12\%p, 0.04\%p, and 0.34\%p, respectively, compared to those without $L_{bnd}$.
This improvement indicates that $L_{bnd}$ enables the optimization of the boundary differences between prediction and GT, unlike conventional loss functions such as CE and Dice, which compute the probability differences and similarities, respectively.
Therefore, effectively refining boundary delineation is crucial for improving segmentation tasks.

\begin{table}
\caption{Ablation study results on the multi-scale boundary loss.}\label{tab:loss_bnd}
\centering
\resizebox{0.85\linewidth}{!}{
\begin{tabular}{c|c|c|c|c|c}
\Xhline{3\arrayrulewidth}
Loss Function & Methods & DICE$\uparrow$ & RV & Myo & LV \\ \hline\hline
\multirow{4}{*}{$L_{CE} + L_{Dice}$} & SwinUNet~\cite{cao2022swinunet} & 91.04 & 89.44 & 88.29 & 95.38 \\
       & MERIT~\cite{rahman2023multi} & 91.18 & 89.35 & 88.76 & 95.43 \\
       & FCT~\cite{tragakis2023fully} & 91.32 & 89.37 & 89.06 & 95.51 \\
       & Ours & \underline{92.91} & \underline{91.71} & \underline{90.85} & \underline{96.16} \\ \hline
\multirow{4}{*}{$L_{CE} + L_{Dice} + L_{bnd}$} & SwinUNet~\cite{cao2022swinunet} & 91.32 & 90.10 & 88.47 & 95.40 \\
             & MERIT~\cite{rahman2023multi} & 91.30 & 89.35 & 88.90 & 95.64 \\
             & FCT~\cite{tragakis2023fully} & 91.36 & 89.19 & 89.28 & 95.60 \\
             & Ours & \textbf{93.25} & \textbf{92.16} & \textbf{91.08} & \textbf{96.50} \\ 
\Xhline{3\arrayrulewidth}
\multicolumn{6}{r}{\textbf{Bold}: The best, \underline{Underline}: The second-best}
\end{tabular}
}
\end{table}

\subsection{Comparison with Single-image-based and DIFM}
\label{sec:single_multi}
The performance comparison between single-image-based and using a dual interactive fusion module (DIFM) is shown in \tableref{tab:synapse_single} and \tableref{tab:acdc_single}.
The `only $I$' and `only $E$' employ multi-head self-attention instead of DIFM.
Our approach achieved superior performance on the ACDC and Synapse datasets compared to using only $I$ and $E$.
This improvement is attributed to effectively and fully leveraging the advantages of both original and enhanced images through DIFM.
Conversely, on the Synapse, there was a slight decrease in the Dice scores for the Aorta, GB, and Liver.
This can be attributed to organ-specific complexities.
These organs have unique anatomical characteristics, such as complex morphologies and variable shapes, which may interact differently with the proposed method due to significant inter-patient variability.
Despite these challenges, the proposed method significantly enhances overall segmentation performance.

\begin{table*}
\caption{Ablation study results between single-image-based and DIFM on the Synapse dataset.}
\label{tab:synapse_single}
\centering
\resizebox{0.9\textwidth}{!}{
\begin{tabular}{c|c|c|c|c|c|c|c|c|c|c}
\Xhline{3\arrayrulewidth}
Methods & Dice$\uparrow$ & HD95$\downarrow$ & Aorta & GB & KL & KR & Liver & PC & SP & SM \\ \hline\hline
only $I$ & 83.25 & 27.32 & \textbf{90.12} & \textbf{75.89} & 84.83 & 83.45 & \textbf{95.26} & 71.49 & 87.21 & 77.73 \\
only $E$~\cite{FuzzyImageEnhancement} & \underline{83.26} & \underline{26.33} & \underline{89.30} & 64.83 & \underline{86.47} & \underline{83.92} & 93.27 & \underline{72.52} & \underline{91.20} & \underline{84.64} \\ \hline
Ours & \textbf{85.49} & \textbf{10.74} & 89.12 & \underline{74.32} & \textbf{87.47} & \textbf{85.85} & \underline{94.71} & \textbf{73.82} & \textbf{92.30} & \textbf{86.31} \\
\Xhline{3\arrayrulewidth}
\multicolumn{11}{r}{\textbf{Bold}: The best, \underline{Underline}: The second-best}
\end{tabular}
}
\end{table*}

\begin{table}
    \centering
    \caption{Ablation study results between single-image-based and DIFM on the ACDC dataset.}
    \label{tab:acdc_single}
    \resizebox{0.9\linewidth}{!}{
    \begin{tabular}{c|c|c|c|c}
    \Xhline{3\arrayrulewidth}
    Methods & Dice$\uparrow$ & RV & Myo & LV \\ \hline\hline
    only $I$ & \underline{92.88} & \underline{91.74} & \underline{90.64} & 96.25 \\
    only $E$~\cite{FuzzyImageEnhancement} & 92.66 & 91.09 & \underline{90.64} & \underline{96.26} \\ \hline
    Ours & \textbf{93.25} & \textbf{92.16} & \textbf{91.08} & \textbf{96.50} \\ 
    \Xhline{3\arrayrulewidth}
    
    \multicolumn{5}{r}{\textbf{Bold}: The best, \underline{Underline}: The second-best}
    \end{tabular}
    }
\end{table}

\subsection{Decoder Constraints and Generalization}

\begin{table}
\caption{Ablation study results on the ACDC test set with decoder structures.}
\label{tab:ACDC_EMCAD}
\centering
\resizebox{0.90\linewidth}{!}{
\begin{tabular}{c|c|c|c|c}
\Xhline{3\arrayrulewidth}
Decoder & Dice$\uparrow$ & RV & Myo & LV \\ \hline\hline
EMCAD~\cite{rahman2024emcad} & 93.02 & 91.77 & 90.91 & 96.39 \\ 
SegFormer~\cite{xie2021segformer} & \textbf{93.25} & \textbf{92.16} & \textbf{91.08} & \textbf{96.50} \\ 
\Xhline{3\arrayrulewidth}
\end{tabular}
}
\end{table}

\begin{table*}
\caption{Ablation study results on the Synapse test set with decoder structures.}
\label{tab:synapse_EMCAD}
\centering
\resizebox{0.80\textwidth}{!}{
\begin{tabular}{c|c|c|c|c|c|c|c|c|c|c}
\Xhline{3\arrayrulewidth}
Decoder & Dice$\uparrow$ & HD95$\downarrow$ & Aorta & GB & KL & KR & Liver & PC & SP & SM \\ \hline\hline
EMCAD~\cite{rahman2024emcad} & 81.67 & 13.78 & \textbf{89.34} & 71.11 & \textbf{88.81} & 82.41 & \textbf{95.07} & 63.36 & 87.75 & 76.13 \\
SegFormer~\cite{xie2021segformer} & \textbf{85.49} & \textbf{10.74} & 89.12 & \textbf{74.32} & 87.47 & \textbf{85.85} & 94.71 & \textbf{73.82} & \textbf{92.30} & \textbf{86.31} \\
\Xhline{3\arrayrulewidth}
\end{tabular}
}
\end{table*}

%
Tables \tabref{tab:ACDC_EMCAD} and \tabref{tab:synapse_EMCAD} represent experiment results using EMCAD~\cite{rahman2024emcad} on the ACDC and Synapse.
Replacing the SegFormer~\cite{xie2021segformer} multi-layer perceptron (MLP) decoder with EMCAD results in a 0.23\%p decrease in Dice score on the ACDC.
Likewise, on Synapse, the results dropped by 3.82\%p.
Although EMCAD efficiently uses convolutional operations, applied gating mechanisms, and depth-wise convolution, its $n \times n$ kernel size inherently limits the receptive field to local regions, making it less effective at capturing long-range dependencies.
In contrast, the SegFormer MLP decoder flattens the multi-scale features extracted from the transformer encoder and processes them globally, thereby incorporating rich contextual information across the entire image.
%
Moreover, since we utilize ConvNext base~\cite{liu2022convnet}, an architecture that imitates transformers through design choices such as 7$\times$7 kernels and layer normalization, it provides a large receptive field and has been shown to outperform previous methods.
Given this, the MLP decoder is better suited for handling such a feature map compared to EMCAD.
However, based on an input resolution of 224 $\times$ 224, the MLP decoder requires 25.27M parameters, whereas EMCAD requires only 20.72M.
In this regard, EMCAD can be considered more parameter-efficient than the MLP decoder.
Nevertheless, we adopt the MLP decoder to achieve higher accuracy, which is particularly critical in the medical domain.

\subsection{Additional Dataset Scope}

To demonstrate the superiority of our method, we additionally conduct experiments on the ClinicDB~\cite{Clinic} and ISIC18~\cite{ISIC18} datasets.
ClinicDB comprises 612 colonoscopic images containing polyps, while ISIC18 consists of 3,594 dermoscopic images of skin lesions.
Both datasets are used for binary segmentation tasks and are split into training, validation, and test sets with a 80:10:10.
For implementation, the input images are resized to 224 $\times$ 224, and random rotation as well as horizontal and vertical flip augmentations are applied.
The hyperparameter settings follow those used in the ACDC and Synapse experiments.
The evaluation metric is the Dice score.
Please kindly note that both the ACDC and Synapse datasets used in the main paper are originally 3D volumetric datasets, which were processed into 2D slices for all experiments.
Therefore, the additional experiments were conducted on 2D image datasets.

Table \tabref{tab:div_data} provides quantitative results on the additional two datasets.
Our method achieved a superior performance compared to previous works.
Specifically, it outperformed EMCAD by 1.79\%p and 2.89\%p Dice scores on the ClinicDB and ISIC18 datasets, respectively.
These results indicated that our proposed method effectively leverages the advantage of the original and enhanced images using DIFM and refines the object boundary by multi-scale boundary loss $L_{bnd}$.
Therefore, we demonstrate the robustness of our method not only for CT and MRI but also for polyps and skin.

\begin{table}
\caption{Quantitative results on the various binary segmentation datasets with previous methods.}
\label{tab:div_data}
\resizebox{0.85\linewidth}{!}{
\centering
\begin{tabular}{c|c|c}
\Xhline{3\arrayrulewidth}
Methods & Clinic & ISIC18 \\ \hline\hline
UNet~\cite{ronneberger2015unet} & 92.11 & 86.67 \\
UNet$++$~\cite{zhou2018unet++} & 92.17 & 87.46 \\
AttnUNet~\cite{oktay2018attentionunet} & 92.20 & 87.05 \\
DeepLabv3+~\cite{chen2017deeplab} & 93.24 & 88.64 \\
PraNet~\cite{fan2020pranet} & 91.71 & 88.56 \\
CaraNet~\cite{lou2022caranet} & 94.08 & 90.18 \\
UACANet-L~\cite{kim2021uacanet} & 94.16 & 89.76 \\
SSFormer-L~\cite{wang2022stepwise} & 94.18 & 90.25 \\
PolypPVT~\cite{dong2021polyp} & 94.13 & 90.36 \\
TransUNet~\cite{chen2021transunet} & 93.90 & 89.16 \\
SwinUNet~\cite{cao2022swinunet} & 92.42 & 89.26 \\
TransFuse~\cite{zhang2021transfuse} & 93.62 & 89.62 \\
UNeXt~\cite{valanarasu2022unext} & 90.20 & 87.78 \\
PVT-CASCADE~\cite{rahman2023medical} & 94.53 & 90.41 \\
PVT-EMCAD-B0~\cite{rahman2024emcad} & 94.60 & 90.70 \\
PVT-EMCAD-B2~\cite{rahman2024emcad} & 95.21 & 90.96 \\ \hline
Ours & \textbf{97.00} & \textbf{93.85} \\
\Xhline{3\arrayrulewidth}
\end{tabular}
}
\end{table}

\section{Image Enhancement}
\label{sec:enhancement}

\begin{figure}
    \centering
    \renewcommand{\arraystretch}{0.2}
    \begin{subtable}[b]{\linewidth}
        \centering
        \begin{tabular}{@{}c@{\hskip 0.003\linewidth}c@{\hskip 0.003\linewidth}c@{\hskip 0.003\linewidth}c@{\hskip 0.003\linewidth}}
            \includegraphics[width=0.224\linewidth]{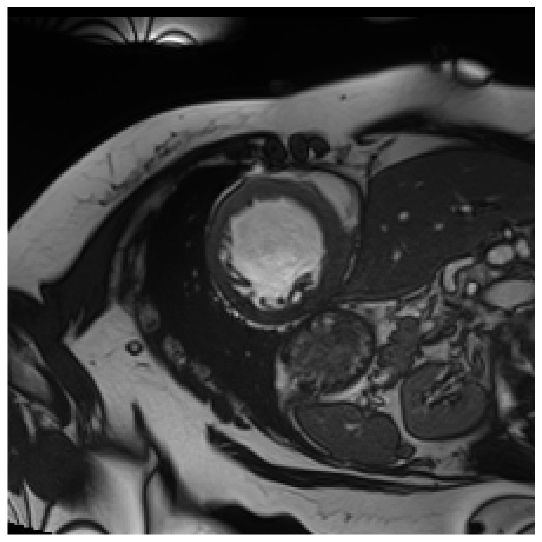} & 
            \includegraphics[width=0.224\linewidth]{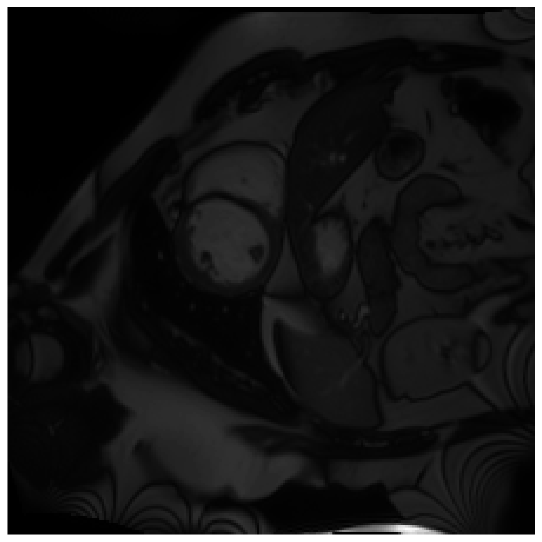} & 
            \includegraphics[width=0.224\linewidth]{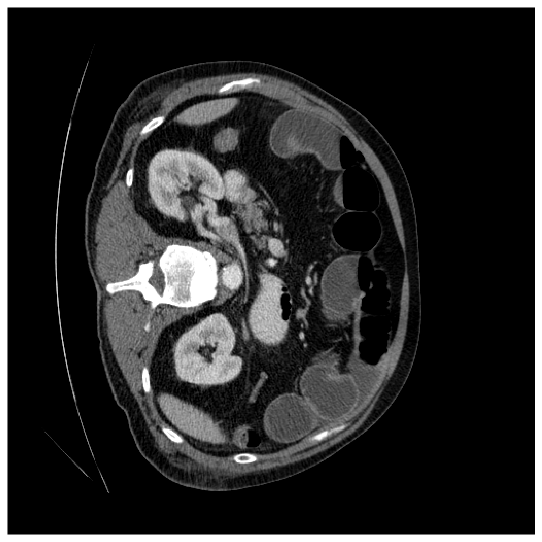} &
            \includegraphics[width=0.224\linewidth]{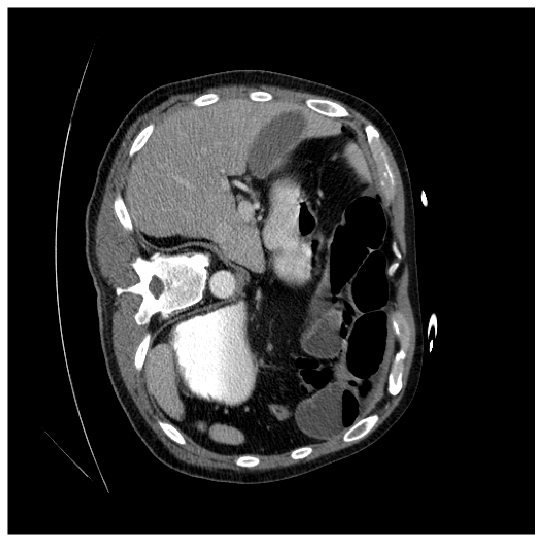} \\
        \end{tabular}
        \caption{}
        \label{subfig:original_image}
    \end{subtable}
    \begin{subtable}[b]{\linewidth}
        \centering
        \begin{tabular}{@{}c@{\hskip 0.003\linewidth}c@{\hskip 0.003\linewidth}c@{\hskip 0.003\linewidth}c@{\hskip 0.003\linewidth}}
            \includegraphics[width=0.224\linewidth]{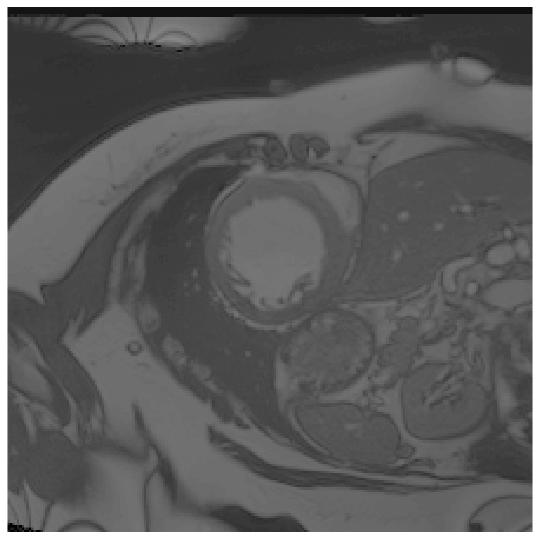} & 
            \includegraphics[width=0.224\linewidth]{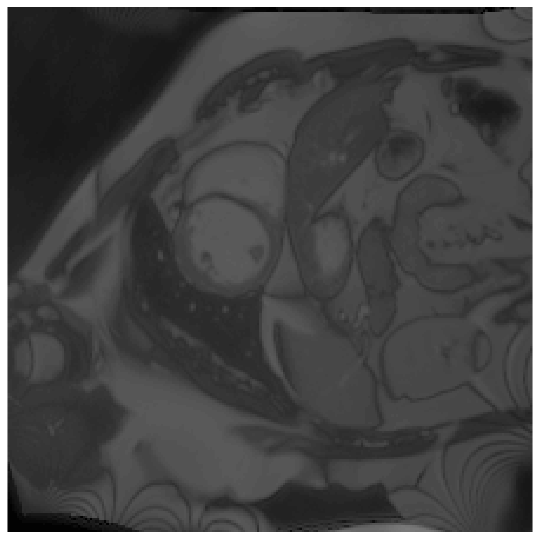} & 
            \includegraphics[width=0.224\linewidth]{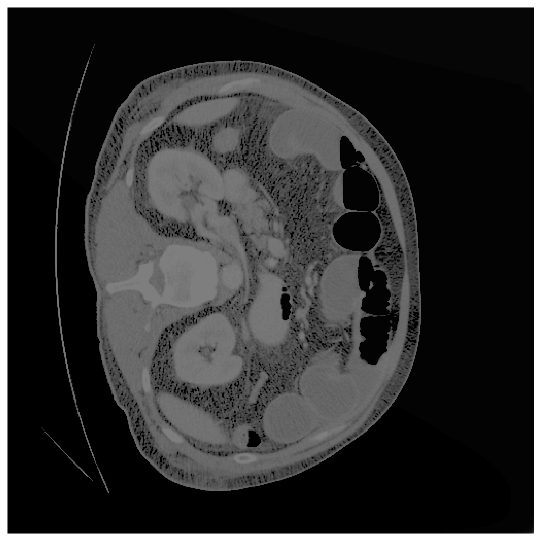} &
            \includegraphics[width=0.224\linewidth]{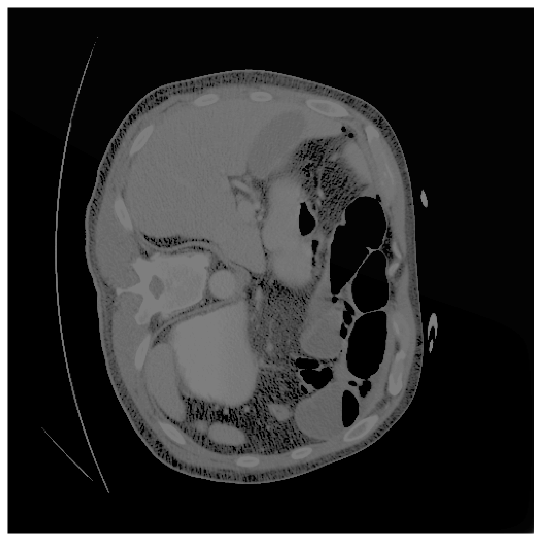} \\
        \end{tabular}
        \caption{}
        \label{subfig:fuzzy_image}
    \end{subtable}
    \label{fig:ori_fuzzy}
    \caption{
    Visual comparisons of enhancement results (a) Original image and (b) Fuzzy image enhancement~\cite {FuzzyImageEnhancement} image.
    }
\end{figure}

In medical imaging, preprocessing through enhancement techniques is necessary for effective noise handling~\cite{perumal2018preprocessing}.
In this paper, we use fuzzy image enhancement~\cite{FuzzyImageEnhancement}, which has proven effective in various works~\cite{gupta2021instaconvnet,iqball2022covid}, for enhancing the original image.
The fuzzy image enhancement utilizes fuzzy rules based on local fuzzy mean and variance to enhance image quality.
The transformation function for each fuzzy window $W_{ij}$ of the original image is formulated as follows:
\begin{equation}
    \begin{split}
        \label{equ:F_enhance_1}
        & \psi_{ij}(f) = \lambda_{ij} \cdot (f + \tau_{ij}), \\
        & \textrm{where} \quad \lambda_{ij} = \frac{\sigma_u}{\sigma_\varphi(f, W_{ij})}, \; \tau_{ij} = -\mu_\varphi(f, W_{ij}),
    \end{split}
\end{equation}
and the final enhanced image $E$ is formulated as follows:
\begin{equation}
    \label{equ:F_enhance_3}
    E = \sum_{i, j} I_{ij} \cdot \psi_{ij}(f),
\end{equation}
where $f$ and $I_{ij}$ denote the fuzzy window patch extracted from the original image and the pixel at coordinates $(i, j)$ in the original image $I$, respectively.
$\lambda_{ij}$ is the scaling factor defined as the ratio of the desired uniform variance $\sigma_u$ to the fuzzy variance $\sigma_\varphi(f, W_{ij})$, and $\tau_{ij}$ is the translation factor, where $\mu_\varphi$ represents the fuzzy mean.
In the logarithmic model, the operator $\cdot$ represents scalar multiplication.
The membership function calculator, $\psi_{ij}$, is weighted based on how each transformed fuzzy window contributes to the final enhanced image.
As a result, the enhanced image exhibits improved quality and a distinct feature distribution compared to the original image.
The original and enhanced images are shown in \figref{subfig:original_image} and \figref{subfig:fuzzy_image}, respectively. 
As shown in \figref{subfig:fuzzy_image}, the image quality is enhanced by \eqnref{equ:F_enhance_3}, emphasizing the organ's attributes, including shape and intensity.
Consequently, previously subtle features in \figref{subfig:original_image}, become more pronounced in \figref{subfig:fuzzy_image}.
When these two images are processed through their respective encoders, they generate feature maps characterized by diverse feature distributions and unique information.
In other words, the boundary information is better preserved in the original image, while the enhanced image highlights the object’s details.
Thus, cross-learning the information from both images leads to improved performance.
Please kindly refer to \cite{FuzzyImageEnhancement} for more details.

\section{Encoder and Decoder}
\begin{table}
\caption{ConvNext-base~\cite{liu2022convnet} architecture.}\label{tab:convnext}
\centering
\resizebox{0.85\linewidth}{!}{
\begin{tabular}{c|c|c}
\Xhline{3\arrayrulewidth}
Feature & Output Size & ConvNeXt-base~\cite{liu2022convnet} \\ \hline \hline
$\hat{f}_{edge}$ & 224$\times$224 & 3$\times$3, 128, stride 1 \\ \hline
\multirow{3}{*}{}& & \\
$\hat{f}_{shape}$ & 224$\times$224 & $ \left[ \begin{array}{cc} \mathrm{d}7\times7, 128 \\ 1\times1, 512 \\ 1\times1, 128 \end{array} \right] \times3 $ \\
 & & \\ \hline
$\hat{f}_{edge}$ & 112$\times$112 & 2$\times$2, 256, stride 2 \\ \hline
\multirow{3}{*}{}& & \\
$\hat{f}_{shape}$ & 112$\times$112 & $ \left[ \begin{array}{cc} \mathrm{d}7\times7, 256 \\ 1\times1, 1024 \\ 1\times1, 256 \end{array} \right] \times3 $ \\
 & & \\ \hline
$\hat{f}_{edge}$ & 56$\times$56 & 2$\times$2, 512, stride 2 \\ \hline
\multirow{3}{*}{}& & \\
$\hat{f}_{shape}$ & 56$\times$56 & $ \left[ \begin{array}{cc} \mathrm{d}7\times7, 512 \\ 1\times1, 2048 \\ 1\times1, 512 \end{array} \right] \times9 $ \\
 & & \\ \hline
$\hat{f}_{edge}$ & 28$\times$28 & 2$\times$2, 1024, stride 2 \\ \hline
\multirow{3}{*}{}& & \\
$\hat{f}_{shape}$ & 28$\times$28 & $ \left[ \begin{array}{cc} \mathrm{d}7\times7, 1024 \\ 1\times1, 4096 \\ 1\times1, 1024 \end{array} \right] \times3 $ \\
 & & \\ \hline
\Xhline{3\arrayrulewidth}
\end{tabular}
}
\end{table}

\subsection{Encoder}
\label{sec:encoder}
In this paper, we employ ConvNext-base~\cite{liu2022convnet} as our encoder to extract diverse structural attributes from the original and enhanced image.
ConvNext is designed to compete with vision transformers, incorporating various architectural enhancements, such as efficient block and layer normalization, which improve both performance and efficiency in image recognition tasks.
ConvNext-base architecture is detailed in \tableref{tab:convnext}.
The encoder architecture consists of three main components: a stem, convolutional blocks (conv-blocks), and convolutional pooling layers.
The stem initially expands the input image channels to 128, facilitating the extraction of $\hat{f}_{edge}$.
Unlike the conventional ConvNext stem, which uses a 4$\times$4 kernel size with a stride of 4, our implementation employs a 3$\times$3 kernel size with a stride of 1 to preserve full-size features.
The conv-blocks utilize an inverted bottleneck structure, positioning depth-wise convolutions, which has 7$\times$7 kernel size, at the front of each block to increase the receptive field, mimicking transformer-like characteristics.
These blocks incorporate layer normalization and GELU activation function~\cite{hendrycks2016gaussian}.
Conv-block is formulated as follows:
\begin{equation}
    y = \mathtt{conv}(\mathtt{GELU}(\mathtt{conv}(\mathtt{LN}(\mathrm{d}\mathtt{conv}(x, k)), k)), k) + x,
\end{equation}
where $x$ and $y$ denote the input and output feature map, respectively.
The terms $\mathrm{d}\mathtt{conv}(\cdot, k)$ and $\mathtt{conv}(\cdot, k)$ represent depth-wise convolution and standard convolution operations, respectively, with $k$ indicating the kernel size.
$\mathtt{LN}$ and $\mathtt{GELU}$ refer to layer normalization and the GELU activation function.
These conv-blocks effectively extract $\hat{f}_{shape}$.
The convolutional pooling layers, implemented with a 2$\times$2 kernel size with a stride of 2, halve the feature size.
We implement two types of pooling layers: one for extracting $\hat{f}_{edge}$ and another for reducing feature size before passing to subsequent conv-blocks.
This encoder design efficiently captures both $\hat{f}_{edge}$ and $\hat{f}_{shape}$, resulting in the extraction of feature maps with abundant information.

\subsection{Decoder}
\label{sec:decoder}
\begin{figure}
\centering
\includegraphics[width=0.5\linewidth]{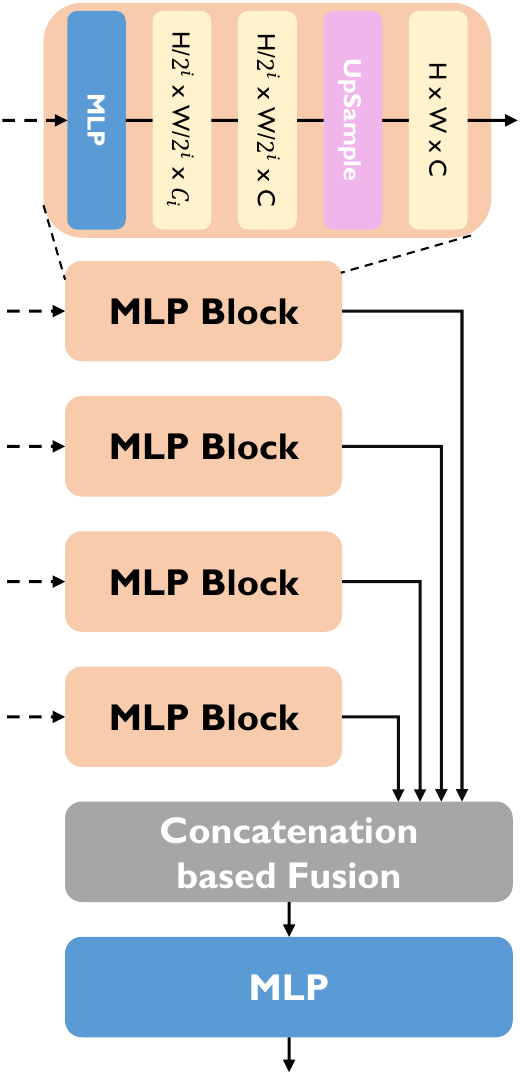}
\caption{Decoder~\cite{xie2021segformer} architecture.}\label{fig:mlp_decoder}
\end{figure}
In this paper, we employ SegFormer's multi-layer perception (MLP) decoder~\cite{xie2021segformer}, as illustrated in \figref{fig:mlp_decoder}.
It offers a more lightweight alternative to convolution neural network (CNN) based decoders while effectively analyzing the large receptive field of the encoder.
This MLP decoder comprises three primary components: MLP blocks, a concatenation-based fusion mechanism, and an MLP layer.
The MLP blocks serve to standardize the feature maps, which vary in size and channel dimensions, to uniform configuration.
Specifically, it transforms all feature maps to a channel dimension of 768 and a feature map size of 224$\times$224, using bilinear interpolation for resizing.
The concatenation-based fusion mechanism integrates the standardized feature maps through a concatenation operation followed by a conv-block.
This conv-block consists of a convolution operation, batch normalization, and ReLU activation function, and is formulated as follows:
\begin{equation}
    y = \mathtt{ReLU}(\mathtt{BN}(\mathtt{conv}(x_{1\sim4}, 3))),
\end{equation}
where $y$ represents the output, $x_{1\sim4}$ denotes the feature maps concatenated along the channel axis.
The role of $\mathtt{conv}$ here is to reduce the feature space by a quarter of the channel dimension.
$\mathtt{BN}$ and $\mathtt{ReLU}$ refer to batch normalization and ReLU activation function.
The final component is an MLP layer dedicated to predicting the segmentation mask, implemented as a convolution operation with a 1$\times$1 kernel size.
This decoder design efficiently and accurately generates the segmentation mask.

\section{Multi-Scale Boundary Loss}
\label{sec:loss}
Ours $L_{bnd}$ implements a gradient extractor to efficiently compute gradients such as boundaries and intensities between adjacent pixels.
The gradient extractor employs the Sobel filter to quickly and accurately identify boundaries.
The Sobel filter, through convolution operations, extracts gradients and is designed to respond maximally to edges running horizontally and vertically relative to the pixel grid.
Compared to the multi-stage-based Canny edge detector, the Sobel filter is computationally simpler and more efficient.
Additionally, the Sobel filter is more robust to noise than the Laplacian filter due to its use of first-order derivatives and the averaging effect inherent in its kernel design.
This robustness allows it to compute gradients more accurately and stably.
Consequently, we implement the gradient extractor based on the Sobel filter, which is defined as follows:
\begin{equation}
    G_{H}= 
    \begin{bmatrix}
    -1 & -2 & -1 \\
    0 & 0 & 0 \\
    1 & 2 & 1
    \end{bmatrix}
    , \quad
    G_{V}=
    \begin{bmatrix}
    -1 & 0 & 1 \\
    -2 & 0 & 2 \\
    -1 & 0 & 1
    \end{bmatrix}
    ,
\end{equation}
where $G_{H}$ and $G_{V}$ denote the horizontal gradient extractor and vertical gradient extractor, respectively.

We opt for the L1 norm over the L2 norm, to quantify the discrepancy between prediction and GT boundaries.
The L2 norm, which calculates the squared difference between predicted and GT values, is effective in reducing overall error and often leads to smoother images.
However, it tends to blur sharp edges and fine details because it penalizes larger errors more heavily.
This characteristic can lead to the smoothing of sharpness, which is crucial for maintaining image clarity and sharpness.
In contrast, the L1 norm calculates the absolute differences between predicted and GT values.
This approach demonstrates robustness to outliers and superior preservation of edges and fine details.
By penalizing large errors linearly, the L1 norm encourages less blurring and more effectively retains sharp features.
This property makes it particularly suitable for our application, where preserving the integrity of boundary details is paramount.
Therefore, we employ the L1 norm in our methodology, as it facilitates more accurate delineation of boundaries.

\section{Computational Cost}
\label{sec:limitatoin}
The computational cost of the proposed method is significant.
This study focuses on achieving high accuracy to support precise diagnostic assistance, which necessitated the use of two heavy encoders, ConvNext-base~\cite{liu2022convnet}, resulting in a total of 323.31M parameters, 178M for the encoders, 119.94M for DIFM, and 25.67M for the MLP decoder, as shown in \tabref{tab:cost}.
Future research should explore the development of more accurate and lightweight encoders and decoders to reduce the computational cost.
Despite this limitation, the study offers valuable insights, laying the groundwork for further research into medical image segmentation.

\begin{table}
\caption{Computational cost comparison with previous methods.}
\label{tab:cost}
\resizebox{0.85\linewidth}{!}{
\centering
\begin{tabular}{c|c|c}
\Xhline{3\arrayrulewidth}
Methods & \#Params (M) & GFLOPs \\ \hline\hline
UNet~\cite{ronneberger2015unet} & 34.53 & 65.53 \\
UNet$++$~\cite{zhou2018unet++} & 9.16 & 34.65 \\
AttnUNet~\cite{oktay2018attentionunet} & 34.88 & 66.64 \\
DeepLabv3+~\cite{chen2017deeplab} & 39.76 & 14.92 \\
PraNet~\cite{fan2020pranet} & 32.55 & 6.93 \\
CaraNet~\cite{lou2022caranet} & 46.64 & 11.48 \\
UACANet-L~\cite{kim2021uacanet} & 69.16 & 31.51 \\
SSFormer-L~\cite{wang2022stepwise} & 66.22 & 17.28 \\
PolypPVT~\cite{dong2021polyp} & 25.11 & 5.30 \\
TransUNet~\cite{chen2021transunet} & 105.32 & 38.52 \\
SwinUNet~\cite{cao2022swinunet} & 27.17 & 6.2 \\
TransFuse~\cite{zhang2021transfuse} & 143.74 & 82.71 \\
UNeXt~\cite{valanarasu2022unext} & 1.47 & 0.57 \\
PVT-CASCADE~\cite{rahman2023medical} & 34.12 & 7.62 \\
PVT-EMCAD-B0~\cite{rahman2024emcad} & 3.92 & 0.84 \\
PVT-EMCAD-B2~\cite{rahman2024emcad} & 26.76 & 5.6 \\ \hline
Ours & 323.31 & 1860.09 \\
\Xhline{3\arrayrulewidth}
\end{tabular}
}
\end{table}

\section{Additional Visual Comparisons}
\label{sec:visual}
\begin{figure*}
    \centering
    \renewcommand{\arraystretch}{0.2}
    \begin{tabular}{@{}c@{\hskip 0.003\linewidth}c@{\hskip 0.003\linewidth}c@{\hskip 0.003\linewidth}c@{\hskip 0.003\linewidth}c@{\hskip 0.003\linewidth}c@{\hskip 0.003\linewidth}c@{\hskip 0.003\linewidth}c@{\hskip 0.003\linewidth}}
    \multicolumn{4}{l}{\includegraphics[width=0.484\linewidth]{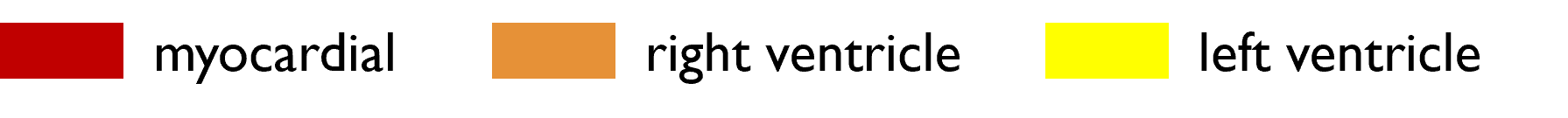}} \\
        \includegraphics[width=0.121\linewidth]{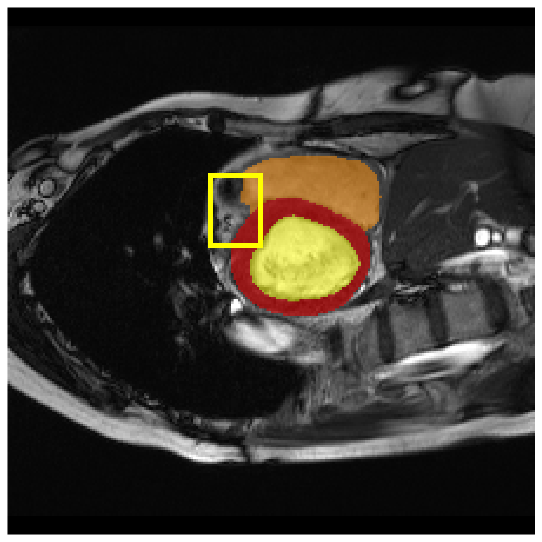} & 
        \includegraphics[width=0.121\linewidth]{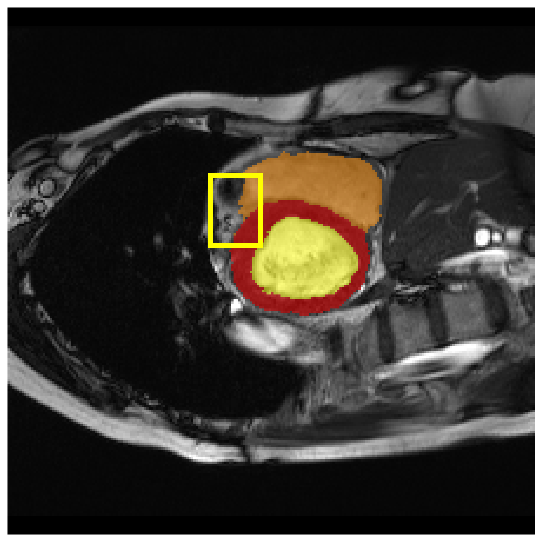} & 
        \includegraphics[width=0.121\linewidth]{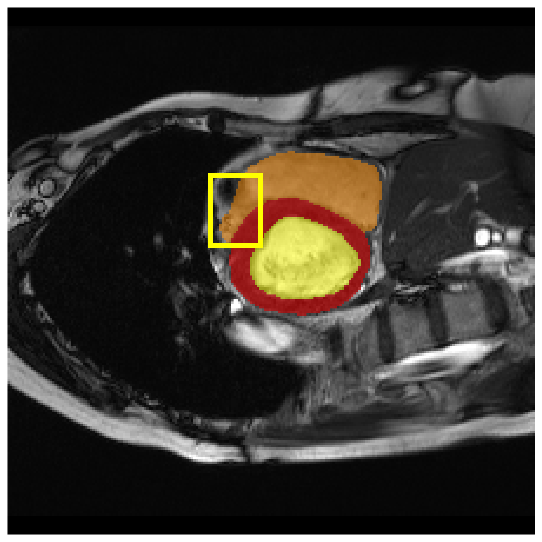} &
        \includegraphics[width=0.121\linewidth]{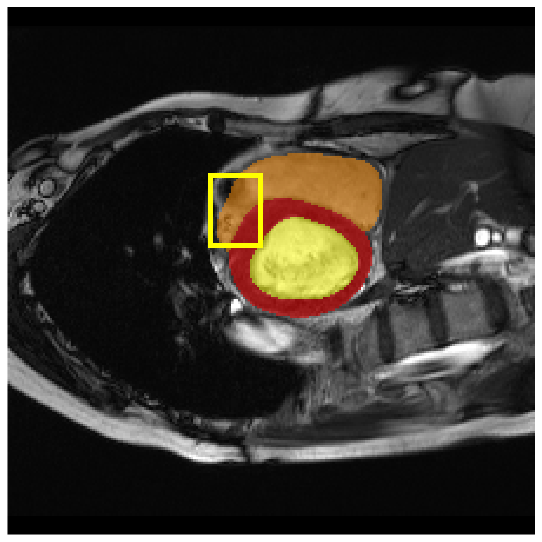} &
        \includegraphics[width=0.121\linewidth]{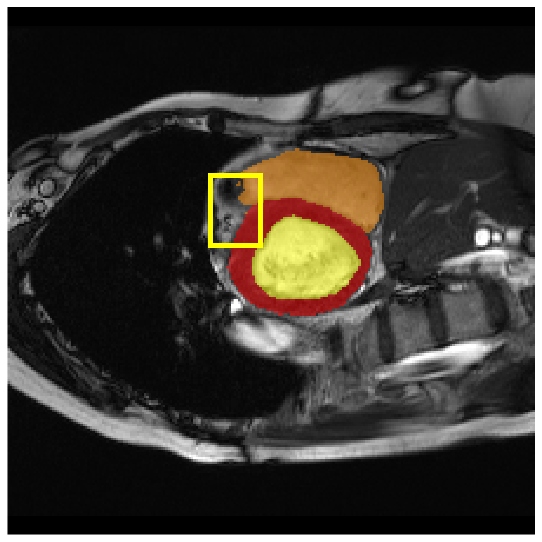} &
        \includegraphics[width=0.121\linewidth]{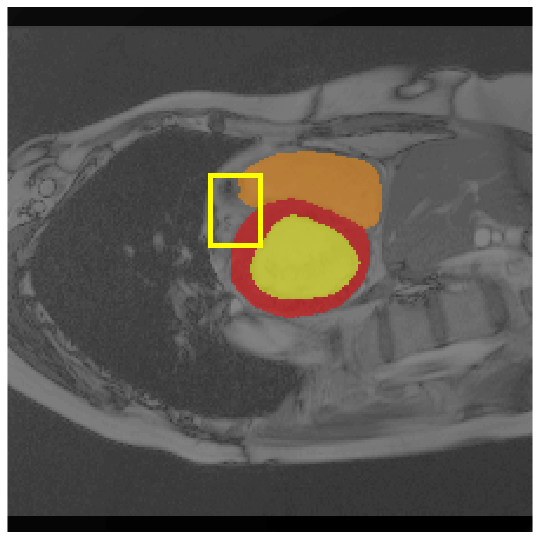} &
        \includegraphics[width=0.121\linewidth]{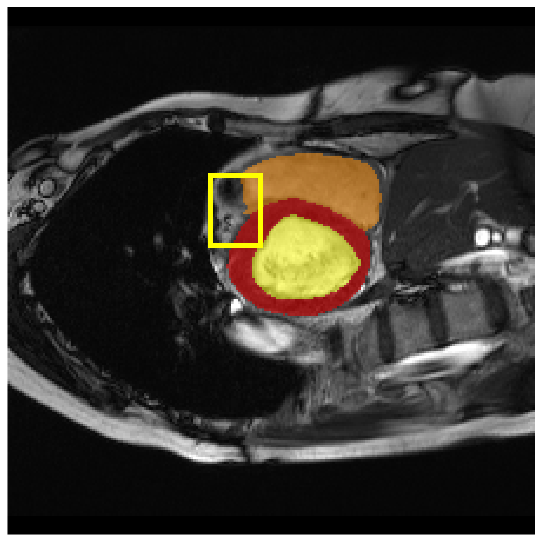} &
        \includegraphics[width=0.121\linewidth]{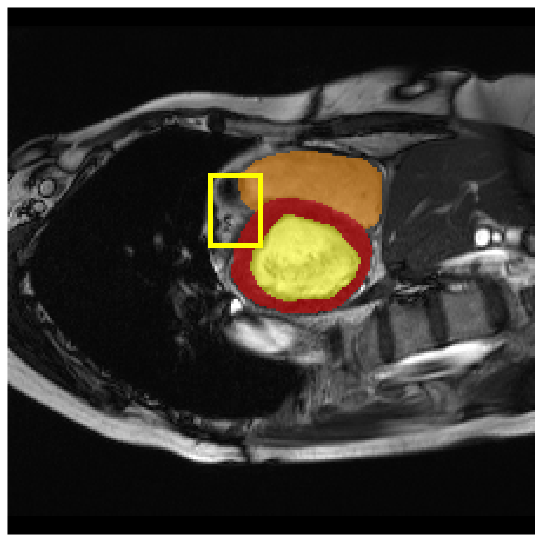} \\
        \includegraphics[width=0.121\linewidth]{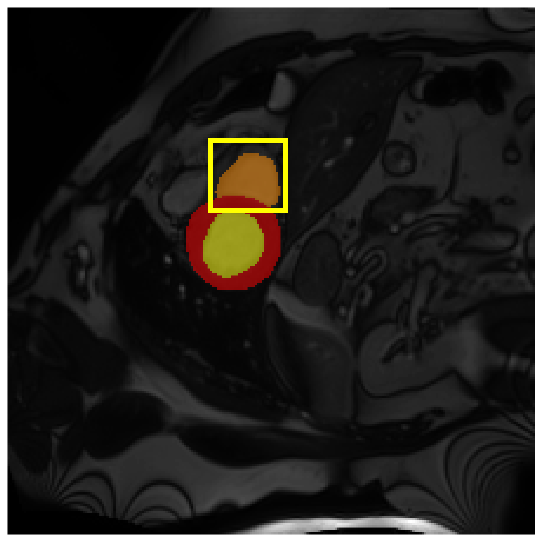} & 
        \includegraphics[width=0.121\linewidth]{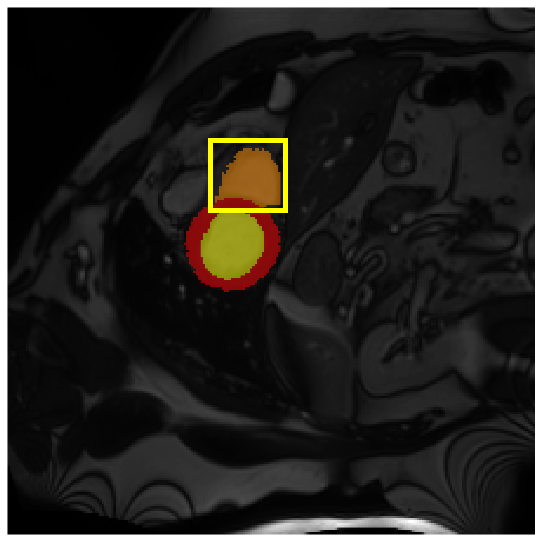} & 
        \includegraphics[width=0.121\linewidth]{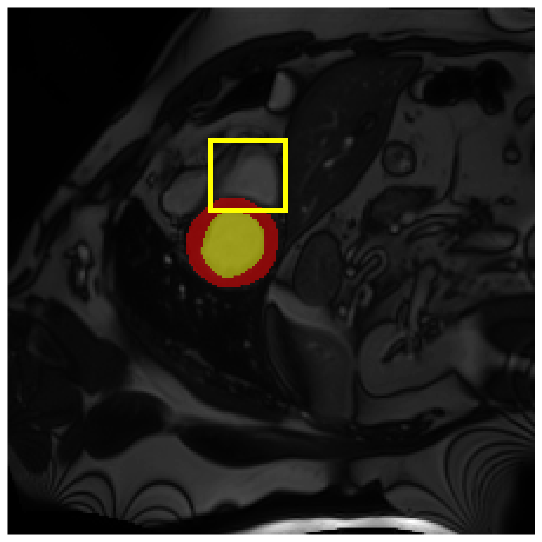} &
        \includegraphics[width=0.121\linewidth]{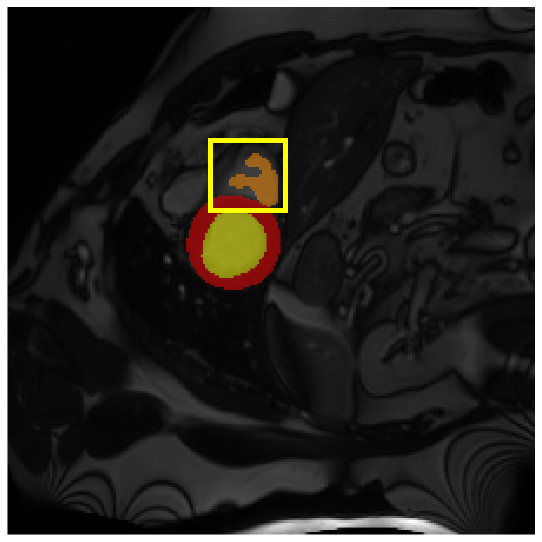} &
        \includegraphics[width=0.121\linewidth]{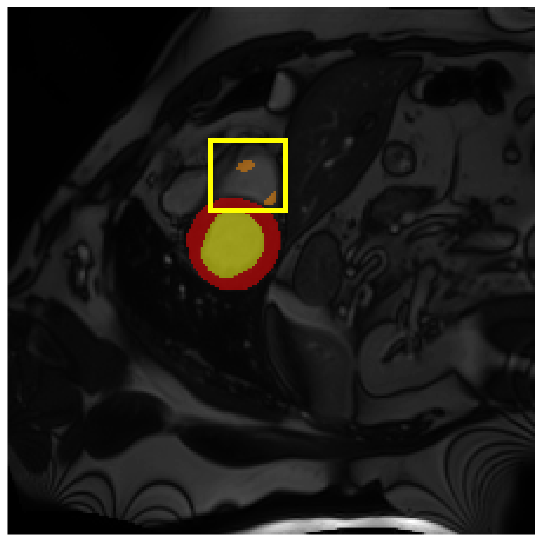} &
        \includegraphics[width=0.121\linewidth]{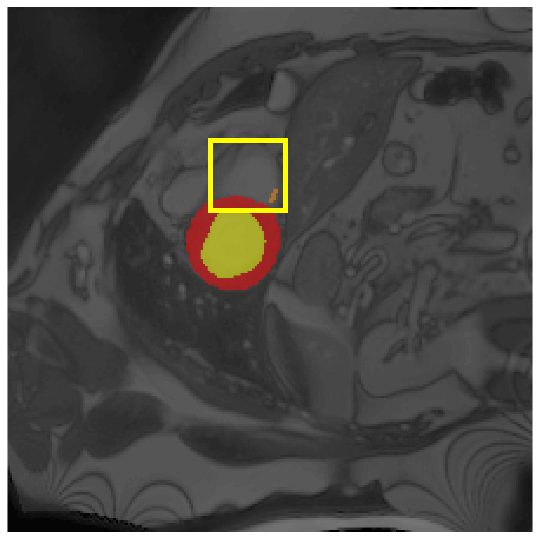} &
        \includegraphics[width=0.121\linewidth]{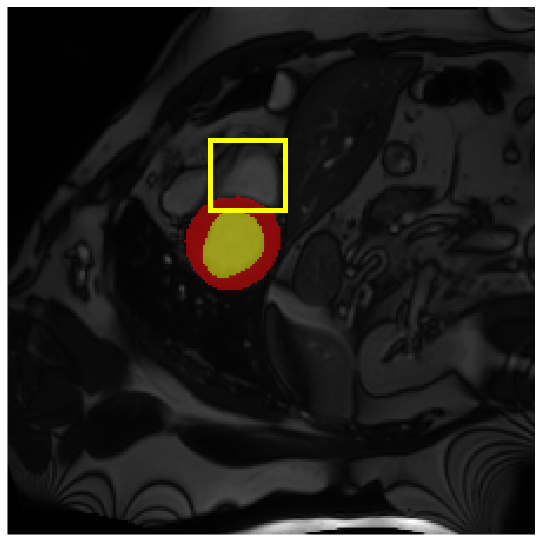} &
        \includegraphics[width=0.121\linewidth]{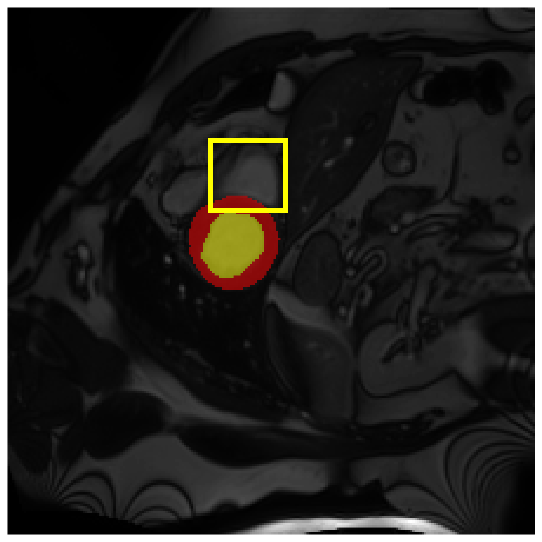} \\
        \includegraphics[width=0.121\linewidth]{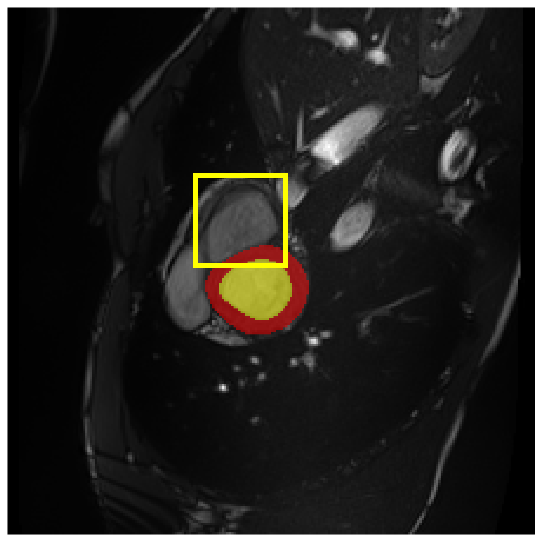} & 
        \includegraphics[width=0.121\linewidth]{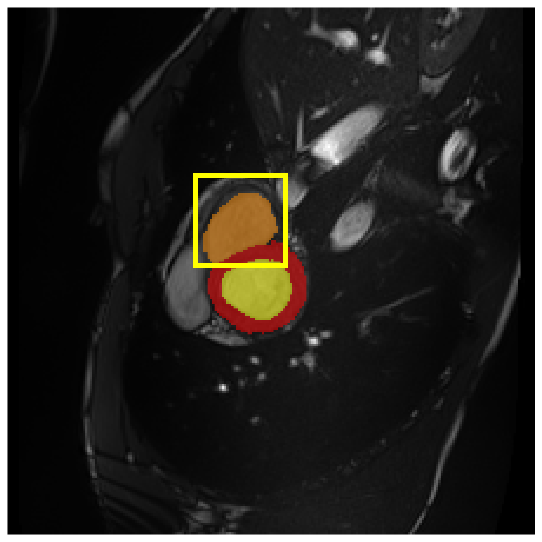} & 
        \includegraphics[width=0.121\linewidth]{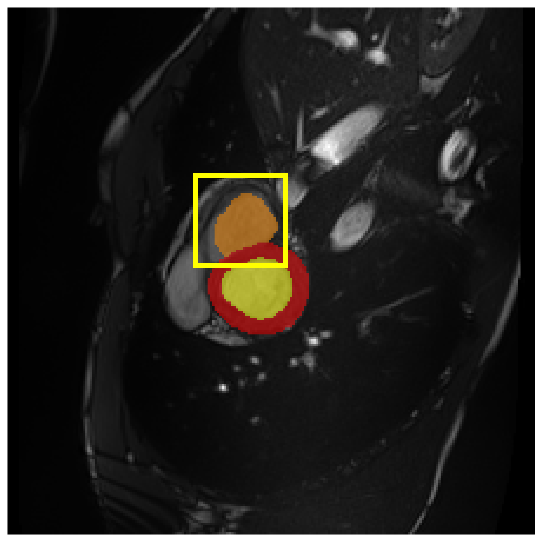} &
        \includegraphics[width=0.121\linewidth]{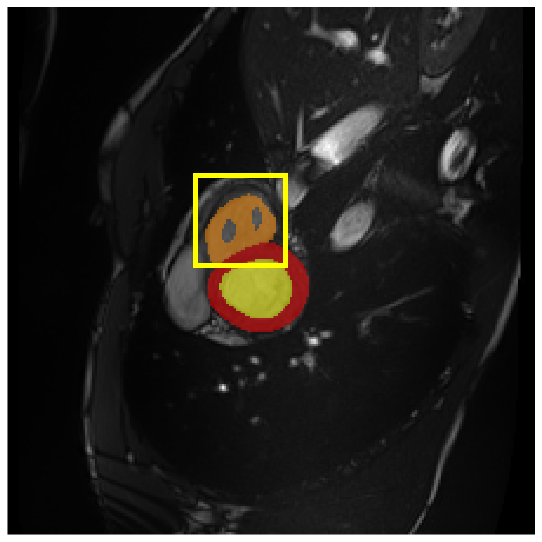} &
        \includegraphics[width=0.121\linewidth]{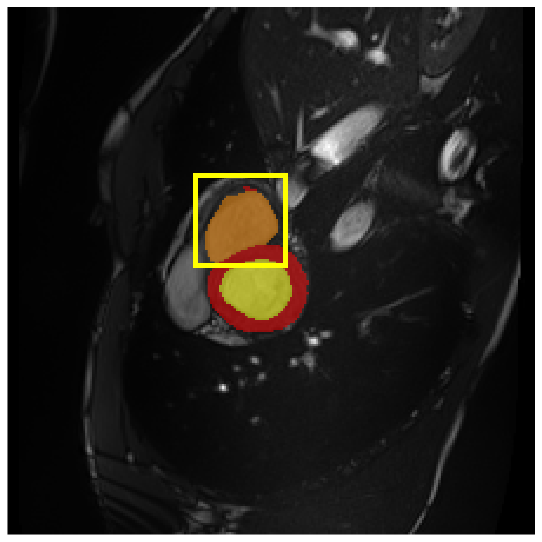} &
        \includegraphics[width=0.121\linewidth]{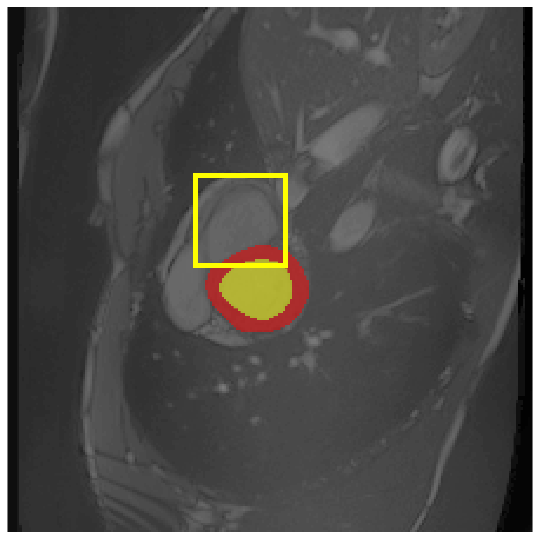} &
        \includegraphics[width=0.121\linewidth]{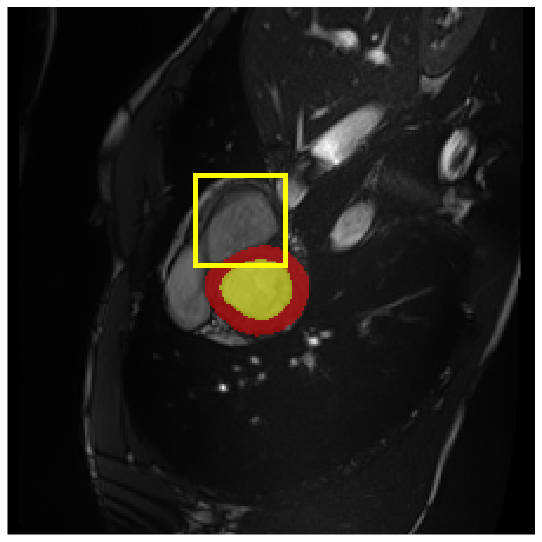} &
        \includegraphics[width=0.121\linewidth]{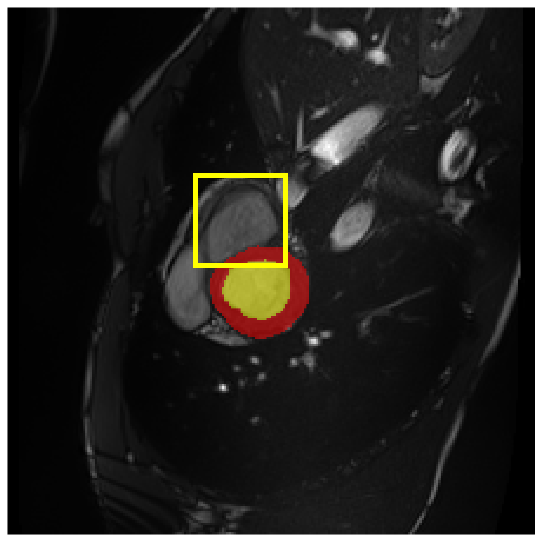} \\
        \includegraphics[width=0.121\linewidth]{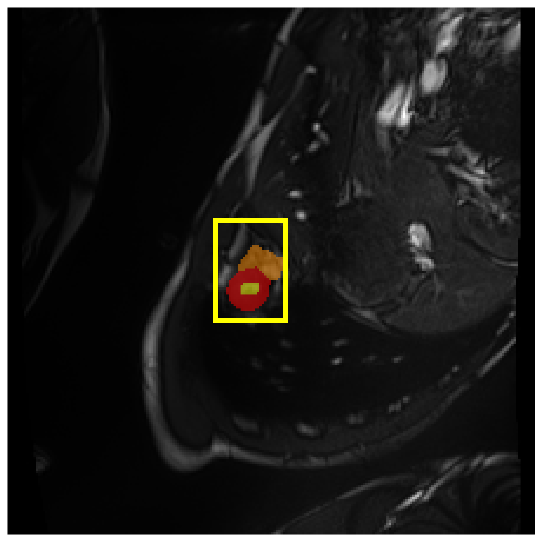} & 
        \includegraphics[width=0.121\linewidth]{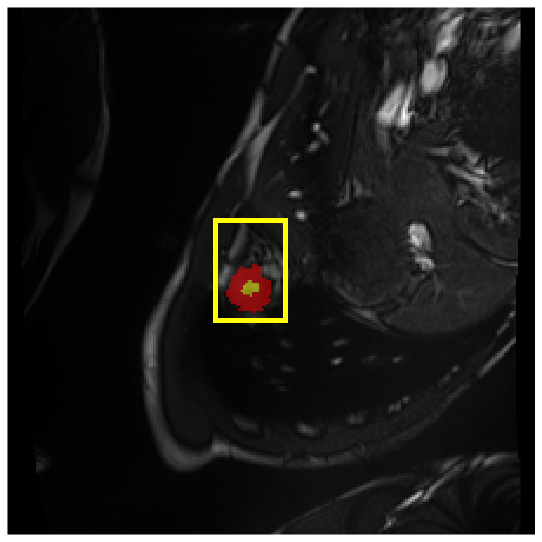} & 
        \includegraphics[width=0.121\linewidth]{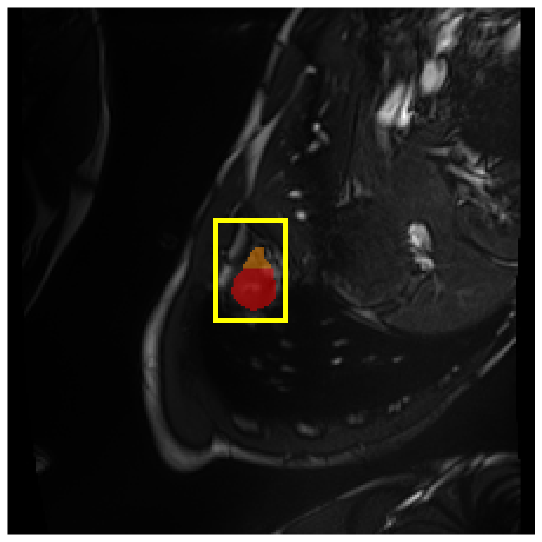} &
        \includegraphics[width=0.121\linewidth]{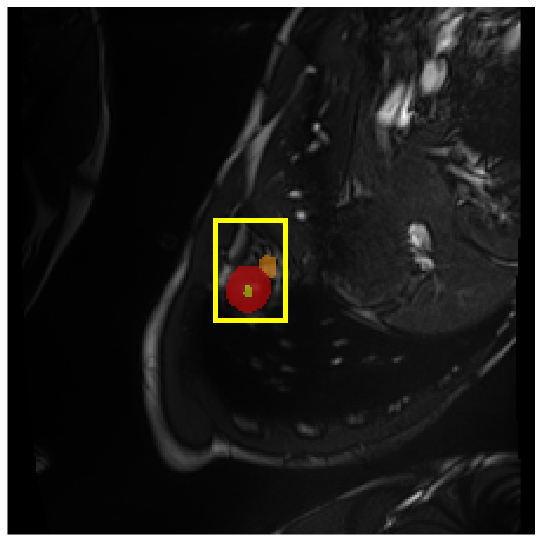} &
        \includegraphics[width=0.121\linewidth]{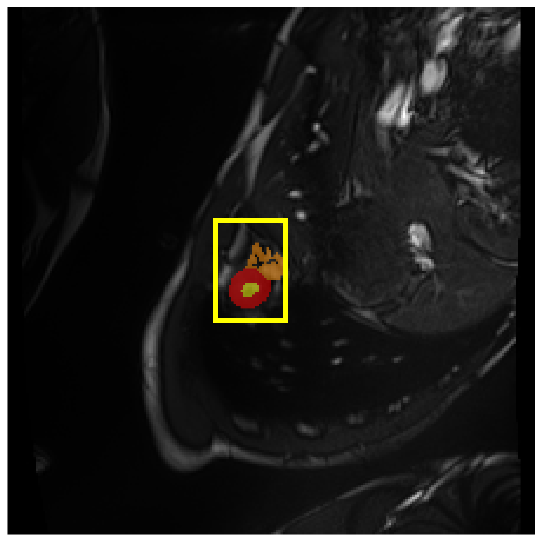} &
        \includegraphics[width=0.121\linewidth]{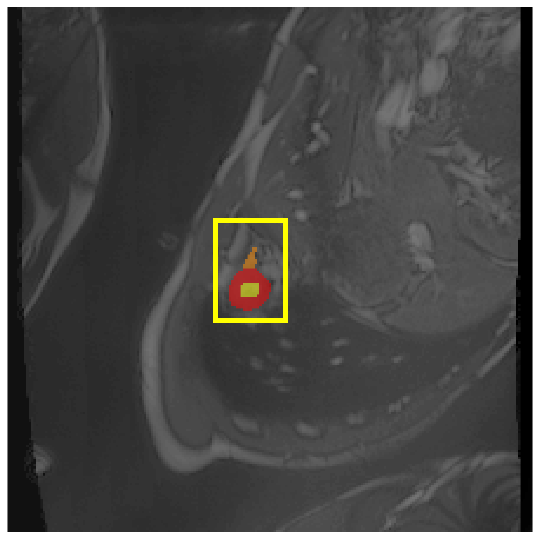} &
        \includegraphics[width=0.121\linewidth]{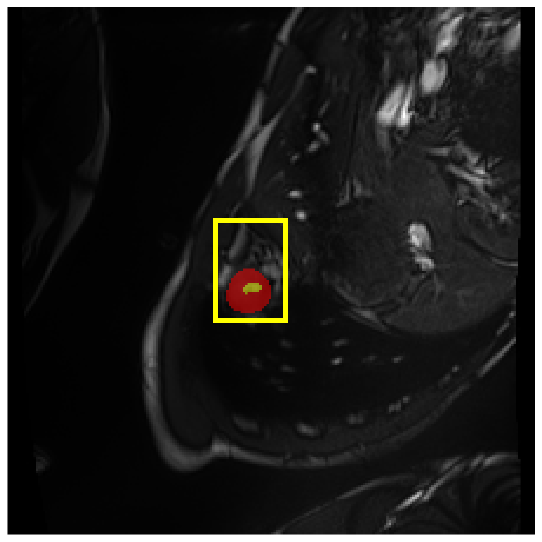} &
        \includegraphics[width=0.121\linewidth]{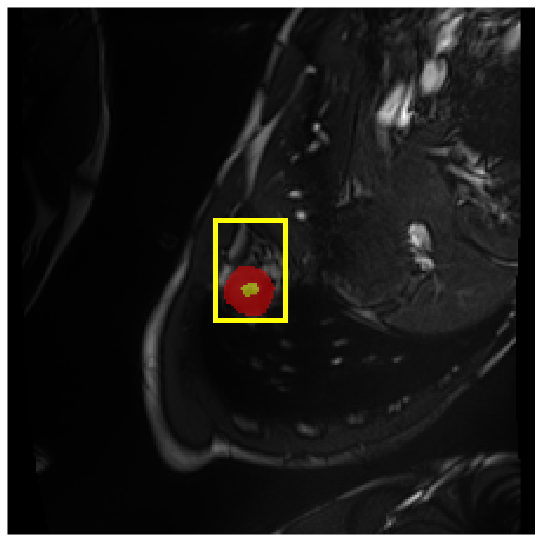} \\
        \includegraphics[width=0.121\linewidth]{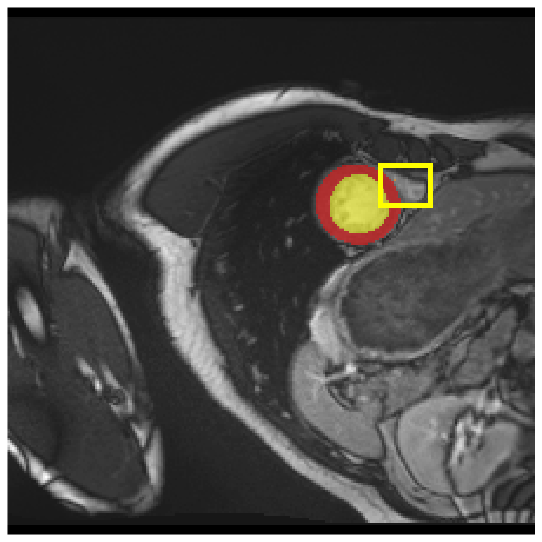} & 
        \includegraphics[width=0.121\linewidth]{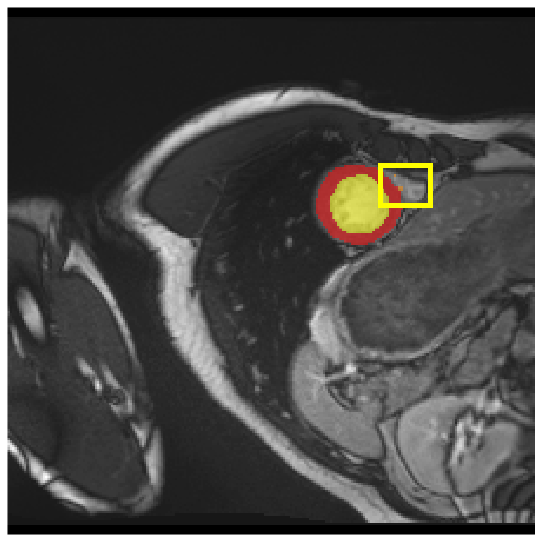} & 
        \includegraphics[width=0.121\linewidth]{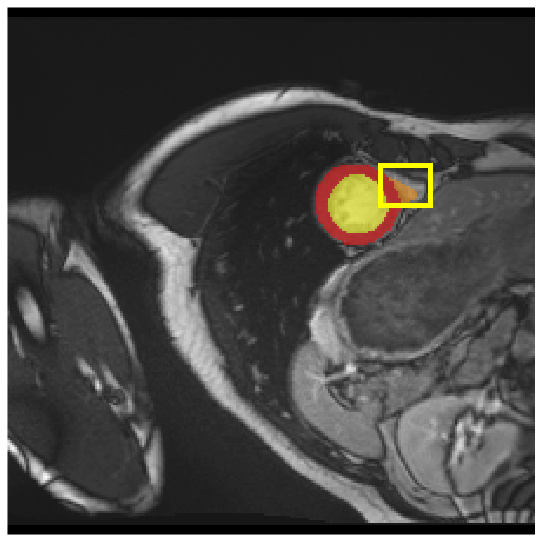} &
        \includegraphics[width=0.121\linewidth]{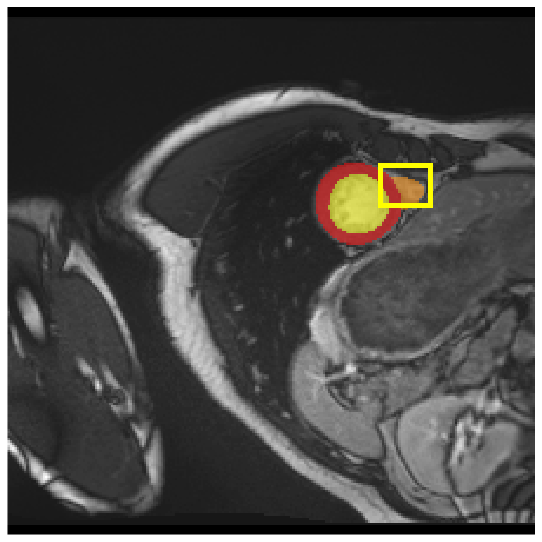} &
        \includegraphics[width=0.121\linewidth]{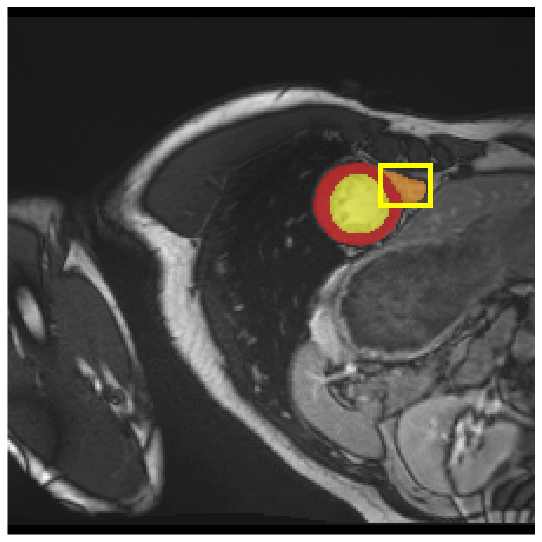} &
        \includegraphics[width=0.121\linewidth]{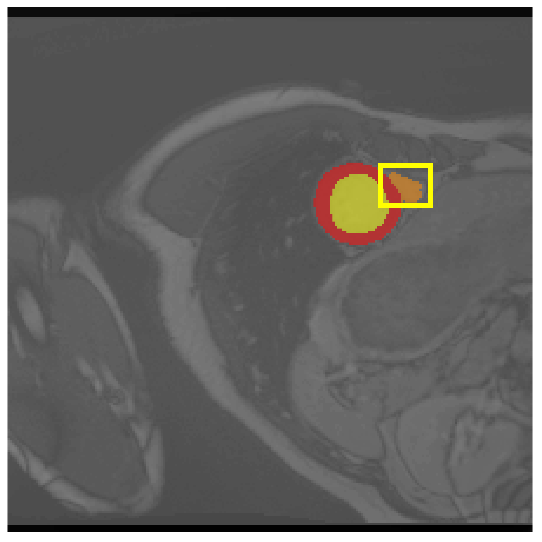} &
        \includegraphics[width=0.121\linewidth]{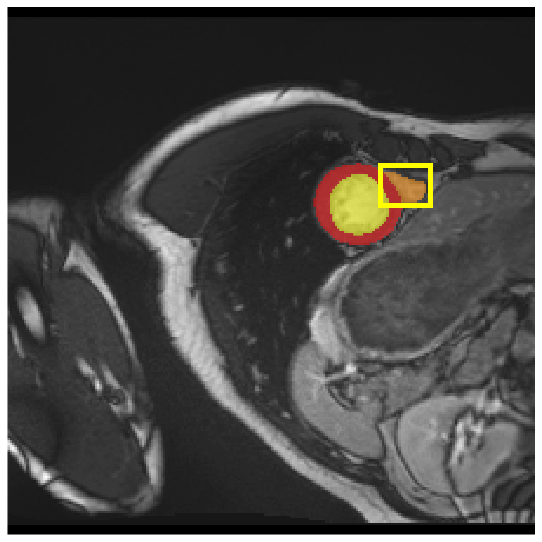} &
        \includegraphics[width=0.121\linewidth]{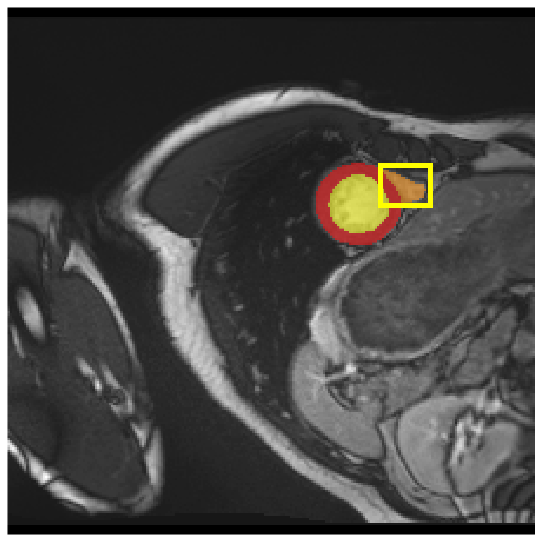} \\
        \includegraphics[width=0.121\linewidth]{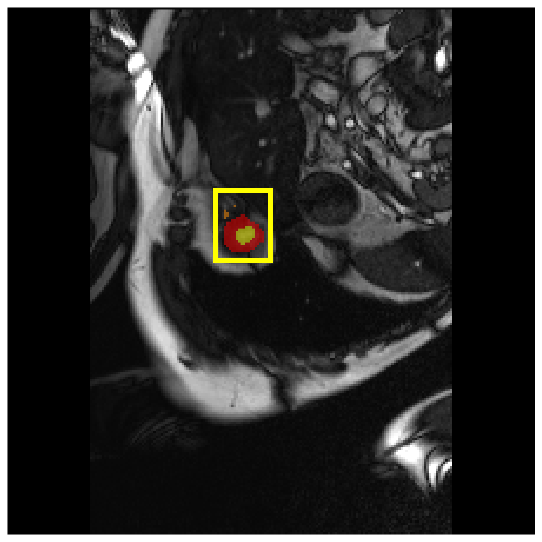} & 
        \includegraphics[width=0.121\linewidth]{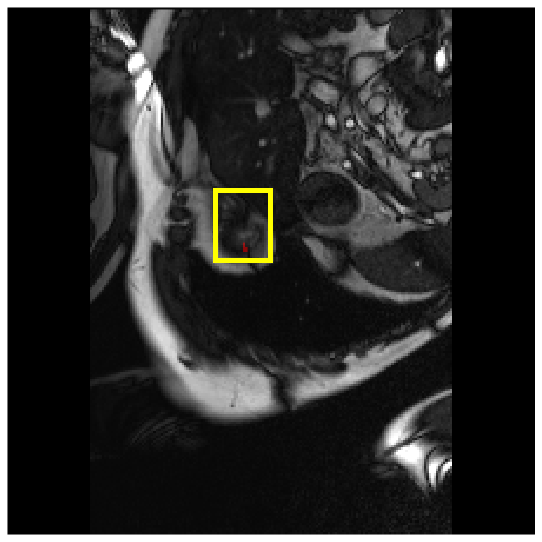} & 
        \includegraphics[width=0.121\linewidth]{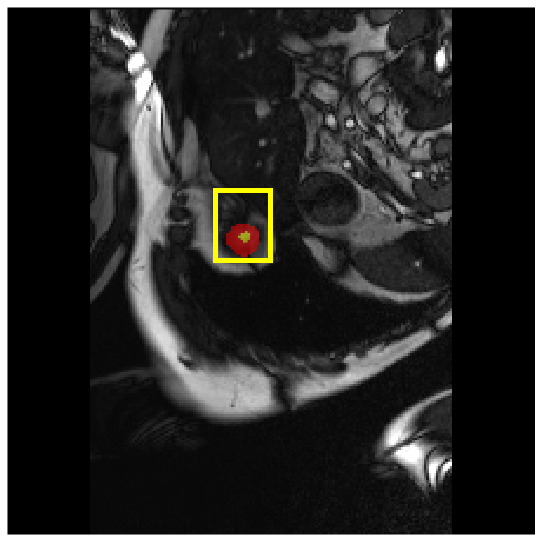} &
        \includegraphics[width=0.121\linewidth]{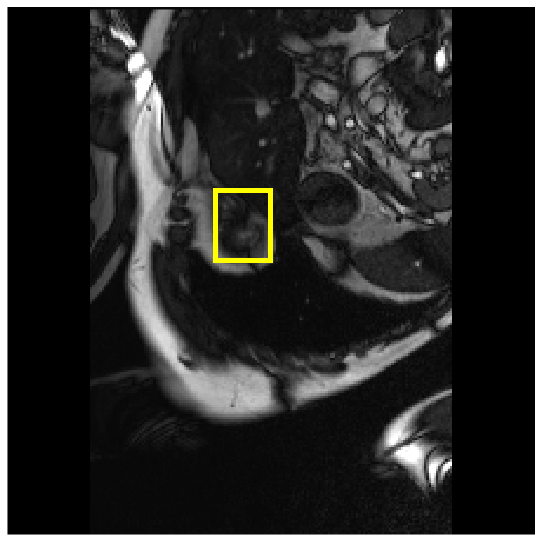} &
        \includegraphics[width=0.121\linewidth]{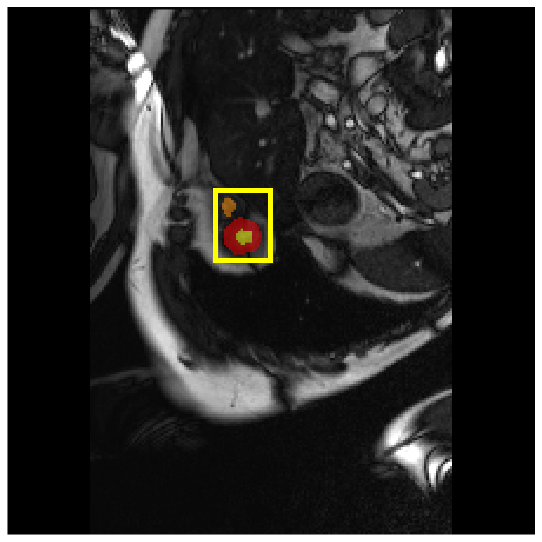} &
        \includegraphics[width=0.121\linewidth]{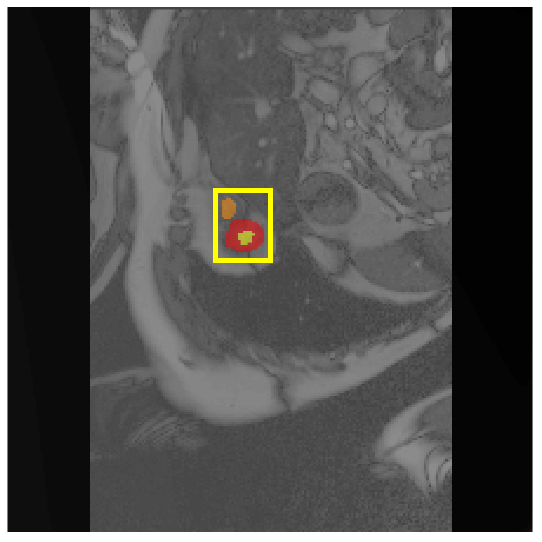} &
        \includegraphics[width=0.121\linewidth]{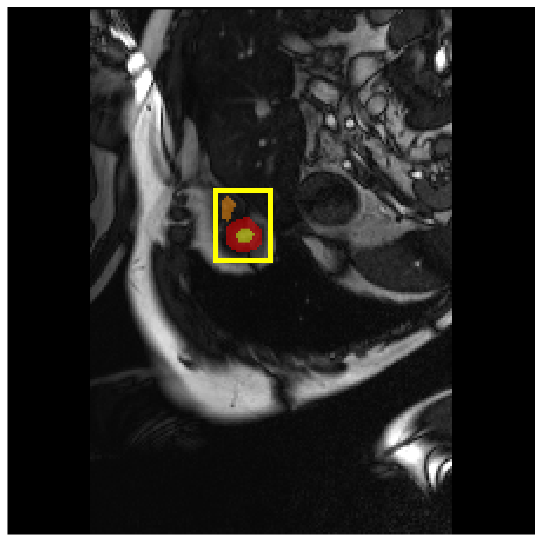} &
        \includegraphics[width=0.121\linewidth]{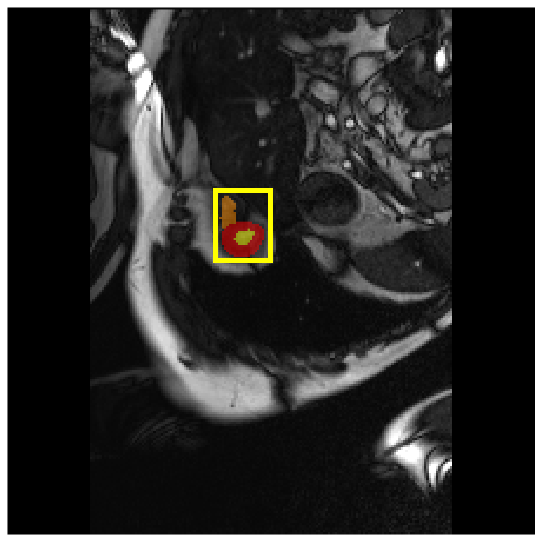} \\
        \includegraphics[width=0.121\linewidth]{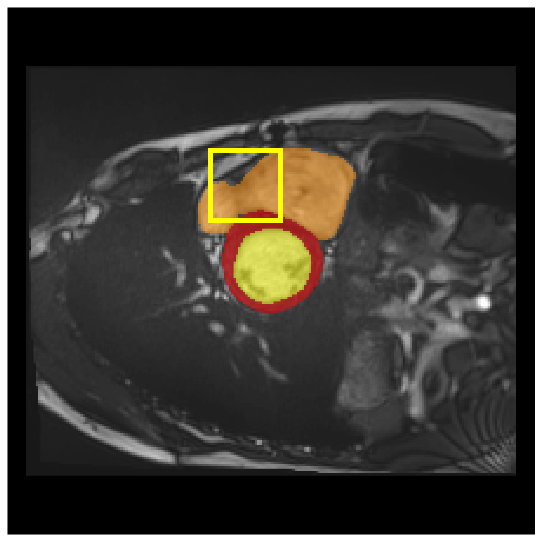} & 
        \includegraphics[width=0.121\linewidth]{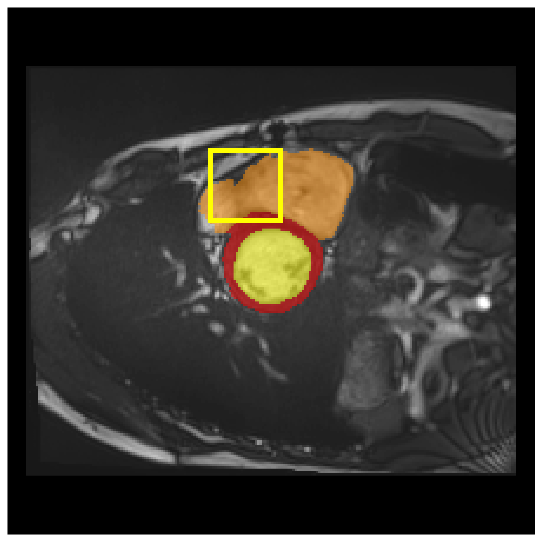} & 
        \includegraphics[width=0.121\linewidth]{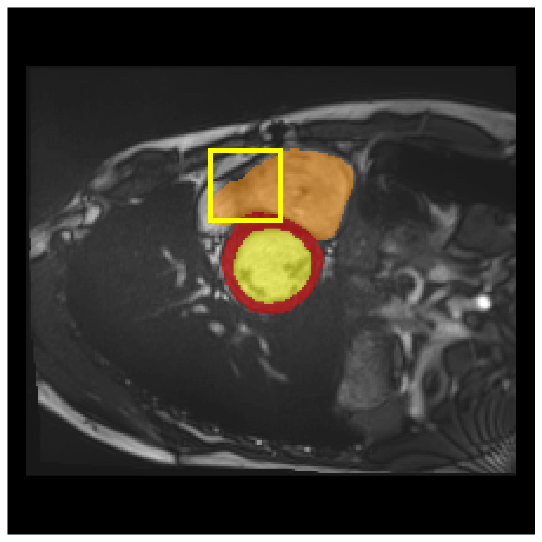} &
        \includegraphics[width=0.121\linewidth]{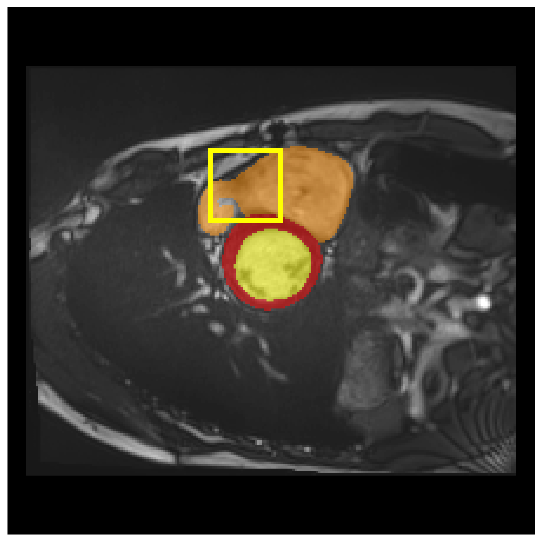} &
        \includegraphics[width=0.121\linewidth]{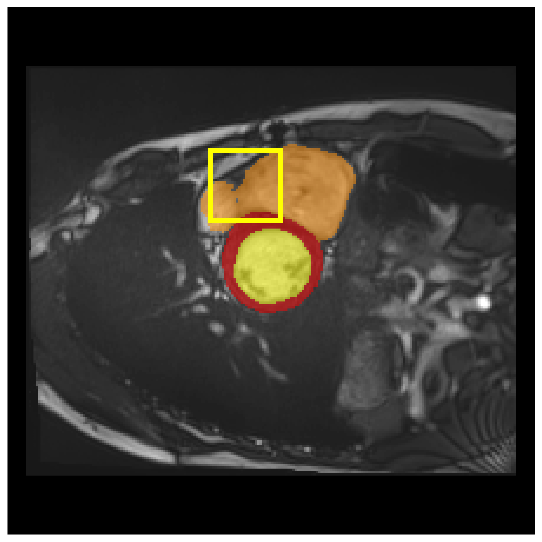} &
        \includegraphics[width=0.121\linewidth]{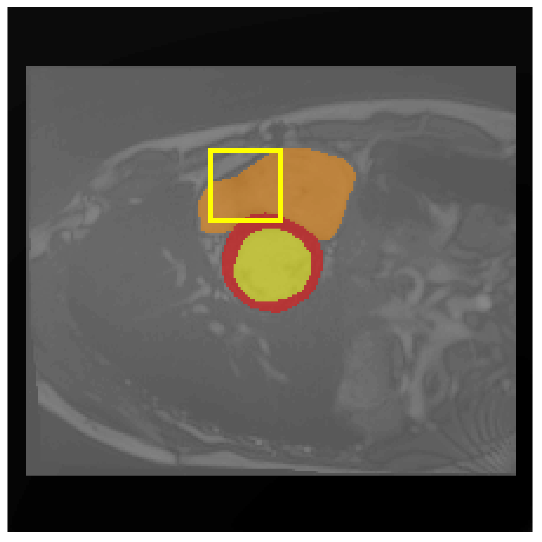} &
        \includegraphics[width=0.121\linewidth]{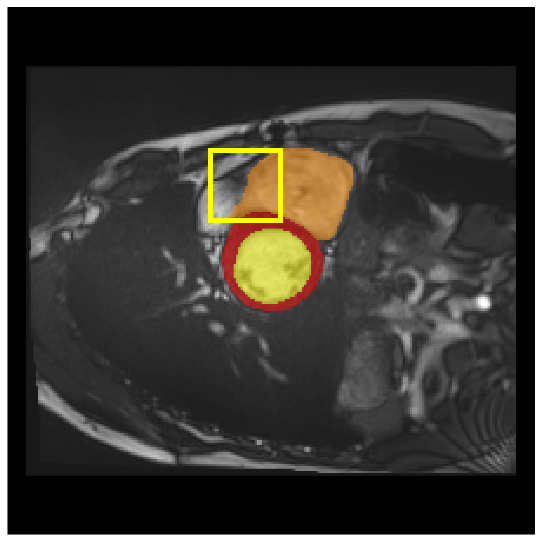} &
        \includegraphics[width=0.121\linewidth]{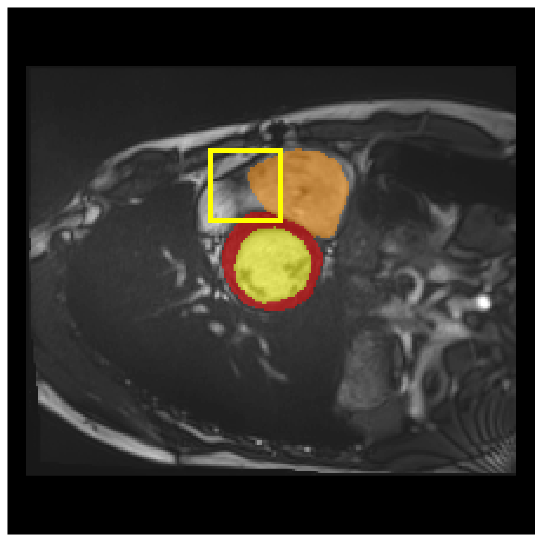} \\
        \includegraphics[width=0.121\linewidth]{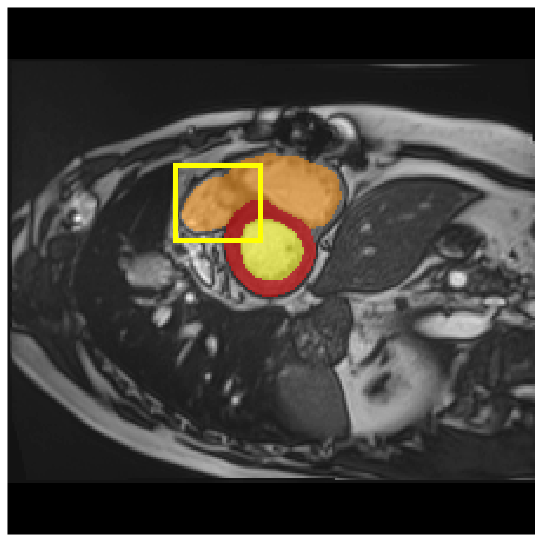} & 
        \includegraphics[width=0.121\linewidth]{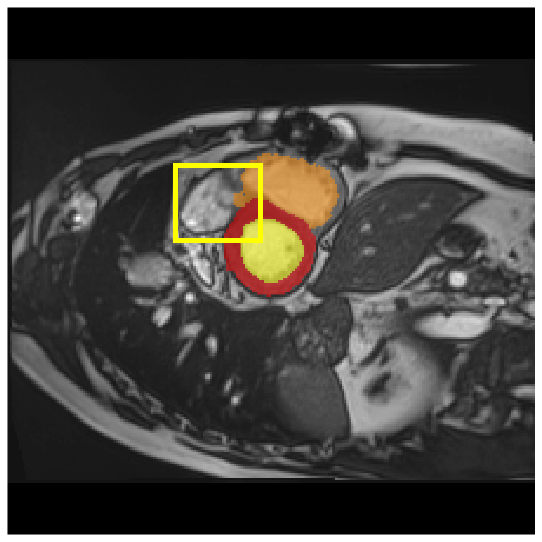} & 
        \includegraphics[width=0.121\linewidth]{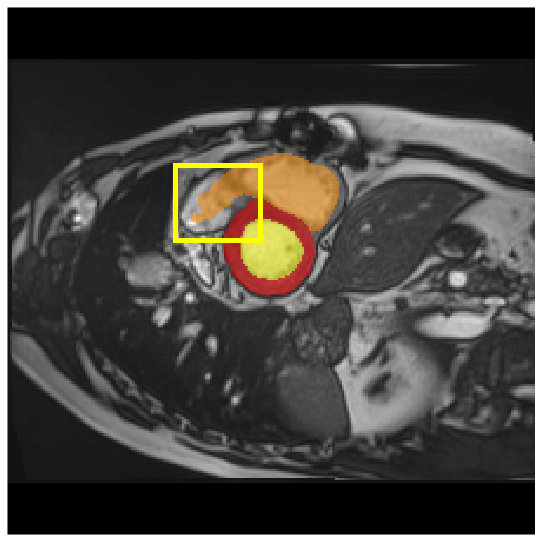} &
        \includegraphics[width=0.121\linewidth]{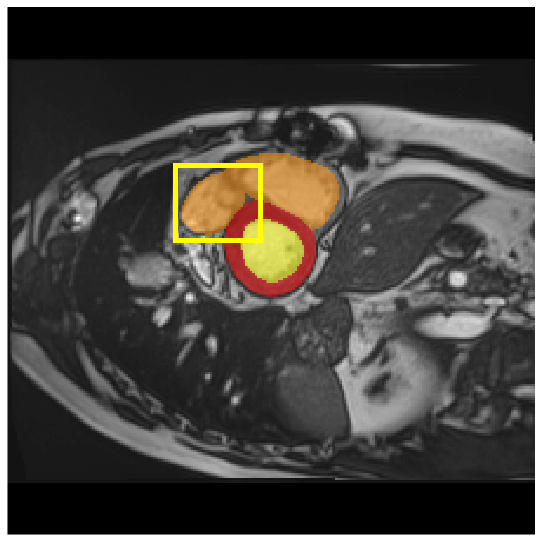} &
        \includegraphics[width=0.121\linewidth]{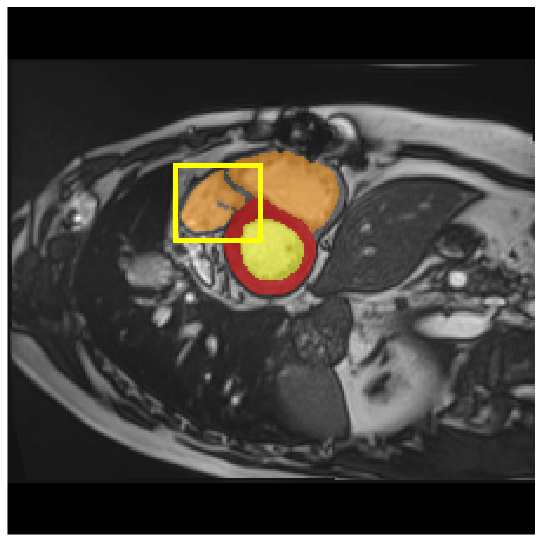} &
        \includegraphics[width=0.121\linewidth]{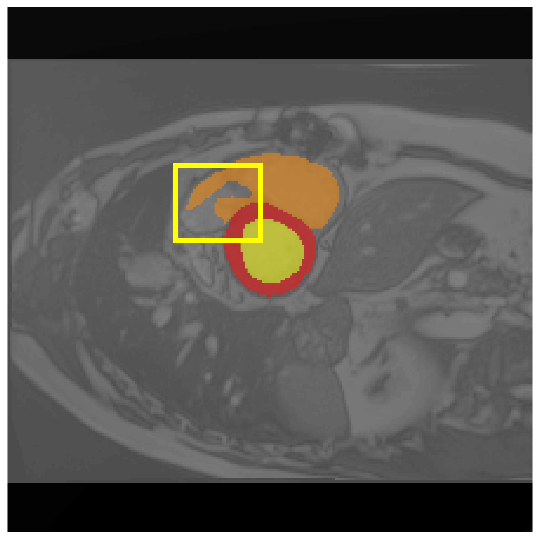} &
        \includegraphics[width=0.121\linewidth]{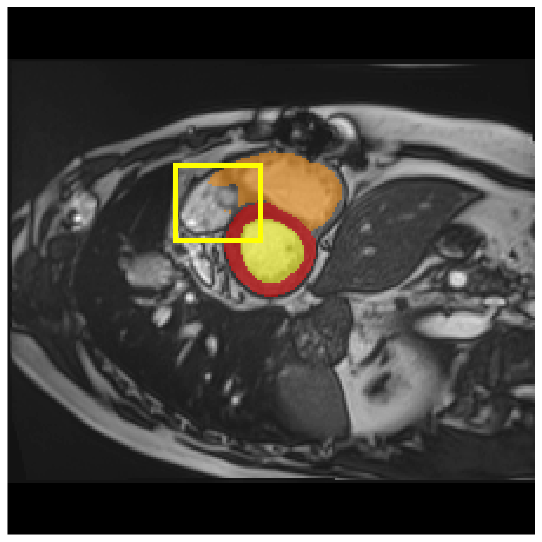} &
        \includegraphics[width=0.121\linewidth]{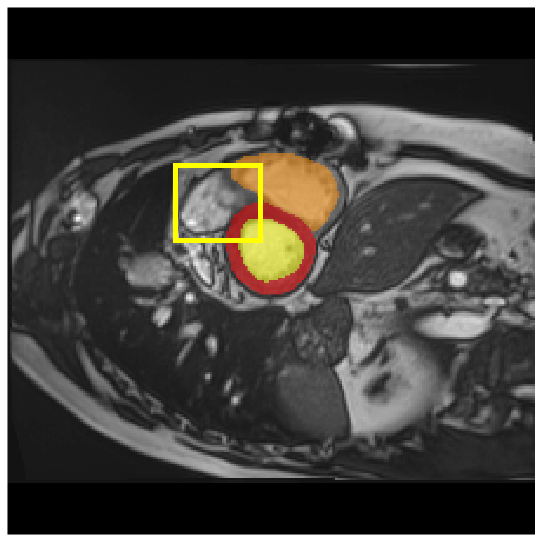} \\
        \footnotesize (a) & \footnotesize (b) & \footnotesize (c) & \footnotesize (d) & \footnotesize (e) & \footnotesize (f) &  \footnotesize (g) & \footnotesize (h)
    \end{tabular}
    \caption{
    Additional visual comparison of segmentation results on the ACDC dataset. (a) TransUNet~\cite{chen2021transunet}, (b) SwinUNet~\cite{cao2022swinunet}, (c) MERIT~\cite{rahman2023multi}, (d) FCT~\cite{tragakis2023fully}, (e) only $I$, (f) only $E$, (g) Ours, and (h) GT, respectively.
    Yellow boxes highlight regions in which our method excels at segmentation.
    %
    }
    \label{fig:acdc_result}
\end{figure*}

\begin{figure*}
    \centering
    \renewcommand{\arraystretch}{0.2}
    \begin{tabular}{@{}c@{\hskip 0.003\linewidth}c@{\hskip 0.003\linewidth}c@{\hskip 0.003\linewidth}c@{\hskip 0.003\linewidth}c@{\hskip 0.003\linewidth}c@{\hskip 0.003\linewidth}c@{\hskip 0.003\linewidth}c@{\hskip 0.003\linewidth}}
    \multicolumn{8}{l}{\includegraphics[width=0.968\linewidth]{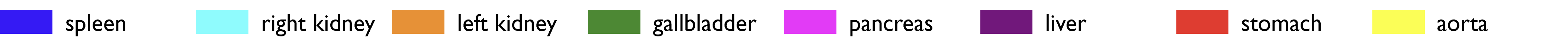}} \\
        \includegraphics[width=0.121\linewidth]{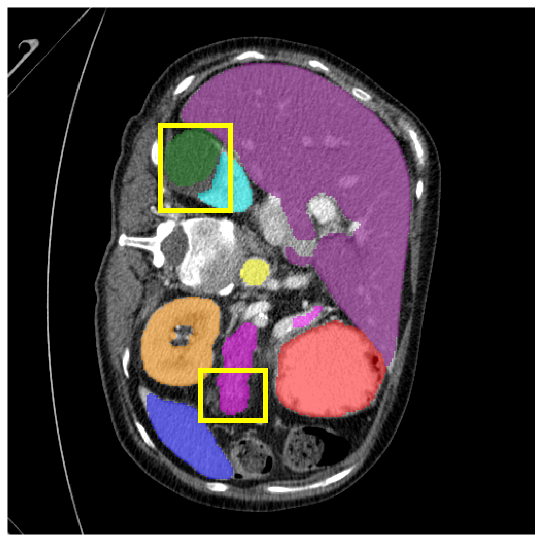} & 
        \includegraphics[width=0.121\linewidth]{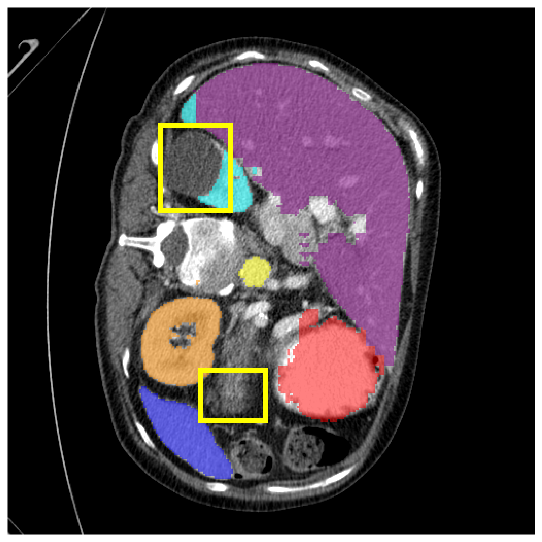} & 
        \includegraphics[width=0.121\linewidth]{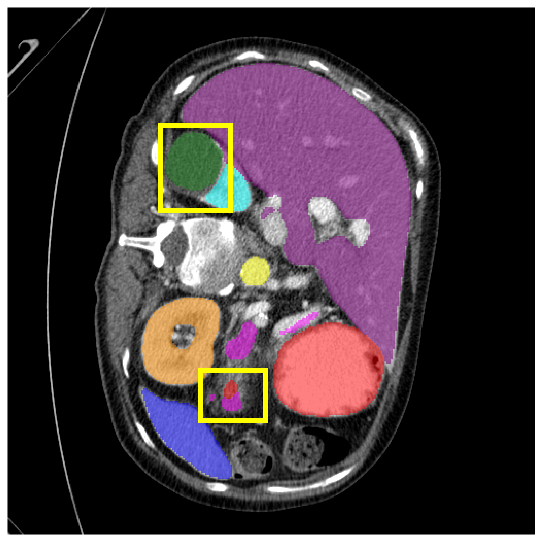} & 
        \includegraphics[width=0.121\linewidth]{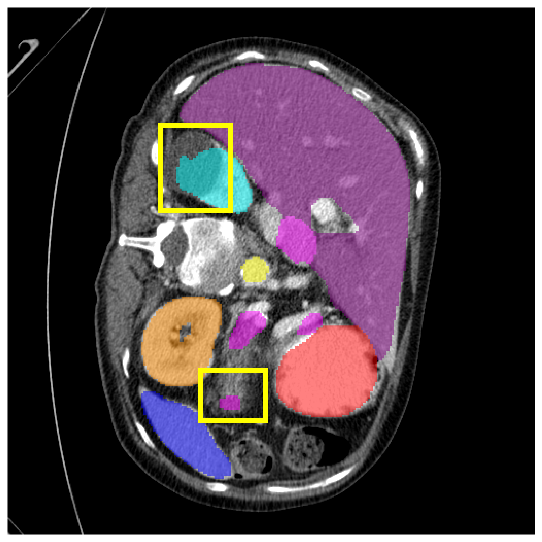} & 
        \includegraphics[width=0.121\linewidth]{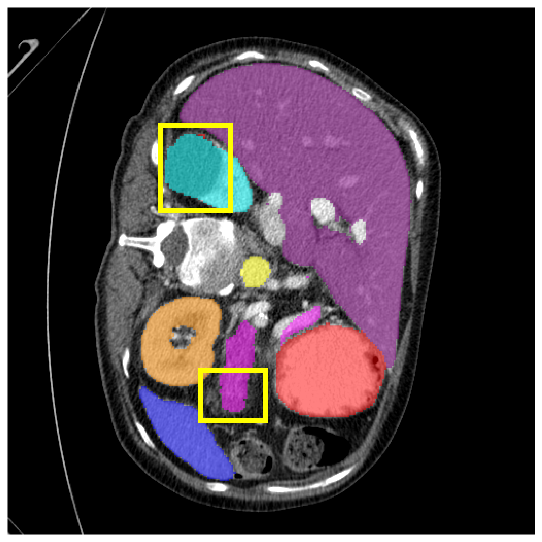} & 
        \includegraphics[width=0.121\linewidth]{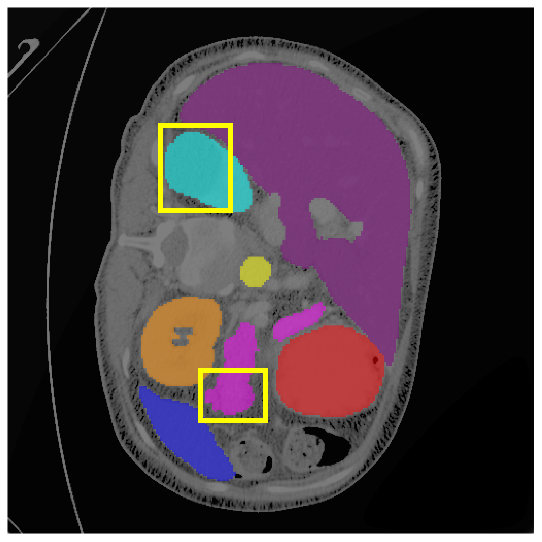} & 
        \includegraphics[width=0.121\linewidth]{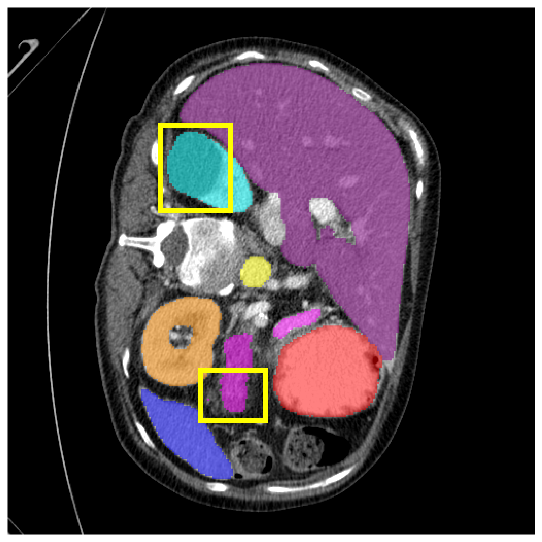} & 
        \includegraphics[width=0.121\linewidth]{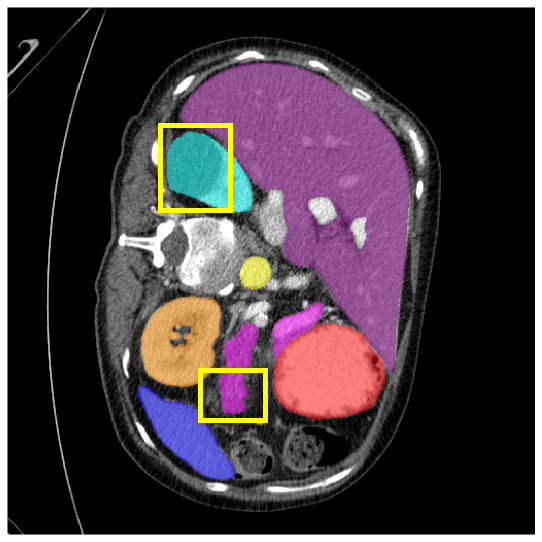} \\
        \includegraphics[width=0.121\linewidth]{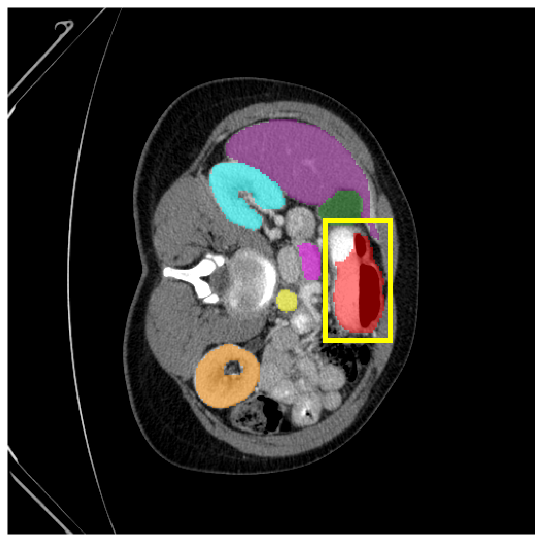} & 
        \includegraphics[width=0.121\linewidth]{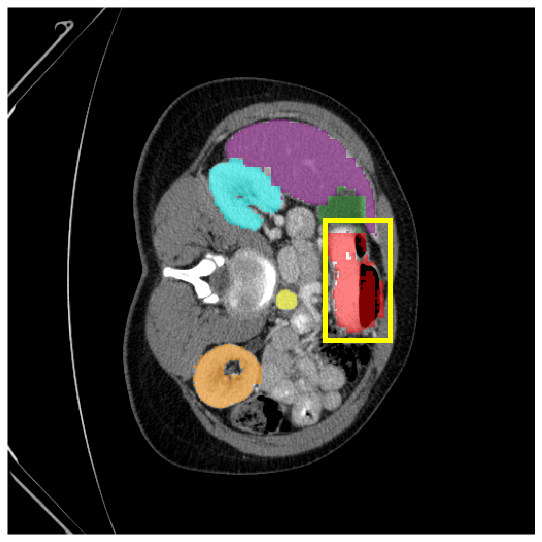} & 
        \includegraphics[width=0.121\linewidth]{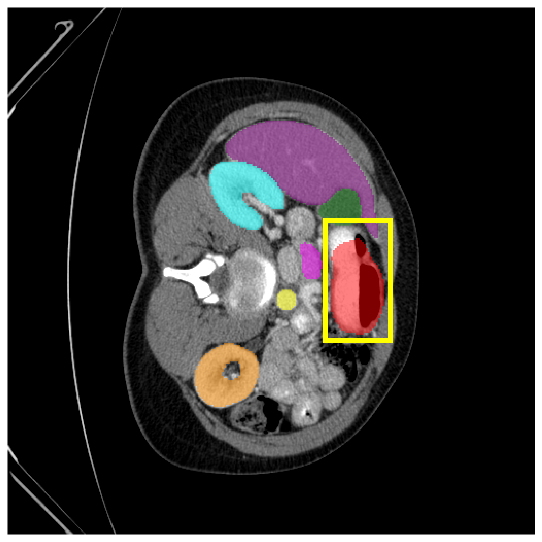} & 
        \includegraphics[width=0.121\linewidth]{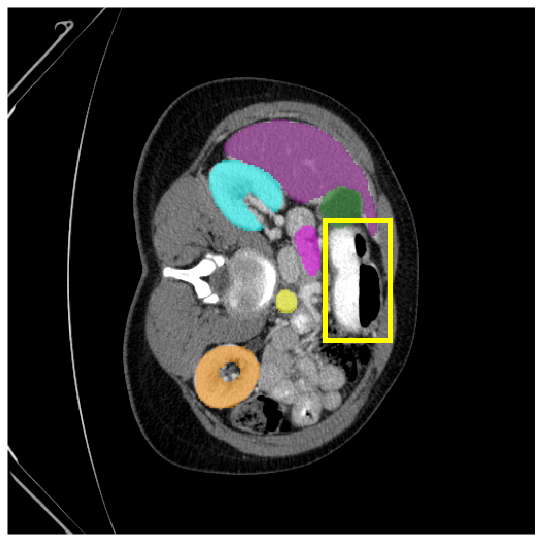} & 
        \includegraphics[width=0.121\linewidth]{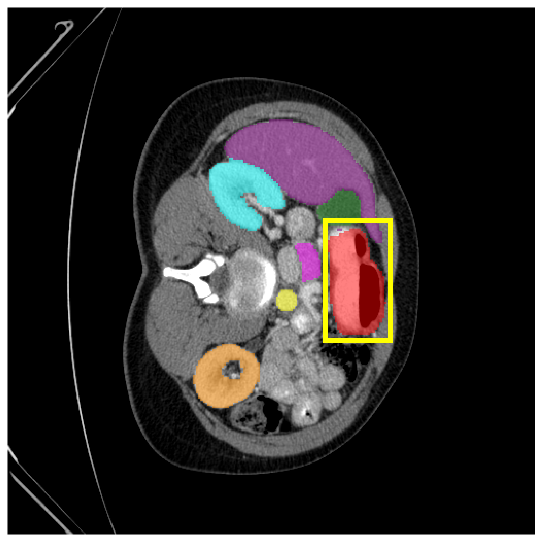} & 
        \includegraphics[width=0.121\linewidth]{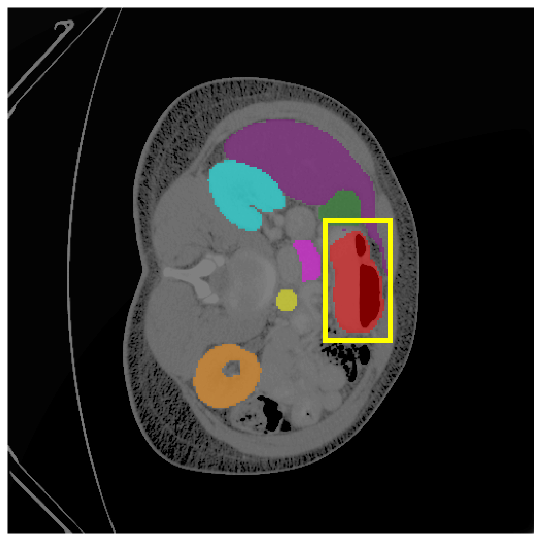} & 
        \includegraphics[width=0.121\linewidth]{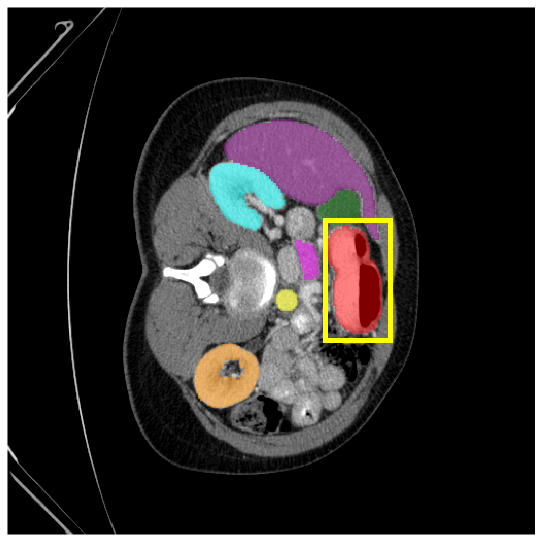} & 
        \includegraphics[width=0.121\linewidth]{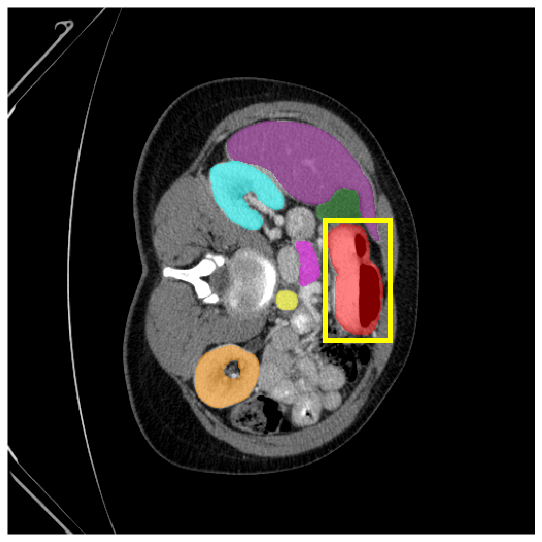} \\
        \includegraphics[width=0.121\linewidth]{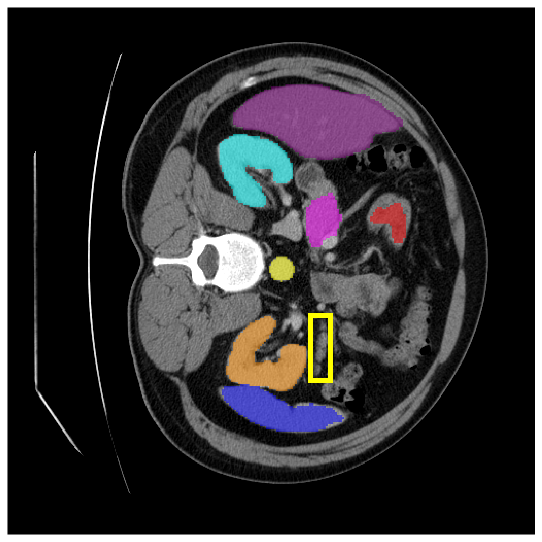} & 
        \includegraphics[width=0.121\linewidth]{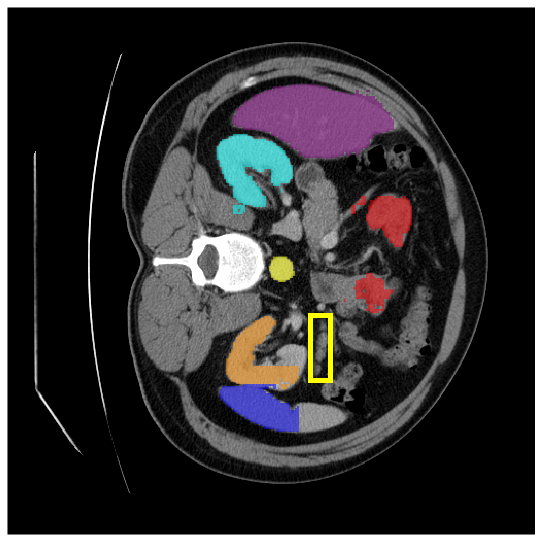} & 
        \includegraphics[width=0.121\linewidth]{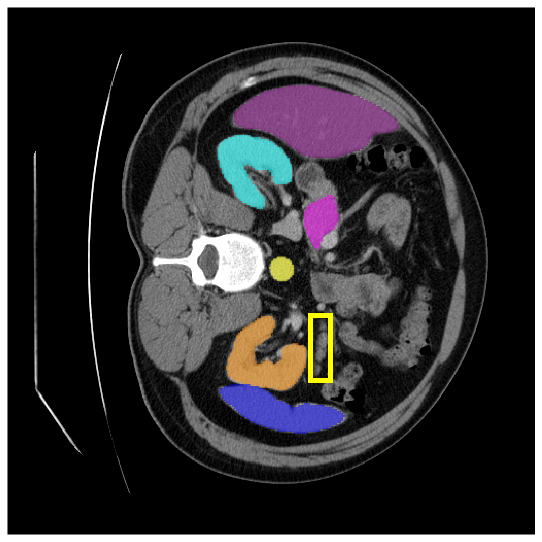} & 
        \includegraphics[width=0.121\linewidth]{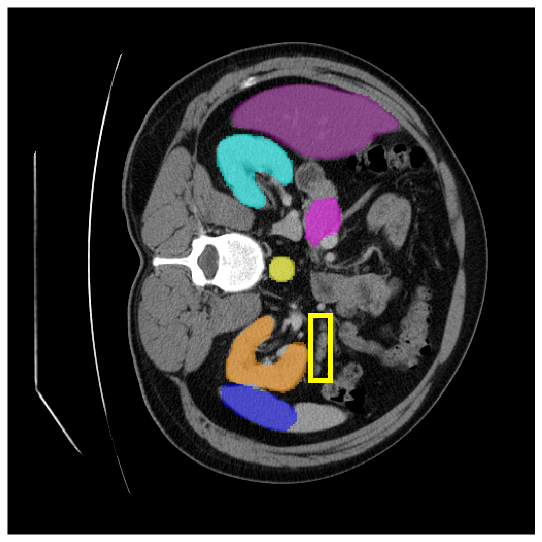} & 
        \includegraphics[width=0.121\linewidth]{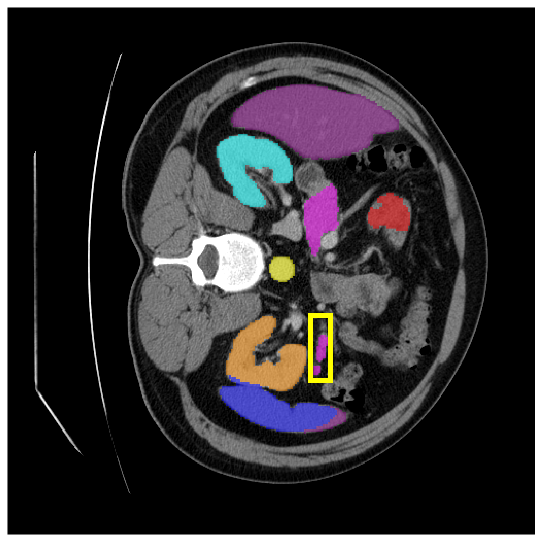} & 
        \includegraphics[width=0.121\linewidth]{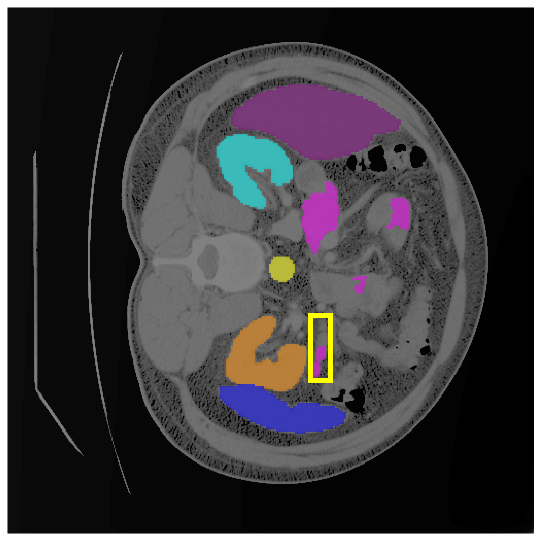} & 
        \includegraphics[width=0.121\linewidth]{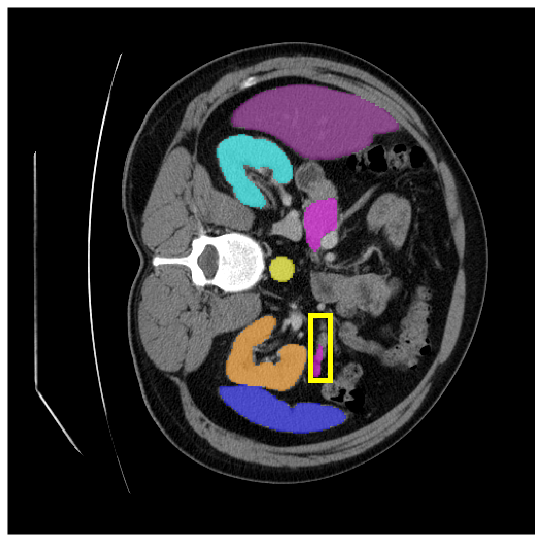} & 
        \includegraphics[width=0.121\linewidth]{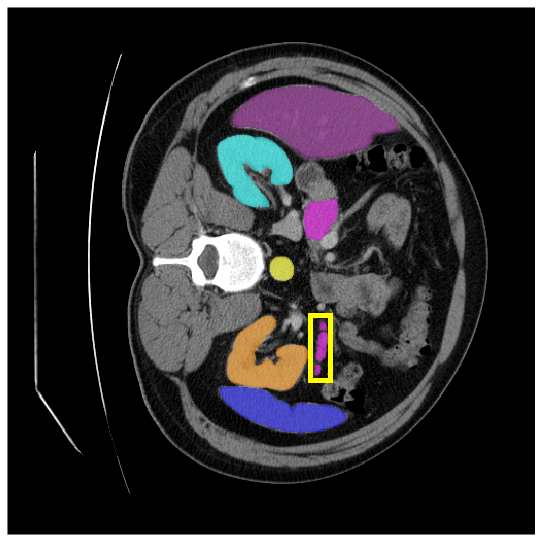} \\
        \includegraphics[width=0.121\linewidth]{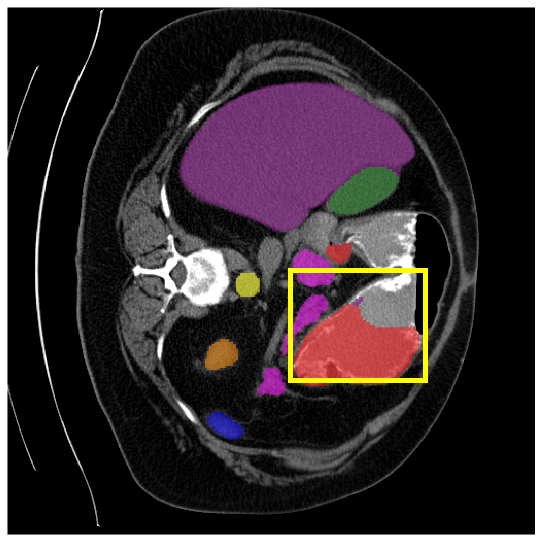} & 
        \includegraphics[width=0.121\linewidth]{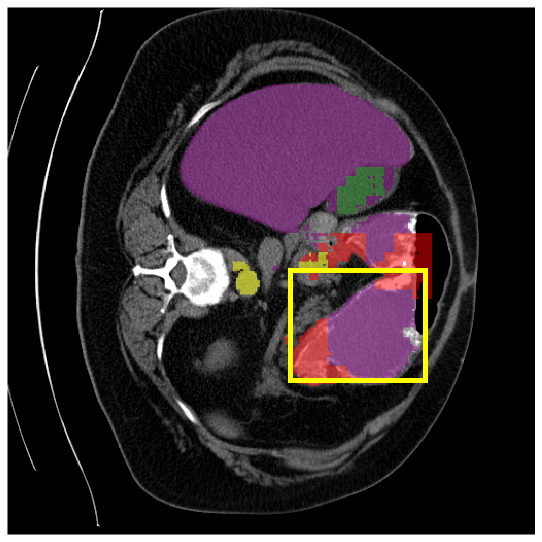} & 
        \includegraphics[width=0.121\linewidth]{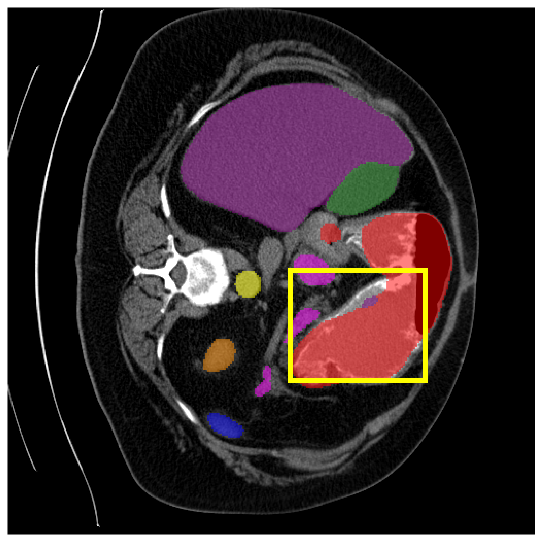} & 
        \includegraphics[width=0.121\linewidth]{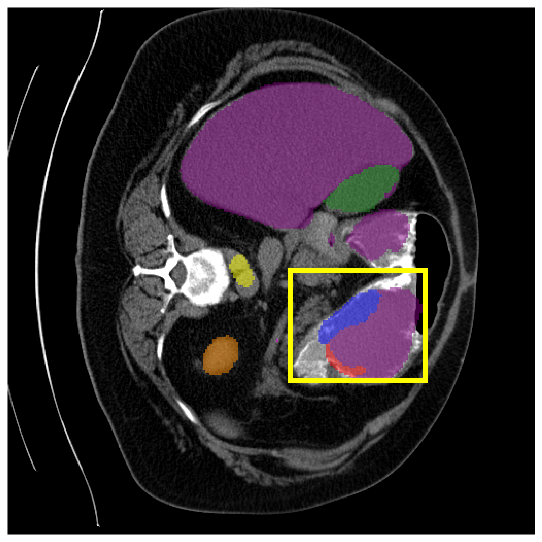} & 
        \includegraphics[width=0.121\linewidth]{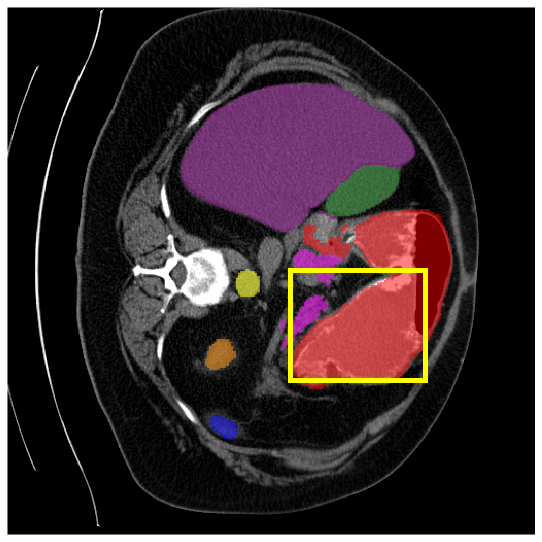} & 
        \includegraphics[width=0.121\linewidth]{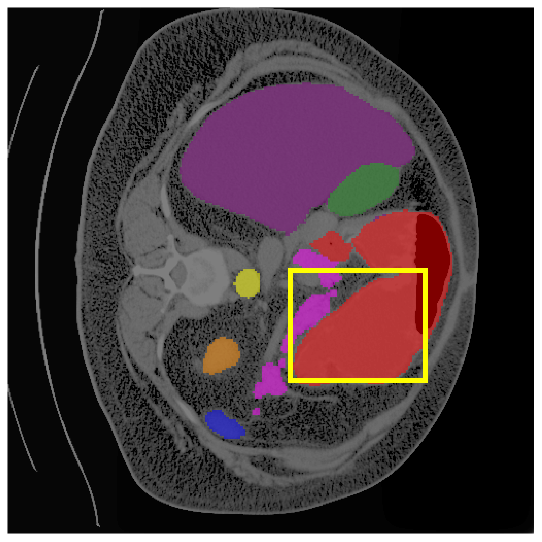} & 
        \includegraphics[width=0.121\linewidth]{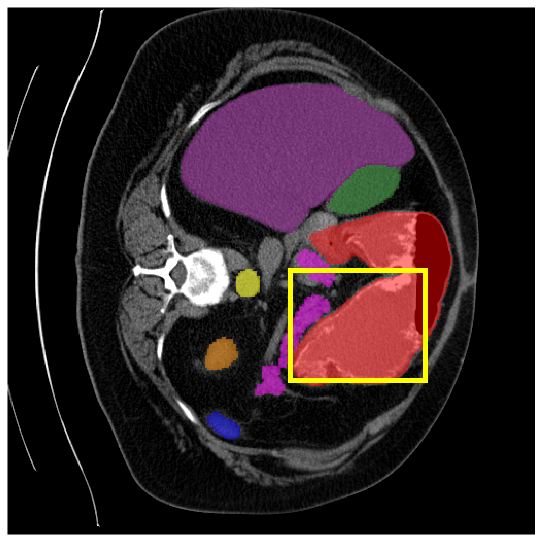} & 
        \includegraphics[width=0.121\linewidth]{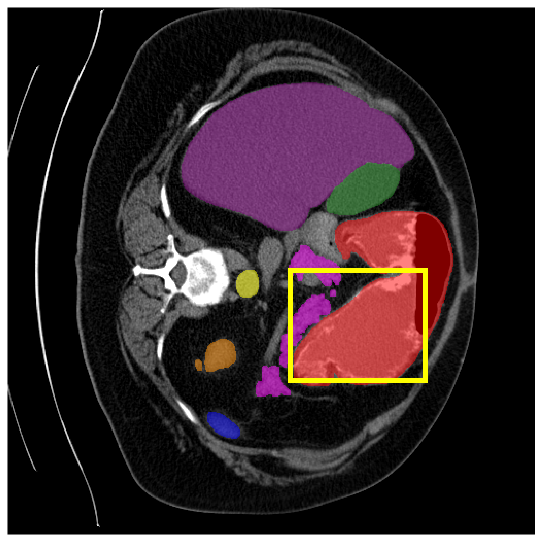} \\
        \includegraphics[width=0.121\linewidth]{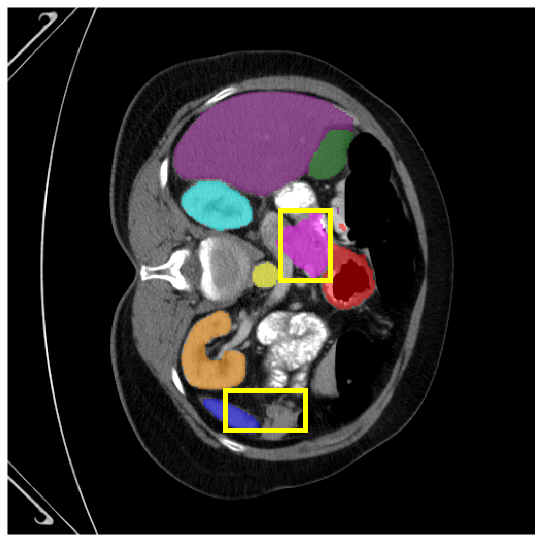} & 
        \includegraphics[width=0.121\linewidth]{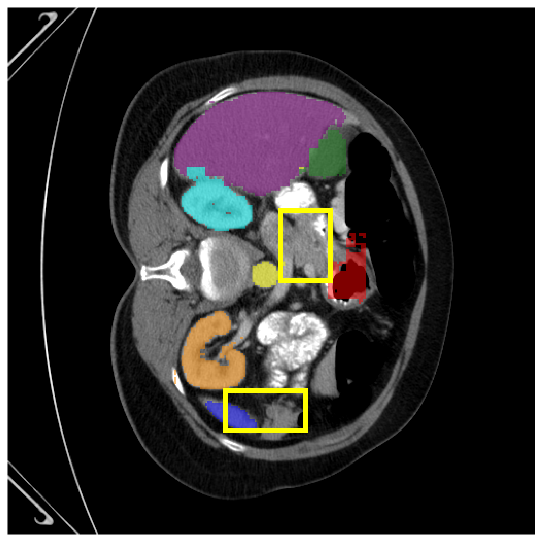} & 
        \includegraphics[width=0.121\linewidth]{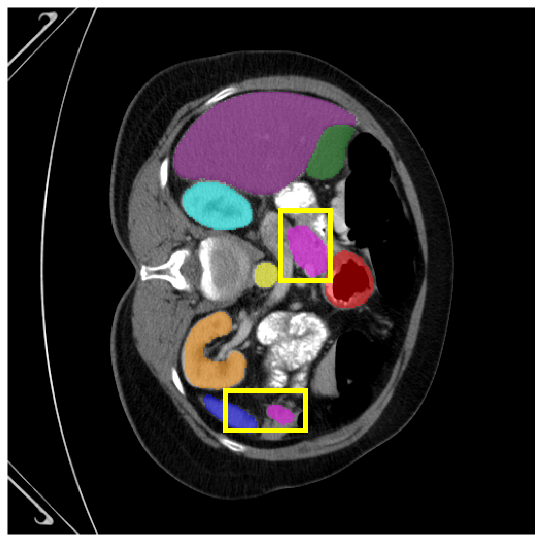} & 
        \includegraphics[width=0.121\linewidth]{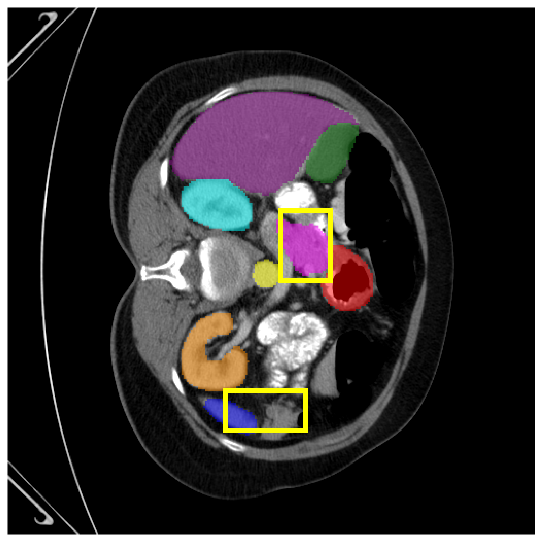} & 
        \includegraphics[width=0.121\linewidth]{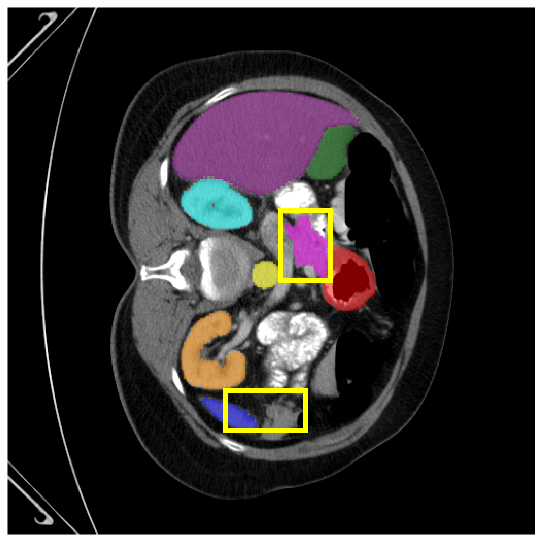} & 
        \includegraphics[width=0.121\linewidth]{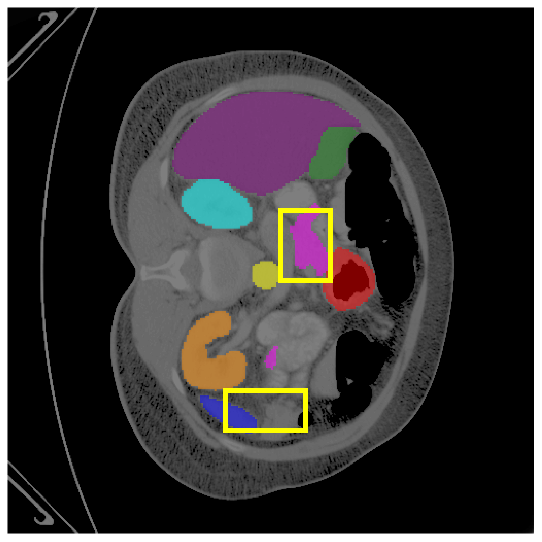} & 
        \includegraphics[width=0.121\linewidth]{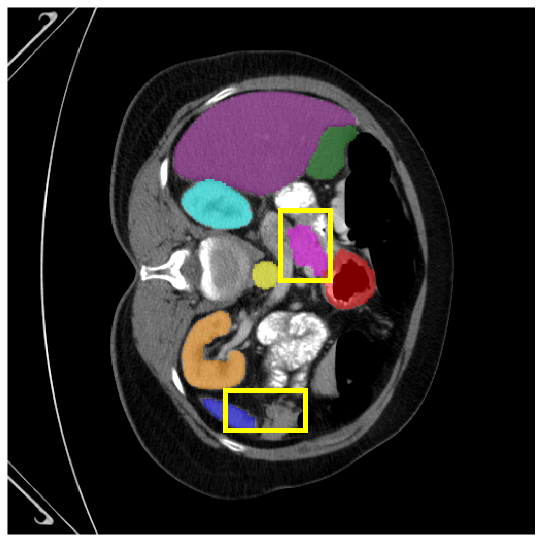} & 
        \includegraphics[width=0.121\linewidth]{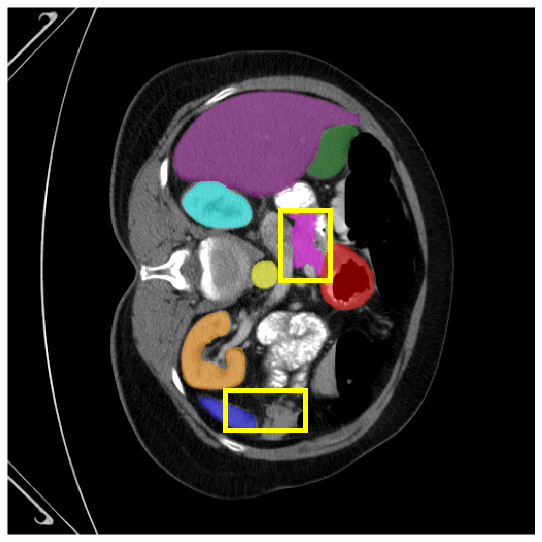} \\
        \includegraphics[width=0.121\linewidth]{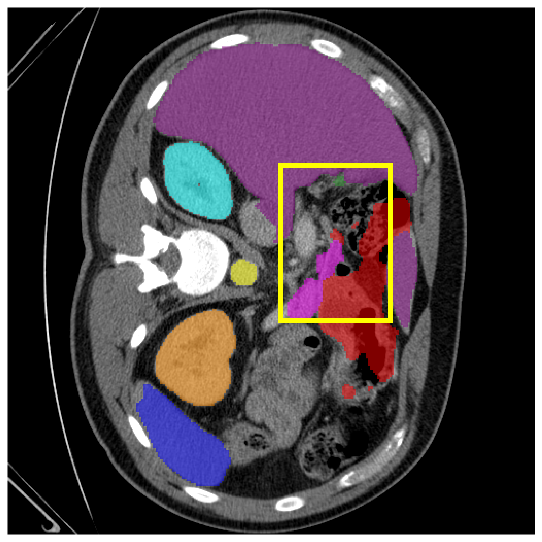} & 
        \includegraphics[width=0.121\linewidth]{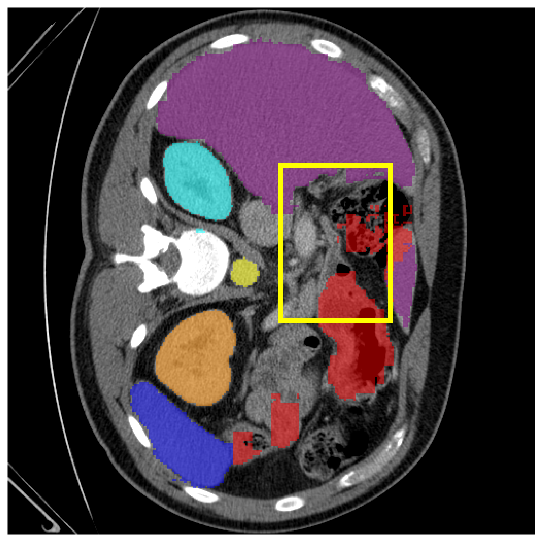} & 
        \includegraphics[width=0.121\linewidth]{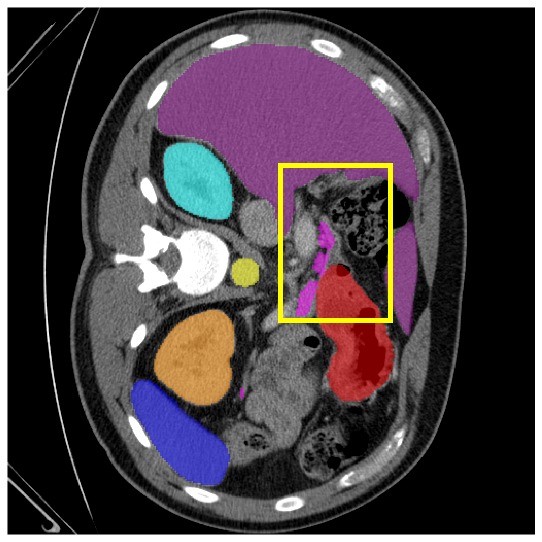} & 
        \includegraphics[width=0.121\linewidth]{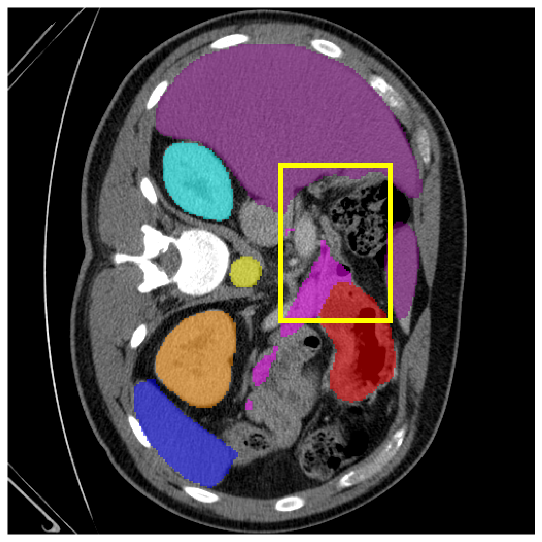} & 
        \includegraphics[width=0.121\linewidth]{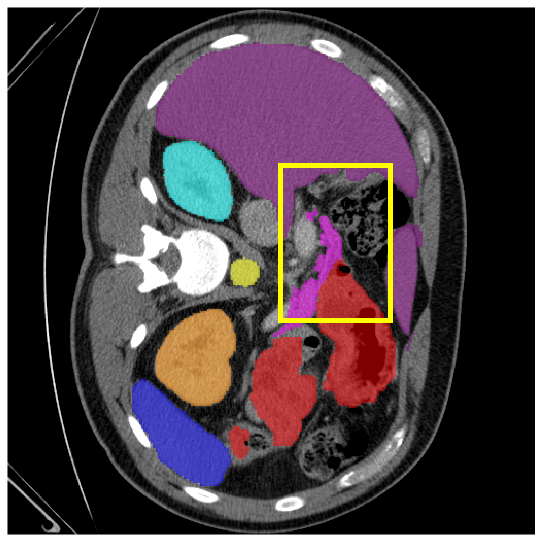} & 
        \includegraphics[width=0.121\linewidth]{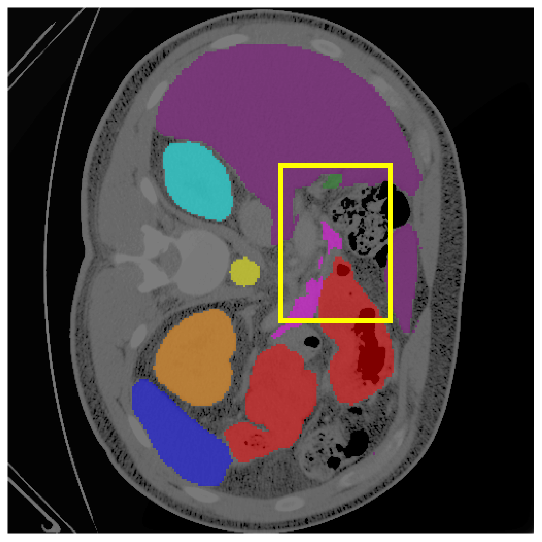} & 
        \includegraphics[width=0.121\linewidth]{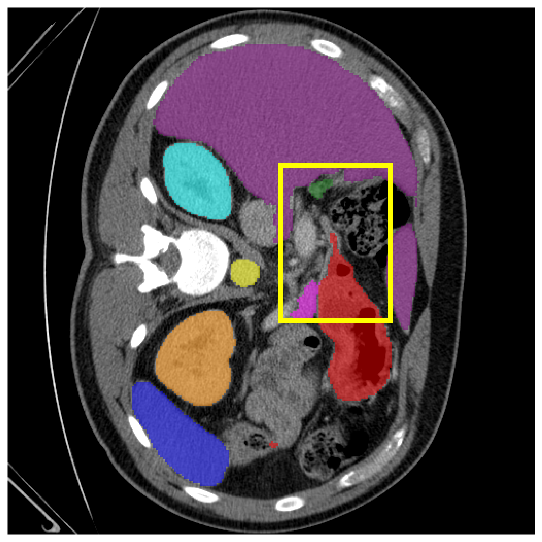} & 
        \includegraphics[width=0.121\linewidth]{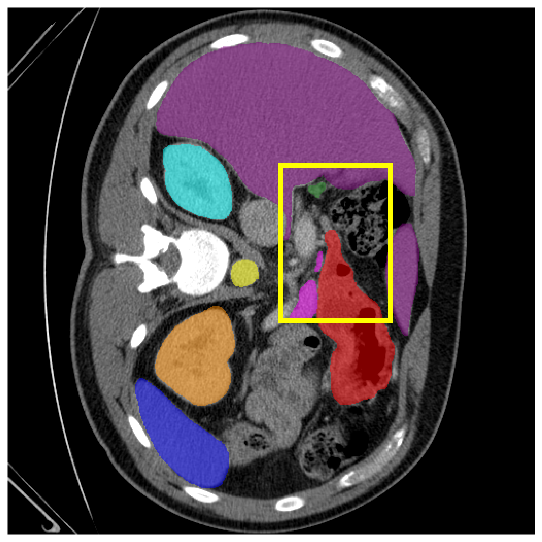} \\
        \includegraphics[width=0.121\linewidth]{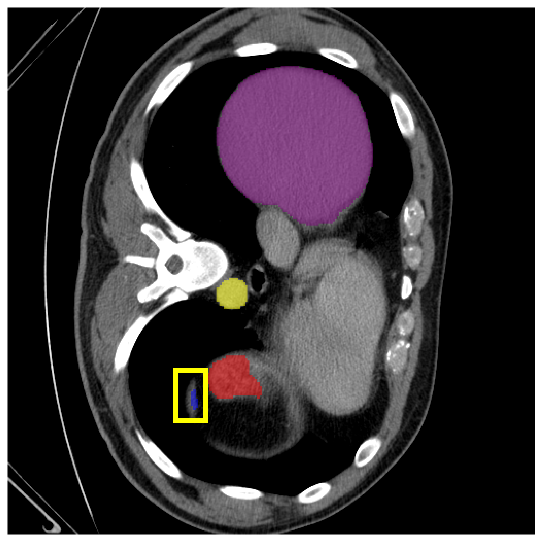} & 
        \includegraphics[width=0.121\linewidth]{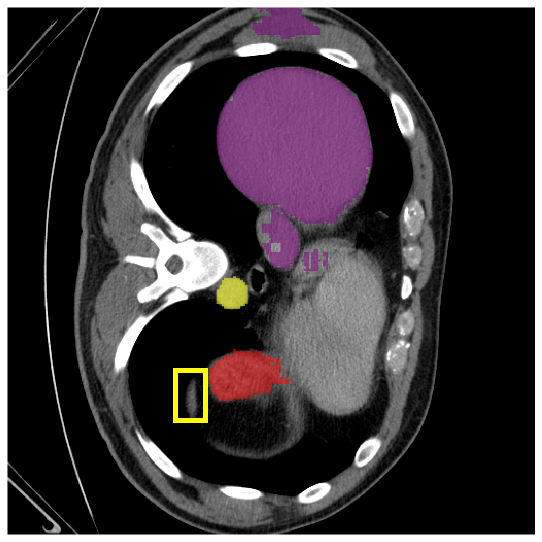} & 
        \includegraphics[width=0.121\linewidth]{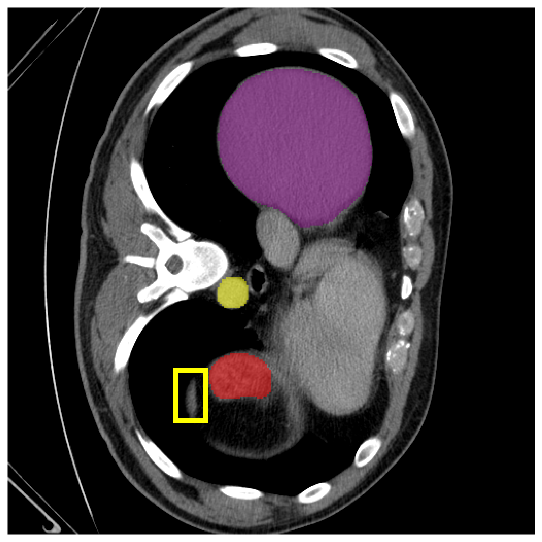} & 
        \includegraphics[width=0.121\linewidth]{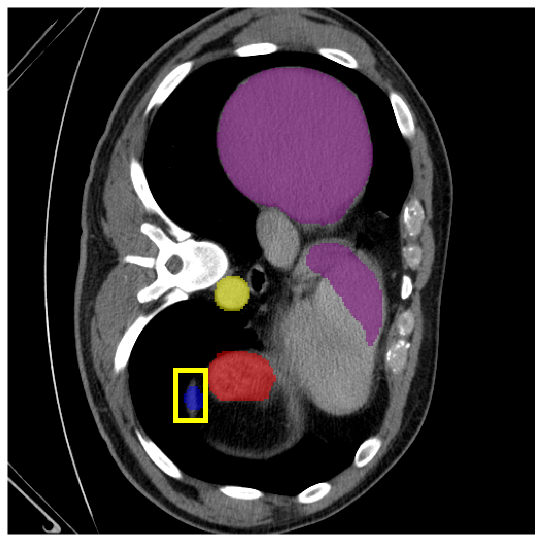} & 
        \includegraphics[width=0.121\linewidth]{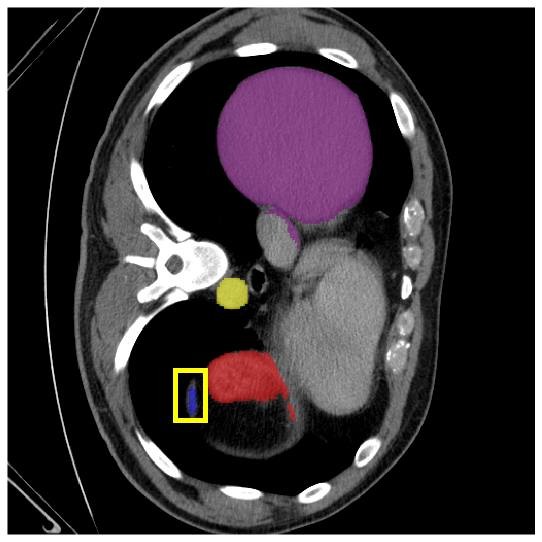} & 
        \includegraphics[width=0.121\linewidth]{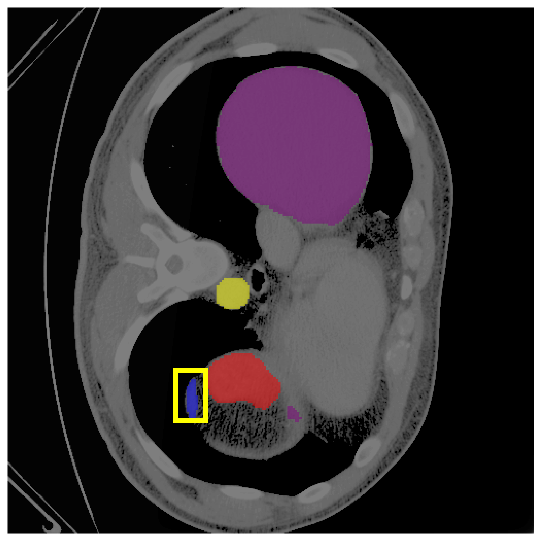} & 
        \includegraphics[width=0.121\linewidth]{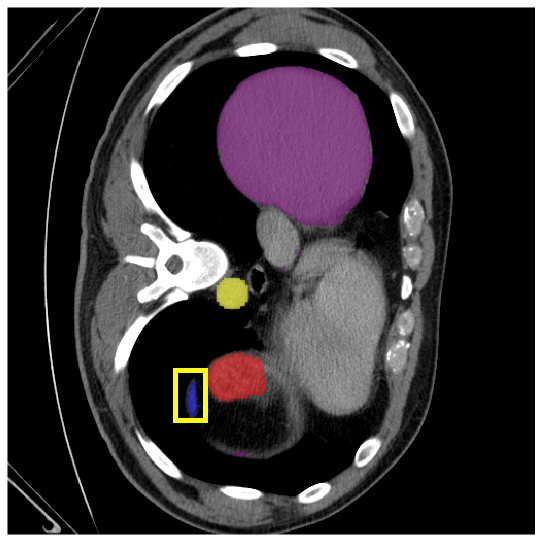} & 
        \includegraphics[width=0.121\linewidth]{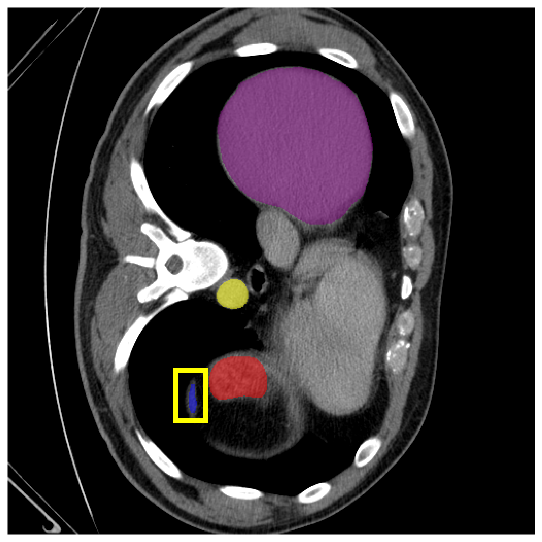} \\
        \includegraphics[width=0.121\linewidth]{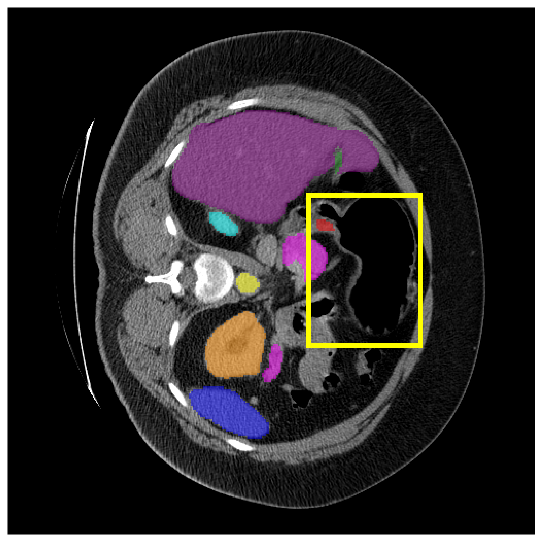} & 
        \includegraphics[width=0.121\linewidth]{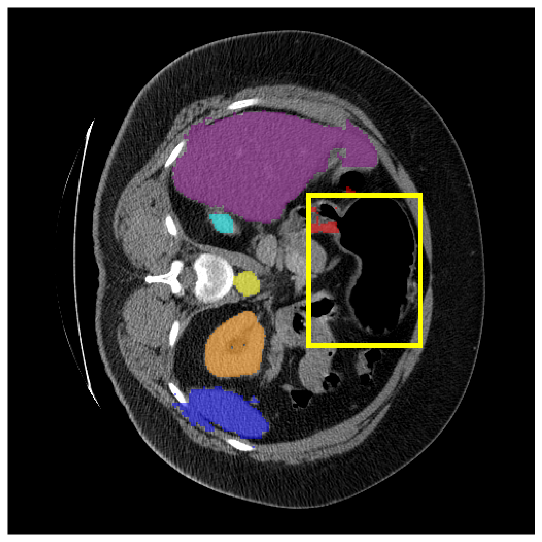} & 
        \includegraphics[width=0.121\linewidth]{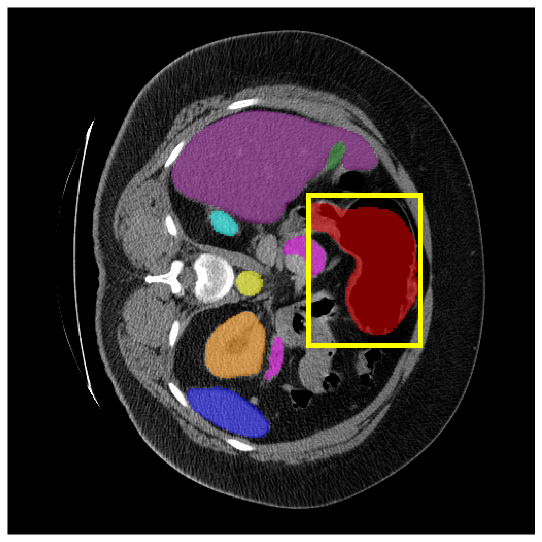} & 
        \includegraphics[width=0.121\linewidth]{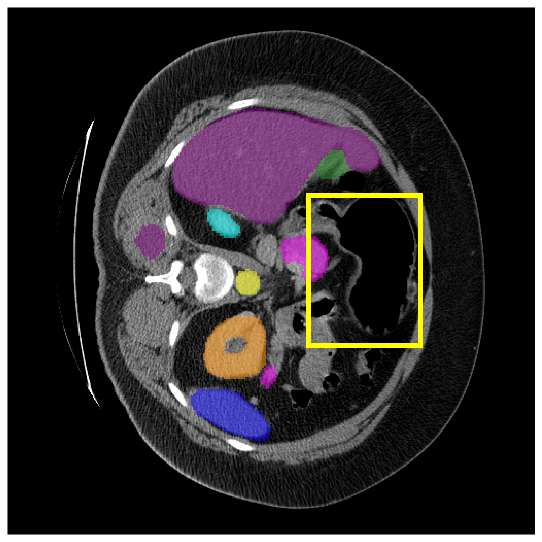} & 
        \includegraphics[width=0.121\linewidth]{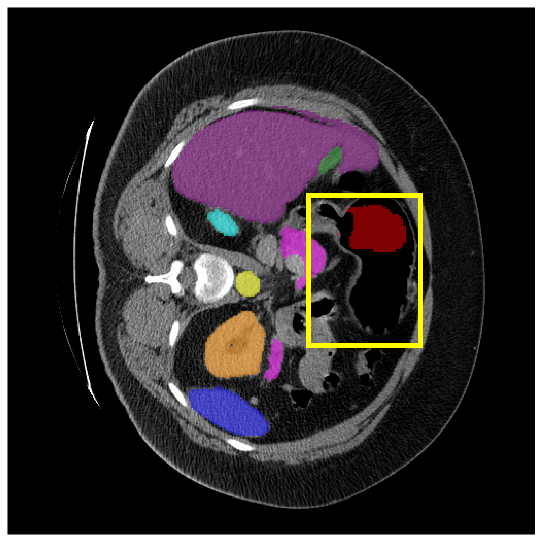} & 
        \includegraphics[width=0.121\linewidth]{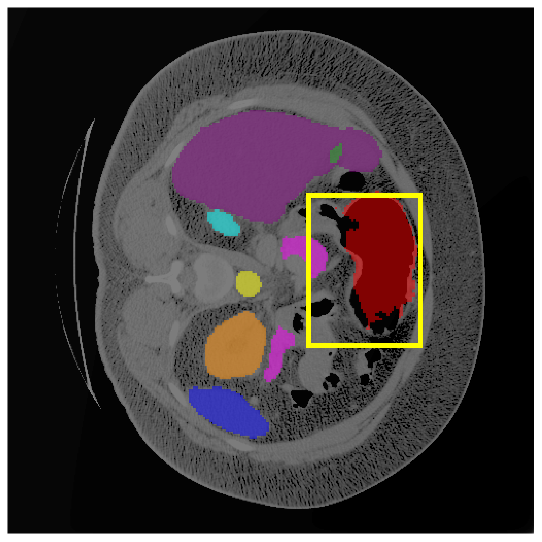} & 
        \includegraphics[width=0.121\linewidth]{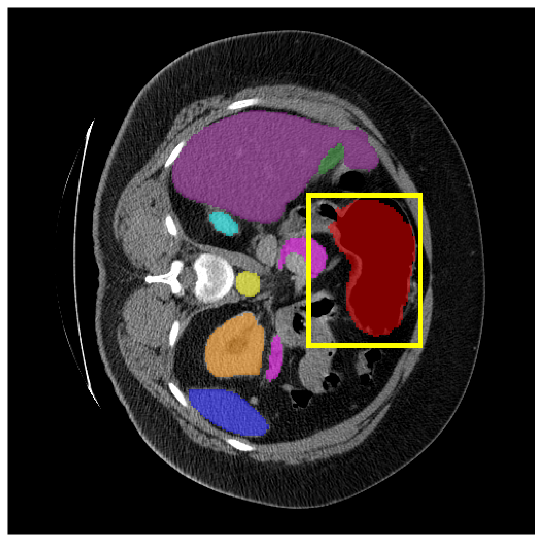} & 
        \includegraphics[width=0.121\linewidth]{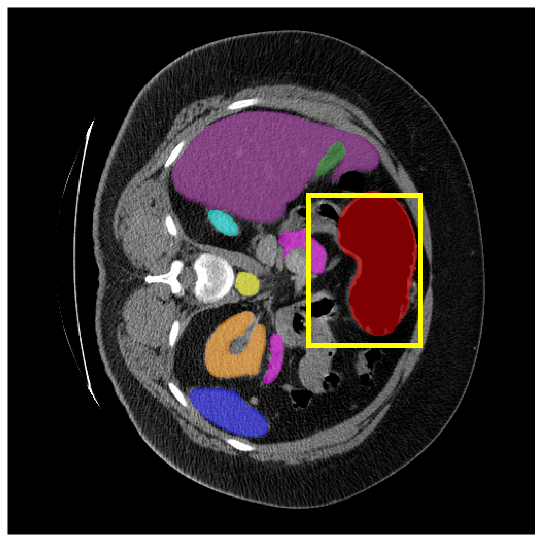} \\
        \footnotesize (a) & \footnotesize (b) & \footnotesize (c) & \footnotesize (d) & \footnotesize (e) & \footnotesize (f) &  \footnotesize (g) & \footnotesize (h)
    \end{tabular}
    \caption{
    Additional visual comparison of segmentation results on the Synapse dataset. (a) TransUNet~\cite{chen2021transunet}, (b) SwinUNet~\cite{cao2022swinunet}, (c) MERIT~\cite{rahman2023multi}, (d) FCT~\cite{tragakis2023fully}, (e) only $I$, (f) only $E$, (g) Ours, and (h) GT, respectively.
    Yellow boxes highlight regions in which our method excels at segmentation.
    %
    }
    \label{fig:synapse_result}
\end{figure*}

To demonstrate the superiority of our method, additional qualitative comparison results for the ACDC and Synapse datasets are presented in \figref{fig:acdc_result}, \figref{fig:synapse_result}, respectively.
%
Notably, in the case of GB, the quantitative results in \figref{fig:synapse_result} show relatively lower performance in `only $E$'.
However, the visualization results in \figref{fig:synapse_result} reveal instances where the enhanced images identified GB regions that the original images failed to segment, albeit with some boundary over-segmentation.
In such scenarios, our proposed method successfully leveraged the information from enhanced images to achieve more accurate GB segmentation.
Note that this finding underscores the fact that even when quantitative performance metrics appear lower, the additional information provided by enhanced images can be valuable in the actual segmentation process.

\clearpage

\end{document}